\documentclass[runningheads]{llncs}
\usepackage{graphicx}
\usepackage{tikz}
\usepackage{comment}
\usepackage{amsmath,amssymb} %
\usepackage{color, colortbl}
\usepackage[accsupp]{axessibility}  %
\usepackage{booktabs}

\usepackage[utf8]{inputenc} %
\usepackage[T1]{fontenc}    %
\usepackage{url}            %
\usepackage{amsfonts}       %
\usepackage{nicefrac}       %
\usepackage{microtype}      %
\usepackage{multirow}       %
\usepackage{algorithm}      %
\usepackage{algorithmic}    
\usepackage{wrapfig}
\usepackage{pifont}%
\usepackage[multiple]{footmisc}
\usepackage{makecell}
\usepackage{caption}
\usepackage{adjustbox}
\usepackage{diagbox} %
\usepackage{changepage}
\usepackage{makecell}
\usepackage{subcaption}
\usepackage{arydshln}
\usepackage{cite}
\usepackage{adjustbox}
\usepackage[pagebackref,breaklinks,colorlinks]{hyperref}

\usepackage{ulem}

\usepackage{letltxmacro}

\newcommand{\bm}[1]{\boldsymbol{#1}}

\newcolumntype{H}{>{\setbox0=\hbox\bgroup}c<{\egroup}@{}}

\definecolor{purple}{RGB}{160, 32, 240}
\definecolor{washblue}{RGB}{186, 224, 228}
\definecolor{sky}{RGB}{128, 128, 128}
\definecolor{seagreen}{RGB}{60, 179, 113}
\definecolor{building}{RGB}{128, 0, 0}
\definecolor{road}{RGB}{128, 64, 128}
\definecolor{sidewalk}{RGB}{0, 0, 192}
\definecolor{fence}{RGB}{64, 64, 128}
\definecolor{vegetation}{RGB}{128, 128, 0}
\definecolor{car}{RGB}{64, 0, 128}
\definecolor{sign}{RGB}{192, 128, 128}
\definecolor{pedestrian}{RGB}{64, 64, 0}
\definecolor{cyclist}{RGB}{0, 128, 192}

\newcommand\mypara[1]{\vspace{1mm}\noindent\textbf{#1}}

\usepackage{mathtools}

\def\be {\begin{equation}}
\def\ee {\end{equation}}
\def\beas {\begin{eqnarray*}}
\def\eeas {\end{eqnarray*}}
\def\bea {\begin{eqnarray}}
\def\eea {\end{eqnarray}}
\def\bes {\begin{equation*}}
\def\ees {\end{equation*}}

\newcommand{\figref}[1]{Fig\onedot~\ref{#1}}

\newcommand{\bbR}{{\mathbb{R}}}

\newcommand{\bx}{\mathbf{x}}
\newcommand{\by}{\mathbf{y}}

\newcommand{\bz}{\mathbf{z}}

\makeatletter
\usepackage{xspace}
\def\@onedot{\ifx\@let@token.\else.\null\fi\xspace}
\DeclareRobustCommand\onedot{\futurelet\@let@token\@onedot}

\makeatother

\begin{document}
\pagestyle{headings}
\mainmatter
\def\ECCVSubNumber{2118}  %

\title{Learning to Generate \\ Realistic LiDAR Point Clouds} %

\titlerunning{Lidar Generation}
\author{Vlas Zyrianov \and Xiyue Zhu \and Shenlong Wang}
\authorrunning{Zyrianov et al.}
\institute{University of Illinois at Urbana-Champaign, IL, USA
\email{\{vlasz2,xiyuez2,shenlong\}@illinois.edu}}
\maketitle

\begin{abstract}
We present LiDARGen, a novel, effective, and controllable generative model that produces realistic LiDAR point cloud sensory readings. Our method leverages the powerful score-matching energy-based model and formulates the point cloud generation process as a stochastic denoising process in the equirectangular view. This model allows us to sample diverse and high-quality point cloud samples with guaranteed physical feasibility and controllability. We validate the effectiveness of our method on the challenging KITTI-360 and NuScenes datasets. The quantitative and qualitative results show that our approach produces more realistic samples than other generative models. Furthermore, LiDARGen can sample point clouds conditioned on inputs without retraining. We demonstrate that our proposed generative model could be directly used to densify LiDAR point clouds. Our code is available at: \url{https://www.zyrianov.org/lidargen/}

\end{abstract}

\section{Introduction}

The past decade witnessed rapid progress in machine perception. Many embodied systems leverage various sensors and the power of deep learning to perceive the world better. LiDAR provides accurate 3D geometry of the surrounding environment, making it one of the most popular sensor choices for various autonomous systems, including self-driving cars~\cite{Kitti360, yang2018pixor, lang2019pointpillars}, surveying drones~\cite{zhang2014loam}, indoor robots~\cite{cao2021tare}, and planetary rovers~\cite{carle2010long}.

Realistic and scalable LiDAR simulation suites are desirable for studying LiDAR-based perception for various reasons. First, LiDAR is an expensive sensor. A 64-beam spinning LiDAR costs over 50,000 USD~\cite{WaymoArsTechnica}. Not everyone can afford one, prohibiting physical sensors from being a scalable and customizable solution for data collection in research. Furthermore, training and testing in safety-critical situations is crucial for autonomy safety. However, collecting data for extreme scenarios in the real world is costly, unsafe, and even unethical. Simulations allow overcoming the above limitations by generating realistic data for long-tail events and training and testing agents at low cost. 

Nevertheless, generating highly realistic and scalable LiDAR data remains an unsolved challenge. Existing approaches are either unrealistic or not scalable. The primary paradigm for creating realistic LiDAR data is through model-based simulation. Early work on LiDAR simulation is purely physics-driven. The general idea is to mimic the time-of-flight (ToF) sensing process of LiDAR~\cite{CARLA}. Specifically, the simulator casts rays in a 3D environment and simulates the receiver's returns by measuring the distance of the hitting surface to the sensor. The reality gap remains substantial because of the imperfect physical model and artist-designed assets. State-of-the-art simulation~\cite{manivasagam2020lidarsim} combines physical modeling with learning components to compensate for complicated rendering artifacts. It produces high realism in both geometry and radiometric appearance. In addition, such a simulation method also gives complete controllability, allowing us to rearrange the scene layout and change viewpoint freely. However, the data-driven approach requires scanning the physical world in advance, which is expensive and not scalable. Recent approaches \cite{caccia} investigated asset-free LiDAR generation using deep generative models to overcome such a limit. Nevertheless, neither controllability nor realism has yet been achieved.

\begin{figure}[tb]
    \centering
    \adjustimage{width=1.0\textwidth}{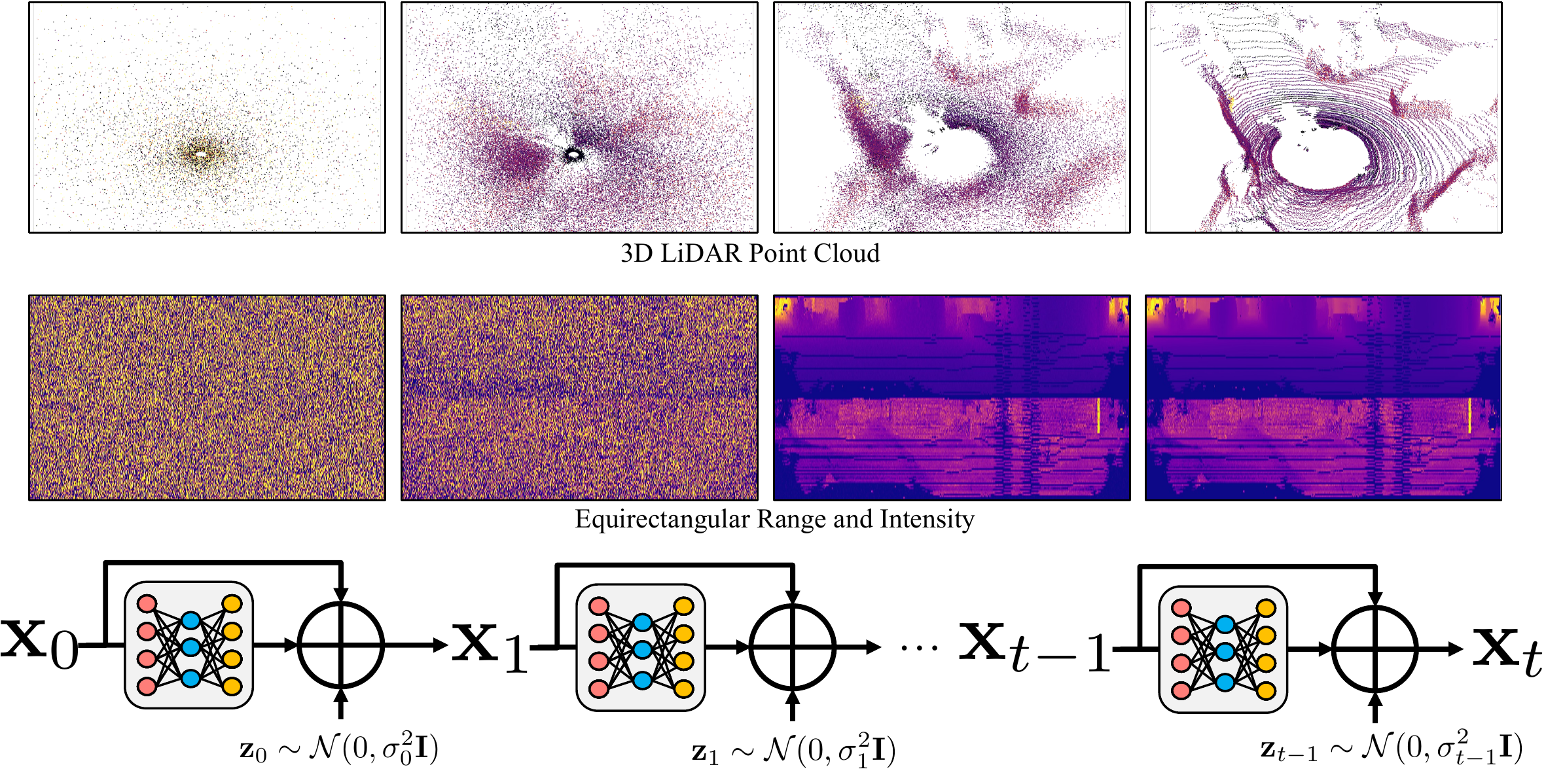}
    \vspace{-5mm}
    \caption{%
 Overview. We sample a LiDAR point cloud by progressively denoising the equirectangular view using a trained score function and Langevin dynamics.     }
    \label{fig:overview}
    \vspace{-5mm}
\end{figure}

In this paper, we present LiDARGen, a \textit{realistic}, \textit{controllable} and \textit{asset-free} LiDAR point cloud generation framework. 
Following the imaging process of spinning LiDAR, we treat each LiDAR scan as an equirectangular view image, a 2.5D panoramic representation encoding information about ray angles, reflectance, and depth range. 
Generating LiDAR points under this representation guarantees physical feasibility.  Inspired by the success of score-matching diffusion models in image generation~ \cite{song2019generative}, LiDARGen then learns a score function~\cite{hyvarinen,vincent2011connection}, modeling the log-likelihood gradient given a sample in the equirectangular image space. 
This score function is trained on real-world LiDAR datasets. 
In the sampling stage, our method gradually converts an initial Gaussian random noise point into a realistic point cloud by progressively applying the score function to remove the noise via Langevin dynamics~\cite{welling2011bayesian, song2019generative}.  \figref{fig:overview} depicts an overview of our approach.

LiDARGen can be applied to conditional generation, such as LiDAR densification, 
by sampling from a posterior distribution~\cite{ScoreSDE}. Specifically, we leverage Bayes' rule to calculate the prior gradient based on the score-matching generative model and the likelihood gradient reflecting the conditions. The generated results are both realistic and plausible with regard to the controlled input. Notably, we also enjoy the simplicity -- such a controlled generation process does not require retraining new models. 

We validate the effectiveness of our method on the KITTI-360 ~\cite{Kitti360} and NuScenes ~\cite{nuscenes2019} datasets. Results demonstrate superior performance compared to other competing methods in various metrics and visual quality. We further evaluate LiDAR densification performance, demonstrating LiDARGen's potential for downstream tasks.

\section{Related Work}

This work studies the problem of generating realistic 3D LiDAR point clouds. It closely relates to point cloud generation and 3D deep learning. 
We also draw upon various efforts in energy-based generative models and LiDAR simulation.

\subsection{Point Cloud Generation}
Our task of generating LiDAR point clouds belongs to the broad category of 3D point cloud generation~\cite{achlioptas_L3DP,gadelha2018multiresolution,zamorski2018adversarial,sun2018pointgrow,3D-AAE,pointsetgen,shu20193d,valsesia2018learning}. Various works successfully apply deep generative models to advance this task. Representative works include varational autoencoder~\cite{pointsetgen, han2019multi, 3D-AAE}, generative adversarial networks~\cite{achlioptas_L3DP, shu20193d, li2018point, zamorski2018adversarial, valsesia2018learning}, flow-based methods~\cite{sun2018pointgrow, yang2019pointflow}, and diffusion processes~\cite{luo2021diffusion, cai2020learning, yang2020energy}. 

Most previous methods on point cloud generation treat point clouds as fixed-size data~\cite{achlioptas_L3DP,zamorski2018adversarial,3D-AAE,pointsetgen,shu20193d,valsesia2018learning}, restricting their practicability in handling real-world data, where the number of points varies significantly. Recent works~\cite{yang2019pointflow, luo2021diffusion} start to look into point cloud generation with a diverse size. However, to achieve this, these approaches require an additional resampling step from an implicit surface distribution or an expensive auto-regressive procedure. In our work, inspired by the physics of LiDAR sensing, we explicitly generate a mask from the range image, mimicking the ray-drop patterns of LiDAR. 
This masking operation allows us to generate point clouds with various sizes and guaranteed physical feasibility. 

Most aforementioned point cloud generation methods are developed and evaluated on clean, synthetic object shapes, such as ShapeNet~\cite{shapenet}. Due to several unique challenges, it remains unclear whether we can directly transfer this success to LiDAR point clouds. First, LiDAR scans are generated through a time-of-flight sensing process. An ideal generator should produce a point cloud following light transport physics. Additionally, LiDAR point clouds are partial observations of a large scene, making the data highly unstructured, sparse, and non-uniform. It is therefore much more challenging to generate realistic LiDAR point clouds. Several recent works explore this direction with moderate success~\cite{caccia, sallab2019lidar}. The pioneering work~\cite{caccia} applies a variational autoencoder and generative adversarial network on LiDAR point clouds. Another line of work also leverages GANs but exploits a hybrid representation~\cite{sallab2019lidar}. However, the level of realism is still limited.

\subsection{Deep Learning for Point Clouds} 

One of the core challenges for generating a 3D point cloud is to select a good representation. 
This subsection briefly categorizes them according to the representation. %

\textbf{Voxelization} methods build a dense volumetric representation in the form of a 3D grid~\cite{qi2016volumetric,choy20163d, graham20183d, pvcnn, pvnas, yang2018pixor}. 3D convolution could then be directly applied to such representation. It is simple and straightforward. 
However, the dense voxel representation suffers from resolution loss and inefficient memory. 
Hence, researchers also improve the 3D voxel method with sparse convolutions~\cite{choy20163d, graham20183d}, hierarchical structures~\cite{yuan2018ocnet} and hybrid representations~\cite{pvcnn, pvnas}.  

Many neural networks directly learn to represent the \textbf{raw point cloud}. The pioneering work PointNet~\cite{pointnet, pointnet++, lin2017structured} leverages aggregation to collect context information. Other lines of research propose novel convolutional operators that can be directly applied on point clouds~\cite{wang2018pccn, li2018pointcnn, wu2019pointconv, xu2018spidercnn, hu2020randla, mao2019interpolated, thomas2019kpconv, su2018splatnet, simonovsky2017dynamic}. Graph-based methods explicitly create graph structures from the point cloud and exploit graph neural networks onto the structures~\cite{simonovsky2017dynamic, zhao2019pointweb, wang2018local, wang2019dgcnn}.

Another popular line of research models 3D data by projecting it onto {2D perspectives}~\cite{su15mvcnn,li2016vehicle,chen2017multi,kanezaki2018rotationnet,lang2019pointpillars, milioto2019rangenet++, yang2018pixor}. One can then directly apply 2D deep learning algorithms for 3D tasks. In these works, the depth value is often encoded in the 2D map, providing partial yet compact information about 3D. \textbf{Bird's eye view representation}~\cite{yang2018pixor, lang2019pointpillars} is established through orthographically projecting the 3D point cloud along the vertical direction. 
\textbf{Perspective projection} obtains a 2D representation that resembles human vision. Several early works exploit perspective projection to produce images from multiple views and fuse the decision in 3D \cite{chen2017multi,su15mvcnn}. 
The most closely related representation to us is the \textbf{equirectangular view representation}, which encodes the polar coordinate into a 2.5D image~\cite{cohen2018spherical, SqueezeSeg, milioto2019rangenet++}. It provides a panoramic view of the surrounding scene that closely resembles the imaging process of LiDAR.  

\begin{figure}[tb]
    \centering
    \adjustimage{width=1\textwidth}{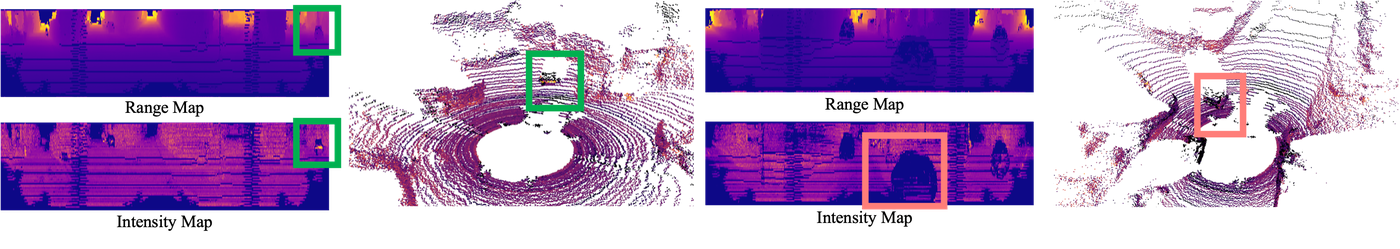}
 \vspace{-7mm}
    \caption{%
 LiDARGen sampling output. Left: equirectangular view. Right: 3D point cloud colored by intensity. Our method learns to generate cars with highly detailed structures, drop rays around transparent region,and produce reflectance intensity for specular objects.}
    \label{fig:details}
 \vspace{-7mm}
\end{figure}

\subsection{LiDAR Simulation}
LiDAR simulation aims to produce a realistic LiDAR point cloud by mimicking the physical process of its imaging~\cite{koenig2004design, CARLA, xiao2021synlidar, manivasagam2020lidarsim, hu2021pattern, sobczak2021lidar, wang2021learning, amini2021vista, xiao2021synlidar, nakashima2021learning}. Physical-based LiDAR simulation uses raycasting methods to simulate LiDAR. Ray intersections are calculated by shooting rays from the origin sensor position outwards onto a geometry surface of the environment. Most self-driving and robotics simulators (e.g., CARLA \cite{CARLA} and Gazebo~\cite{koenig2004design}) typically use this approach. However, physical-based approaches can suffer from a lack of realism because it requires very high-quality 3D assets (e.g., car models with material reflectance information, detailed maps with realistic foliage, etc.), and many LiDAR effects are difficult to simulate (e.g., atmospheric noise or LiDAR ray drop, which occurs when a LiDAR ray reflects off a surface and never gets a reading back).
To approach this problem, research has recently focused on building data-driven LiDAR simulations~\cite{manivasagam2020lidarsim, hu2021pattern, wang2021learning, amini2021vista}. The representative work, LiDARSim~\cite{manivasagam2020lidarsim} leverages machine learning models to learn ray-dropping patterns and exploits 3D assets that are built from modeling the urban environment. 
Many cut-and-paste data augmentation techniques could also be treated as a special form of LiDAR simulation, where objects are removed, inserted, or rearranged into a real LiDAR point cloud to create new ones. However, these methods build upon manipulating real-world point cloud data, restricting their scale and controllability. Furthermore, object insertion often aims to improve the performance of specific perception methods with little consideration for physical plausibility and realism~\cite{fang2020augmented, li2021fooling, gusmao2020development}.

\section{Background}

\subsection{Energy-based Models}
Given a dataset $\{ \bx_1, ... \bx_N \}$, where each data sample $\bx_i$ is assumed to be independently sampled from an underlying distribution, energy-based generative models aim to find a probability model in the following form that best fits the dataset:
$
      p(\bx) = \frac{e^{-E_\theta(\bx)}}{Z_\theta}
$
where the energy function $E_\theta(\bx): \bbR^D \rightarrow \bbR$ is a real-valued function and $\theta$ represents its learnable parameters. $Z_\theta$ is the normalization term $Z_\theta = \int e^{-E_\theta(\bx)} d\bx$ that ensures the $p(\bx)$ to be a valid probability.  Energy-based modeling is a family of expressive yet general probabilistic models that capture underlying dependencies among associated variables in high-dimensional space. Many generative models are instantiations of energy-based models, such as restricted Boltzmann machine~\cite{hinton2006reducing}, conditional random field~\cite{lafferty2001conditional}, factor graphs~\cite{kschischang2001factor} and recent deep energy-based generative models~\cite{song2019generative, song2019sliced}. 
Nevertheless, learning a generic energy model by maximizing log-likelihood is difficult, since computing the partition function $Z_\theta$ or estimating its gradient $\nabla_\theta \log Z_\theta$ is computationally intractable due to the integration over a high-dimensional space.

\subsection{Score-based Energy Models}
To alleviate this problem, researchers try to approximate the computation of the log-likelihood (or the gradient of the partition), with methods such as pseudo-likelihood~\cite{besag1975statistical}, variational inference~\cite{VAE}, and contrastive divergence~\cite{hinton2006reducing}. Other methods bypass it using other learning objectives to train energy-based models, such as structured loss minimization~\cite{hazan2010direct}. Among them, score matching~\cite{hyvarinen2005estimation} recently became a popular choice thanks to its simplicity. Formally speaking, the goal of score matching is to minimize the following objective:
\begin{equation}
    \mathbb{E}_{p_\mathrm{data}(\bx)} \left[ \mathrm{tr}(\nabla_{\bx}^2 \log p_\theta(\bx)) + \frac{1}{2} \| \nabla_{\bx} \log p_\theta(\bx) \|^2_2 \right]
\label{eq:scorematching}
\end{equation}
As shown in this equation, the objective only involves the first and second-order gradient w.r.t. the $\bx$, both of which are independent of the partition function. 

Based on this, Hyvarinen~\cite{hyvarinen2005estimation} proposes the score-based model. It directly models the gradient of log-likelihood $\log p(\bx)$ with a parametric score function $s_\theta(\bx) = \nabla_\bx \log p(\bx): \bbR^D \rightarrow \bbR^D$. However, minimizing the original score matching objective in Eq.~\ref{eq:scorematching} involves calculating the gradient of the Hessian of the log-likelihood $\nabla_\theta \nabla_\bx s_\theta(\bx)$, which is computationally expensive. {Vincent et al.~\cite{vincent2011connection} further reformulates} the score matching energy and shows that score-based models could be more efficiently trained with the following denoising objective:
\begin{equation}
    \frac{1}{2} \mathbb{E}_{p_\mathrm{data}(\bx)} \mathbb{E}_{\Tilde{\bx} \sim \mathcal{N}(\bx, \sigma^2 I)}
    \left[ \left \| s_\theta(\Tilde{\bx}) + \frac{\Tilde{\bx} - \bx}{\sigma^2} \right \|_2^2 \right]
\label{eq:denoisingscore}
\end{equation}
where $\Tilde{\bx}$ is the Gaussian-noise perturbed sample and $\sigma$ is the standard deviation.   

Sampling from a score-based model $s_\theta(\bx)$ can be done with Langevin dynamics~\cite{welling2011bayesian}. It is a Markov chain Monte Carlo (MCMC) process that can be interpreted as a noisy form of gradient ascent. For each step, Langevin dynamics sums the value of the previous step, the current gradient estimation based on the score function, and a Gaussian-distributed random noise: 
\begin{equation}
    {\bx}_t = {\bx}_{t-1} + \frac{\epsilon_t}{2}\nabla_{\bx} \log p({\bx}_{t-1}) + \sqrt{\epsilon_t}\bm{z}_{t}
\label{eq:ld}
\end{equation}
where $\bm{z}_t \sim \mathcal{N}(0,I)$ and $\epsilon_t$ is the learning rate, which is usually decreased (annealed) with a schedule~\cite{song2019generative}. When $\epsilon \rightarrow 0$ and $t \rightarrow +\infty$ Langevin dynamics is convergent to a true samples from the distribution $p(\bx)$ under certain mild conditions~\cite{song2019generative}. The denoising score-matching model and its variants have shown state-of-the-art performance in data generation~\cite{song2019sliced,yang2020energy}.

\begin{table}[!t]
\scriptsize
\centering
\begin{tabular}{ccccc}
Ground-Truth & LiDARGAN~\cite{caccia} & ProjectedGAN~\cite{ProjectedGAN} & \textbf{Ours}\\

\adjincludegraphics[width=0.24\textwidth, trim={{.2\width} {.3\height} {.2\width} {.3\height}},clip]{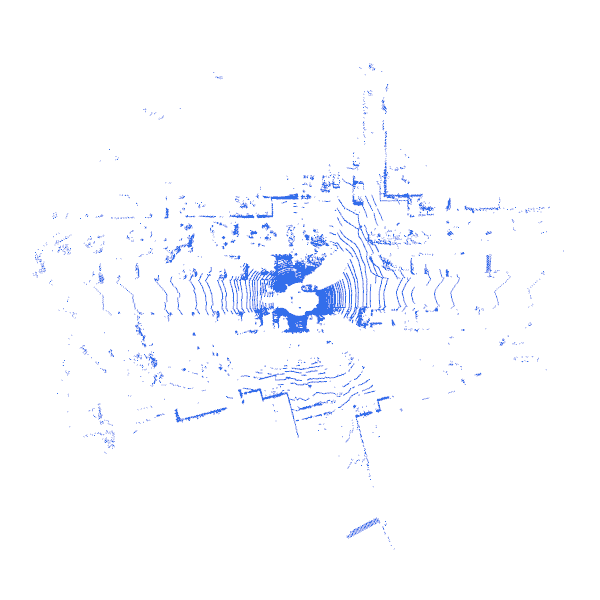} & %
\adjincludegraphics[width=0.24\textwidth, trim={{.2\width} {.3\height} {.2\width} {.3\height}},clip]{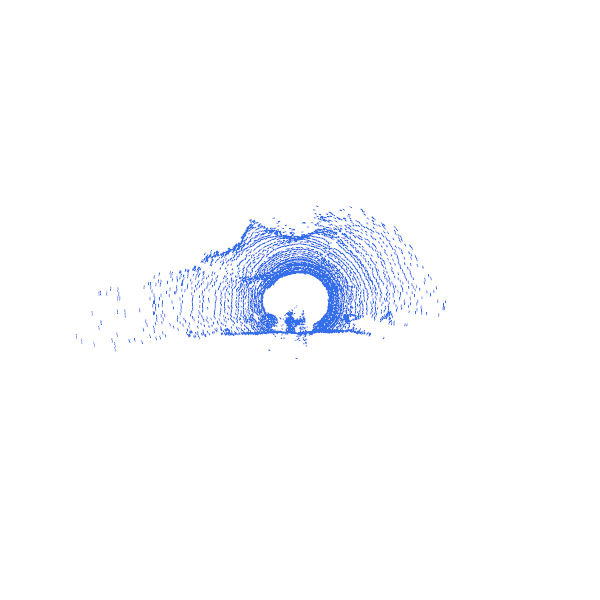} &
\adjincludegraphics[width=0.24\textwidth, trim={{.2\width} {.3\height} {.2\width} {.3\height}},clip]{statement_images/projectedGAN_2.png}  & \adjincludegraphics[width=0.24\textwidth, trim={{.2\width} {.3\height} {.2\width} {.3\height}},clip]{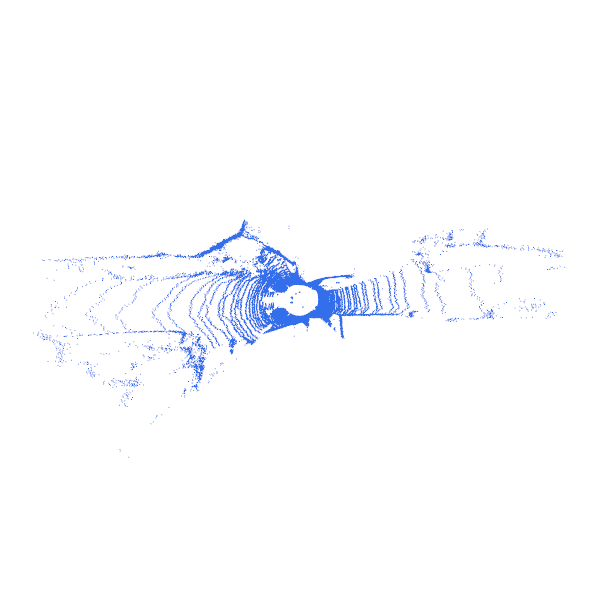}\\

\adjincludegraphics[width=0.24\textwidth, trim={{.2\width} {.3\height} {.2\width} {.3\height}},clip]{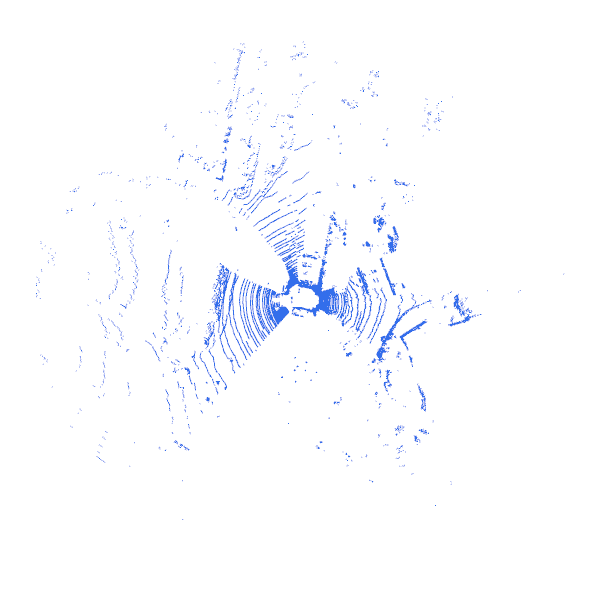} & %
\adjincludegraphics[width=0.24\textwidth, trim={{.2\width} {.3\height} {.2\width} {.3\height}},clip]{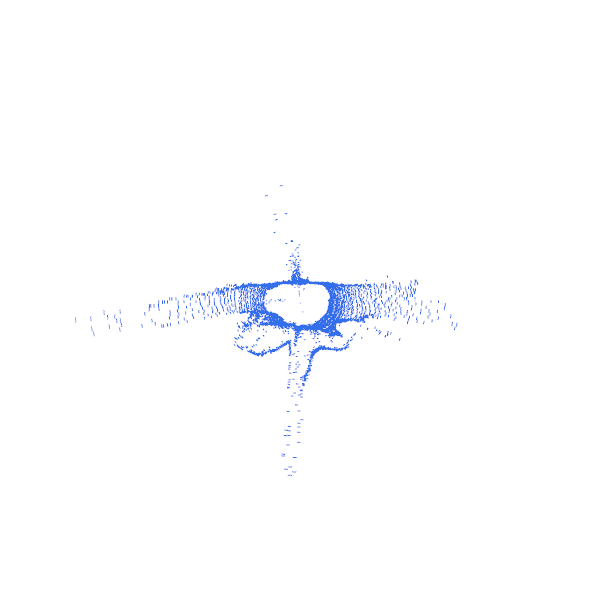} &
\adjincludegraphics[width=0.24\textwidth, trim={{.2\width} {.3\height} {.2\width} {.3\height}},clip]{statement_images/projectedGAN_3.png}  & \adjincludegraphics[width=0.24\textwidth, trim={{.2\width} {.3\height} {.2\width} {.3\height}},clip]{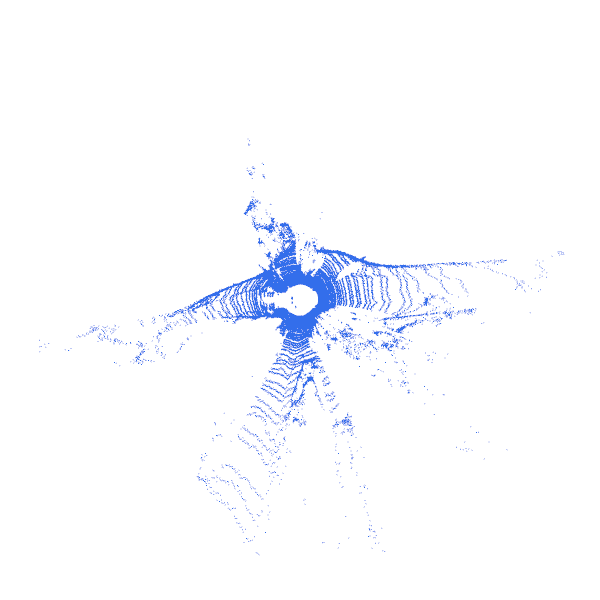}\\

\adjincludegraphics[width=0.24\textwidth, trim={{.2\width} {.3\height} {.2\width} {.3\height}},clip]{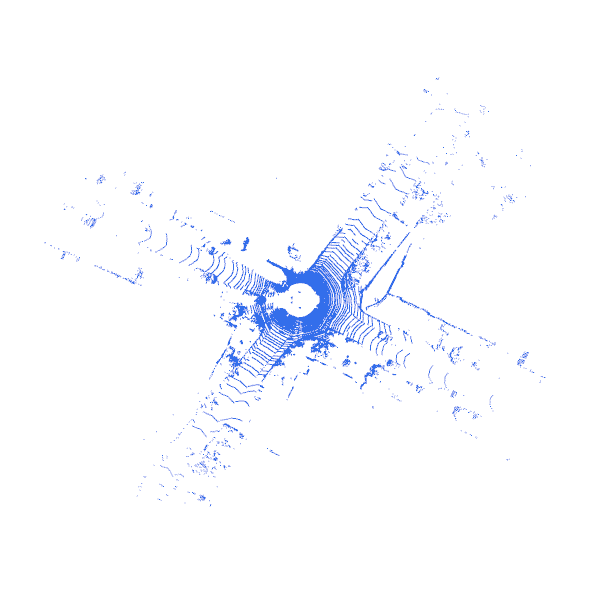} & %
\adjincludegraphics[width=0.24\textwidth, trim={{.2\width} {.3\height} {.2\width} {.3\height}},clip]{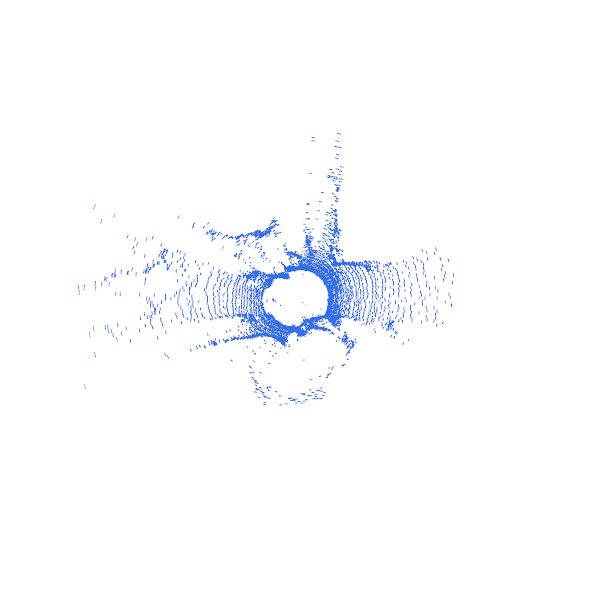} & 
\adjincludegraphics[width=0.24\textwidth, trim={{.2\width} {.3\height} {.2\width} {.3\height}},clip]{statement_images/projectedGAN_4.png}  & \adjincludegraphics[width=0.24\textwidth, trim={{.2\width} {.3\height} {.2\width} {.3\height}},clip]{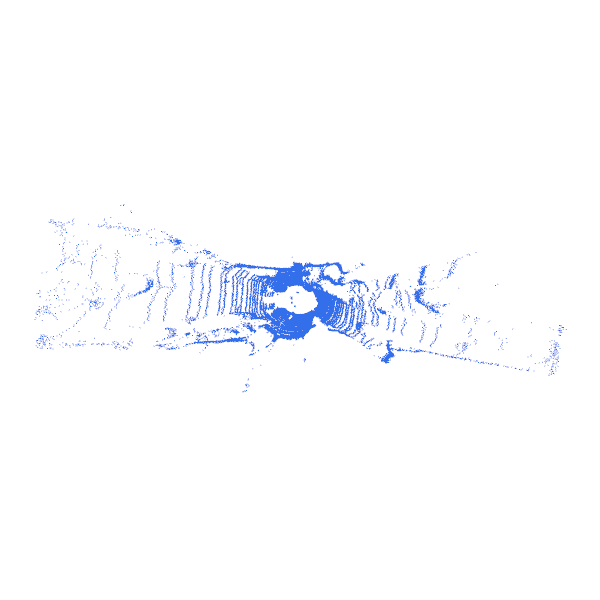}\\

\adjincludegraphics[width=0.24\textwidth, trim={{.2\width} {.3\height} {.2\width} {.3\height}},clip]{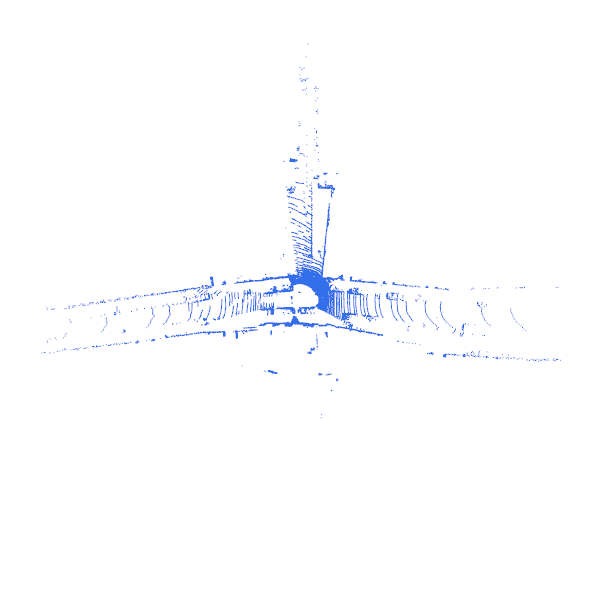} & 
\adjincludegraphics[width=0.24\textwidth, trim={{.2\width} {.3\height} {.2\width} {.3\height}},clip]{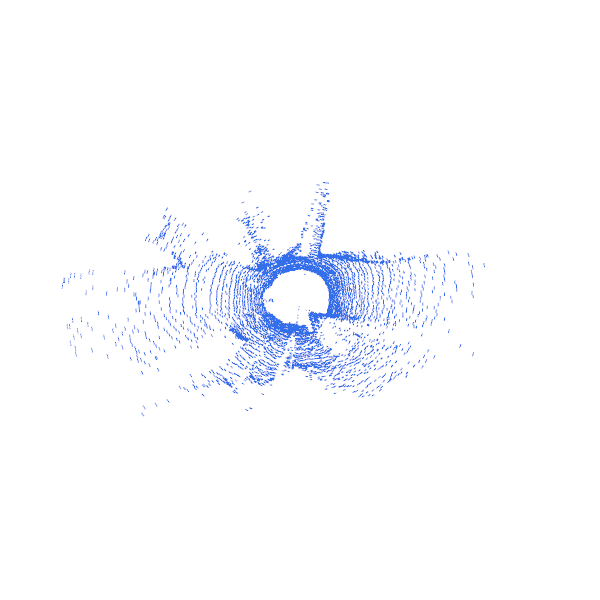} & 
\adjincludegraphics[width=0.24\textwidth, trim={{.2\width} {.3\height} {.2\width} {.3\height}},clip]{statement_images/projectedGAN_5.png}  & \adjincludegraphics[width=0.24\textwidth, trim={{.2\width} {.3\height} {.2\width} {.3\height}},clip]{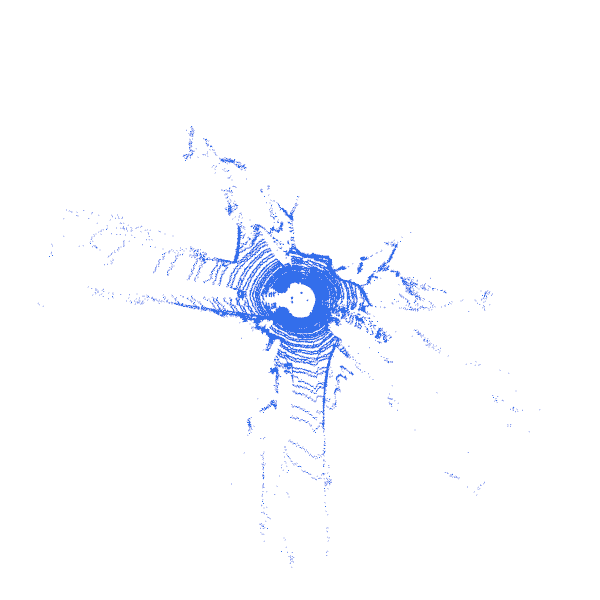}\\
\end{tabular}
\captionof{figure}{Qualitative Results for LiDAR Point Cloud Generation on KITTI-360.}
\vspace{-6mm}
\label{fig:qual}
\end{table}

\section{Method}
Our goal is to model the underlying distribution of LiDAR point clouds in an urban driving scenario. We could then leverage such a generative model to sample new point clouds or use it for downstream MAP inference tasks. The challenge of LiDAR generation is to model the diverse structures that exist in the real world while still maintaining physical plausibility. Towards this goal, we leverage the denoising score-matching generative model~\cite{song2019generative} to model the gradient of its log-probability. Formally speaking, our training dataset consists of a list of raw LiDAR point clouds $\{ (\bx_1, r_i), ... (\bx_N, r_N) \}$, where $\bx_i\in \bbR^{D_i\times 3}$ represents the 3D location and $r_i \in \bbR$ is a scalar representing the reflectance intensity value for each point. Our method learns a score function $s_\theta(\bx)$ to approximate $\nabla_{\bx} \log p(\bx)$ with score-matching \cite{vincent2011connection}. Sampling can then be conducted with Langevin dynamics, which gradually denoises an initial random Gaussian point cloud and returns a clean and realistic point cloud. Inspired by LiDAR's imaging process, we leverage the equirectangular representation as our underlying representation to ensure physical feasibility and develop an encoder-decoder network on top of it as the score function. {Fig.~\ref{fig:overview} gives an overview of our approach. }

\subsection{LiDAR Generation}
\label{sec:generation}

\mypara{Input Representation} 
Our model starts by converting the input parameterization from an unstructured point cloud sparsely distributed in euclidean space into a dense multi-channel equirectangular perspective image, with one channel representing depth and the other representing intensity. More specifically, we first convert each point from the Cartesian coordinate $\bx = ( x, y, z ) \in \bbR^3$ into the spherical coordinate $\bz \in (\theta, \phi, d)$:
$
d = \sqrt{x^2 + y^2 + z^2}, \theta = \arccos\frac{z}{\sqrt{x^2 + y^2 + z^2}}, \phi = \mathrm{atan2}(y, x),
$
where $\theta$ is the inclination, $\phi$ is azimuth and $d$ is the depth range; $\mathrm{atan2}$ is the standard 2-argument arctangent function taking into account the discontinuity across quadrant boundaries of $\mathrm{atan}(y/x)$. Furthermore, we remap the depth range so that it is normalized from 0 to 1: 
The two-channel rectangular image is then produced through quantizing the two angles and rasterization. Concretely, for each point $\bz_i=(\theta_i, \phi_i, d_i), r_i$:
$\mathcal{I}(\lfloor\theta_i / s_\theta\rfloor, \lfloor\phi_j / s_\phi\rfloor) = \left(\frac{1}{6} \log_2(d_i + 1), \frac{1}{255} r_i \right)$. Both channels of the image are normalized to the range $(0, 1)$ and we use a logarithm mapping to esure nearby points have a higher geometry resolution. For simplicity, throughout the rest of section we will also use $\bx$ to represent the point cloud in its equirectangular representation. 
Fig.~\ref{fig:details} demonstrates one example of the range view representation. Our input representation enjoys several benefits. Firstly, it encodes information into a dense and compact 2D map, allowing us to exploit efficient network architecture transferred from the 2D image generation domain. Secondly, due to the ray casting nature, most spinning LiDAR scans will only return the peak pulse for each beam\footnote{some sensors return two beams for a small fraction of beams}. In other words, encoding the point cloud into this representation will not lose any information, and the generated point cloud properly reflects the scanning and ray-casting nature of the sensor. %

\mypara{Network Architecture} Our score-based network $s_\theta$ uses a \textbf{U-Net} architecture~\cite{ronneberger2015u} following its success in image generation~\cite{isola2017image, Ncsnv2}. Specifically, at each step, it takes a $W\times H \times 2$ input image and outputs a $W\times H \times 2$ score map at the same size. We also make important changes suitable for our LiDAR point cloud generation task. Firstly, standard 2D images have disconnected left and right boundaries. Hence zero-padding or symmetry padding is often sufficient for dense prediction. However, equirectangular images are inherently circular. Applying standard convolutions does not take into account such constraints. To alleviate this issue, \textbf{circular convolution} \cite{CircularConvolution} treats left and right boundaries as connected neighbors in its topology. Inspired by this, we substitute all the convolution and pooling layers in our network with circular versions. Second, LiDAR point clouds collected from urban driving environments have a highly structured geometry. And this geometric structure is often viewpoint-aware. For instance, the depth range of the lower beam might have a strong bias due to the ground height; the depth range of the frontal facing positions tends to be larger since the car is mostly driving forward along a straight road. To better encode this prior, our model takes the \textbf{angular coordinate as an additional input} to the convolution, similarly to CoordConv~\cite{liu2018intriguing}.

\mypara{Training} %
One of the difficulties for training denoising score matching models is the choice of a proper noise level for Eq.~\ref{eq:denoisingscore}, which heavily influences the accuracy of score estimation. In practice, we find that having a noise-conditioned extension is crucial for its success. More specifically, we expand our score network $s_\theta({\bx}, \sigma_i)$ to be dependent on the current noise perturbation level $\sigma_i$. At the training stage, following the noise-conditioned score-matching model~\cite{song2019generative}, we adopt a multi-scale loss function, with a re-weighting factor for the loss at each noise level:
\begin{equation}
    \frac{1}{2L} \sum^L_{i = 1} \sigma_i^2  \mathbb{E}_{p_{data}(\bx)} \mathbb{E}_{\Tilde{\bx} \sim \mathcal{N}(\bx, \sigma_i^2 I)}
    \left[ \left \| s_\theta(\Tilde{\bx}, \sigma_i) + \frac{\Tilde{\bx} - \bx}{\sigma_i^2} \right \| \right]
\end{equation}
where $\Tilde{\bx}$ is the randomly perturbed noisy signal at each level, and $\sigma_i$ is the standard deviation of the noise distribution. 

\mypara{Sampling} We exploit annealed Langevin dynamics sampling~\cite{Ncsnv2} for our point generation task to increase sampling efficiency. Specifically, we start from the highest pretrained noise level and gradually reduce the noise level:
\begin{equation}
    {\bx}_t = {\bx}_{t-1} + \gamma \frac{\sigma_i^2}{2 \sigma^2_L} s_\theta({\bx}_{t-1}, \sigma_i) + \gamma \frac{\sigma_i}{\sigma_L}\bz_{t}
\end{equation}
where $\gamma$ is the learning rate and $\sigma_L$ is the smallest noise level. The final step of Langevin dynamic sampling gives us a clean equirectangular range image. We unproject this resulting image back into 3D Cartesian space to recover the 3D point cloud. Please refer to Fig.~\ref{fig:overview} for the full sampling procedure.

\subsection{Posterior Sampling}

Learning the unconditional prior distribution $p(\bx)$ of LiDAR point clouds enables many applications. In particular, we often expect our generated LiDAR point cloud to satisfy a specific property or to be consistent with certain conditions. For instance, we might want to create a LiDAR point cloud that agrees with some partial observation; 
or we might generate a LiDAR point cloud conditioned on its semantic layout. Conventional methods, such as GANs, often require training different conditional generative models for each task. However, thanks to the gradient-based approach used in score-based models, we could efficiently conduct the tasks mentioned above with only the pretrained unconditional generation model $p(\bx)$. Next, we will show how to achieve this in LiDARGen. 

Specifically, given a pretrained generation model $p(\bx)$ and an input condition $\by$, we formulate the agreement between the condition $\by$ and the LiDAR point cloud $\bx$ as a likelihoood function $p(\by | \bx)$. Our goal is to sample new point cloud that reflect the input condition $p(\bx | \by)$. According to the Bayes' rule we have:
\begin{equation}
    p(\bx | \by) = p(\by | \bx) p(\bx) / {p(\by)}, 
    \nabla_\bx \log p(\bx | \by)  = \nabla_{\bx} \log p(\by | \bx) +  \nabla_{\bx} \log  p(\bx)
    \label{eq:posterior}
\end{equation}
where $\nabla_\bx \log p(\bx)$ is our pretrained score function $s_\theta(\bx)$. In many situations, the likelihood model has an analytical gradient or is a neural network; hence calculating the gradient with respect to $\bx$ is straightforward. We, therefore, leverage the following Langevin dynamics to sample from the posterior distribution: 
\begin{equation}
    \bx_t = \bx_{t-1} + \frac{\epsilon}{2} \left( s_\theta(\bx_{t-1}) + \nabla_{\bx} \log p(\by | \bx_{t-1})\right) + \sqrt{\epsilon}\bz_{t}.
    \label{eq:densification}
\end{equation}
Next, we will discuss a concrete application of posterior sampling.

\mypara{LiDAR Densification} Spinning LiDAR used on autonomy is expensive due to its complicated mechanical design. In particular, the price of LiDAR sensors grows exponentially as the number of beams increases. Therefore, there is a practical need to produce high-beam LiDAR readings with a low-beam model. 
Given a low-beam LiDAR point cloud $\by$, our goal is to recover its high-beam version $\bx$. Assuming that $\mathbf{m}$ is the visibility mask denoting pixels with a provided gt ray, the gradient of the posterior can be computed as follows: 
\[ \nabla_\bx\log p(\bx | \by ) = \nabla_\bx \left(\log p(\bx) +  \frac{\lambda}{2} || (\bx - \by) \odot \mathbf{m} ||^2 \right) = s_\theta(\bx) + \lambda \left[ (\bx - \by) \odot \mathbf{m} \right ],\]
where $\odot$ is the Hadamard product. 
Intuitively, each Langevin dynamic step pushes the samples towards the direction of being both realistic and consistent with the partial observation $\by$.

\begin{table}[!t]
\small
\centering
\begin{tabular}{ccc}
Reference & 16-Beam Input & \textbf{Ours}\\
\adjincludegraphics[width=0.32\textwidth, trim={{.2\width} {.3\height} {.2\width} {.3\height}},clip]{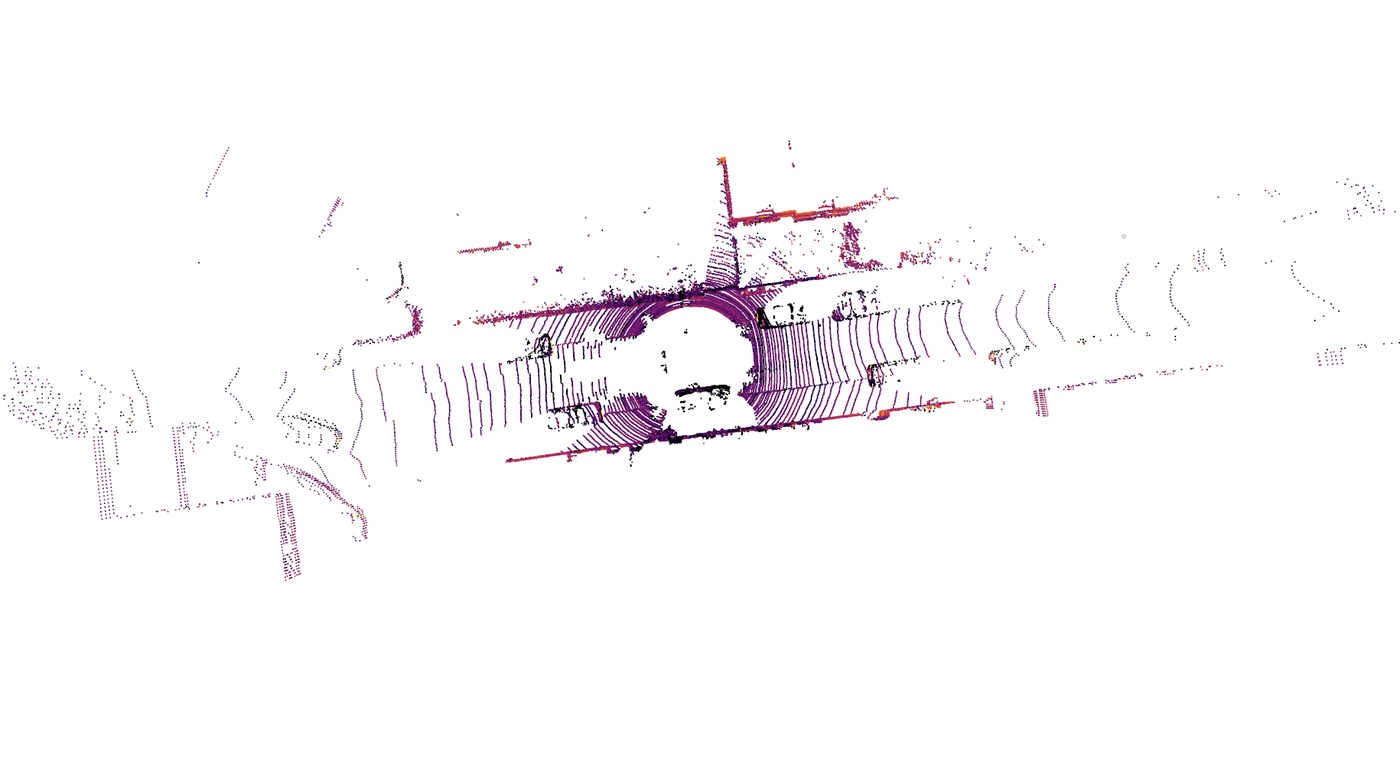} & 
\adjincludegraphics[width=0.32\textwidth, trim={{.2\width} {.3\height} {.2\width} {.3\height}},clip]{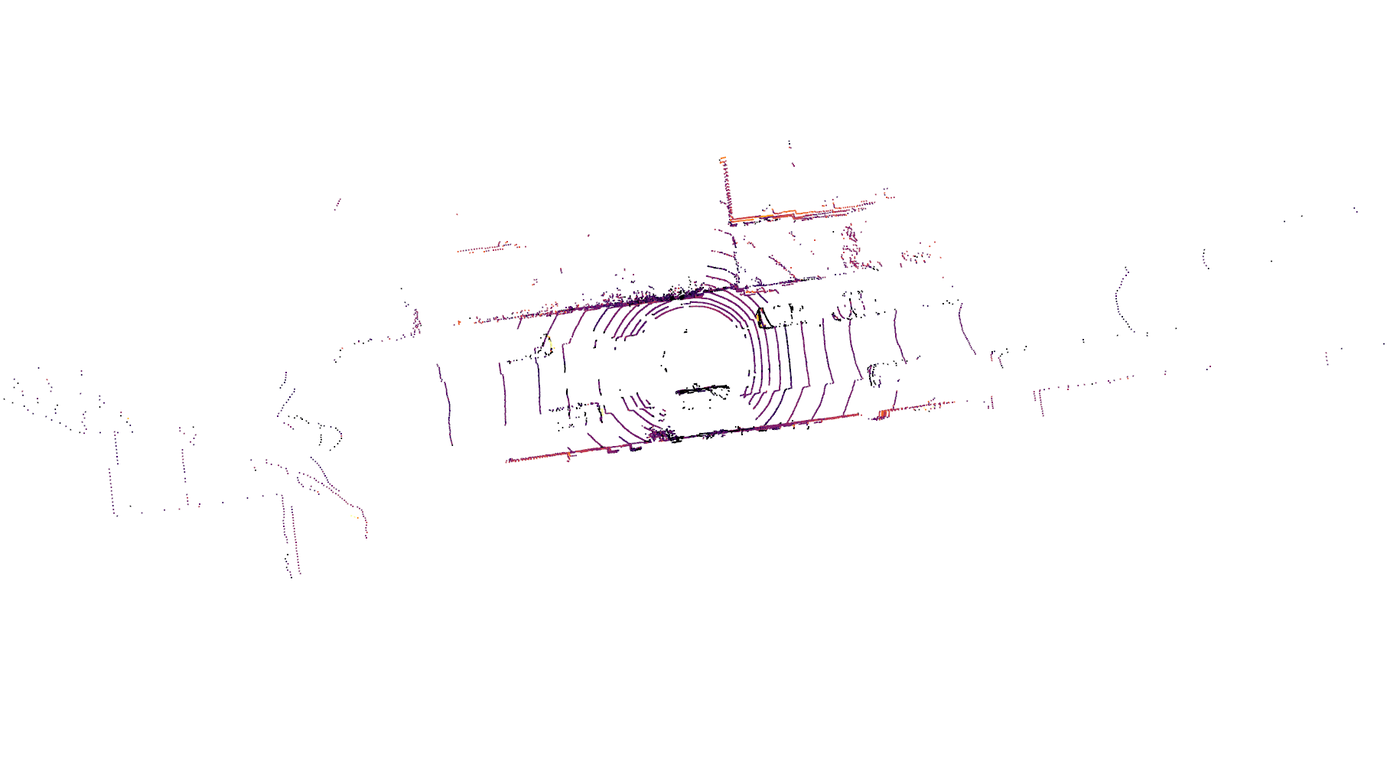} & 
\adjincludegraphics[width=0.32\textwidth, trim={{.2\width} {.3\height} {.2\width} {.3\height}},clip]{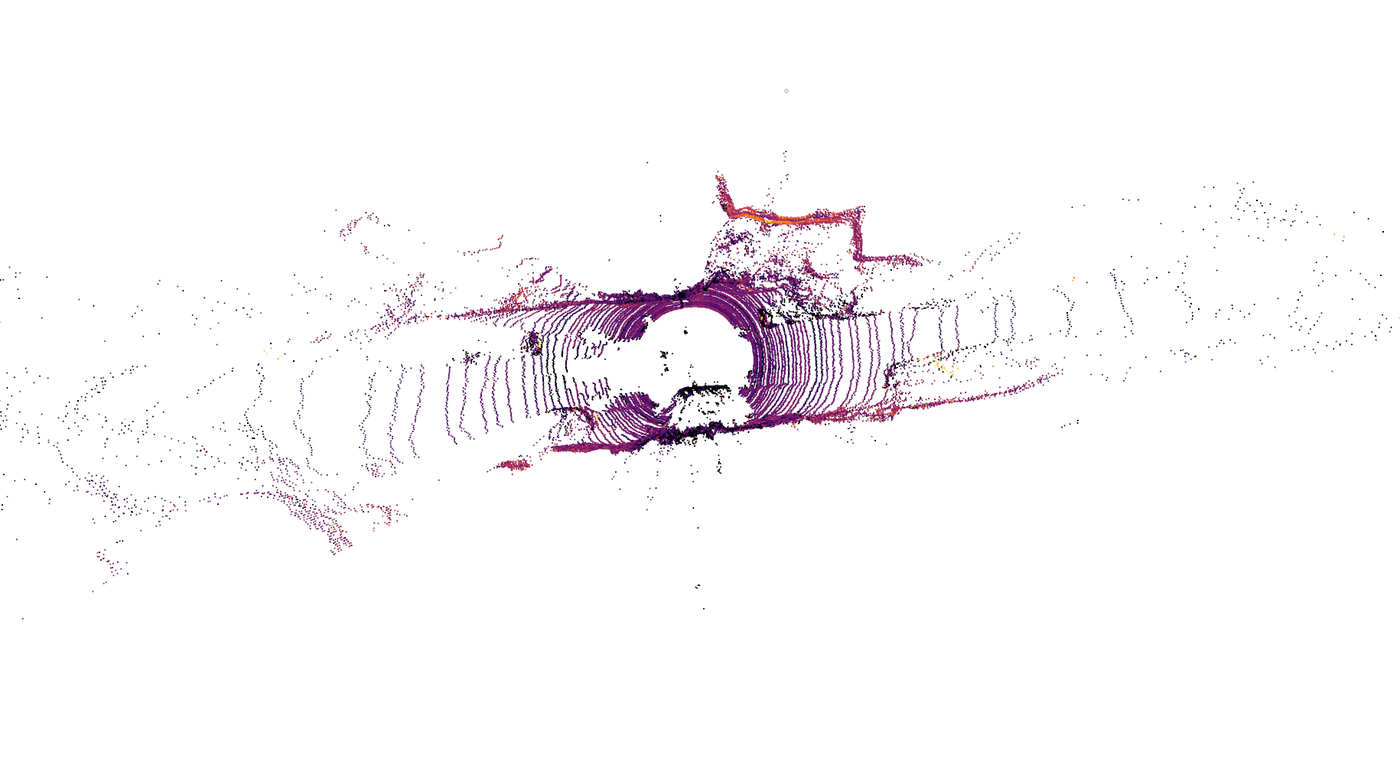} \\
\adjincludegraphics[width=0.32\textwidth, trim={{.2\width} {.3\height} {.2\width} {.3\height}},clip]{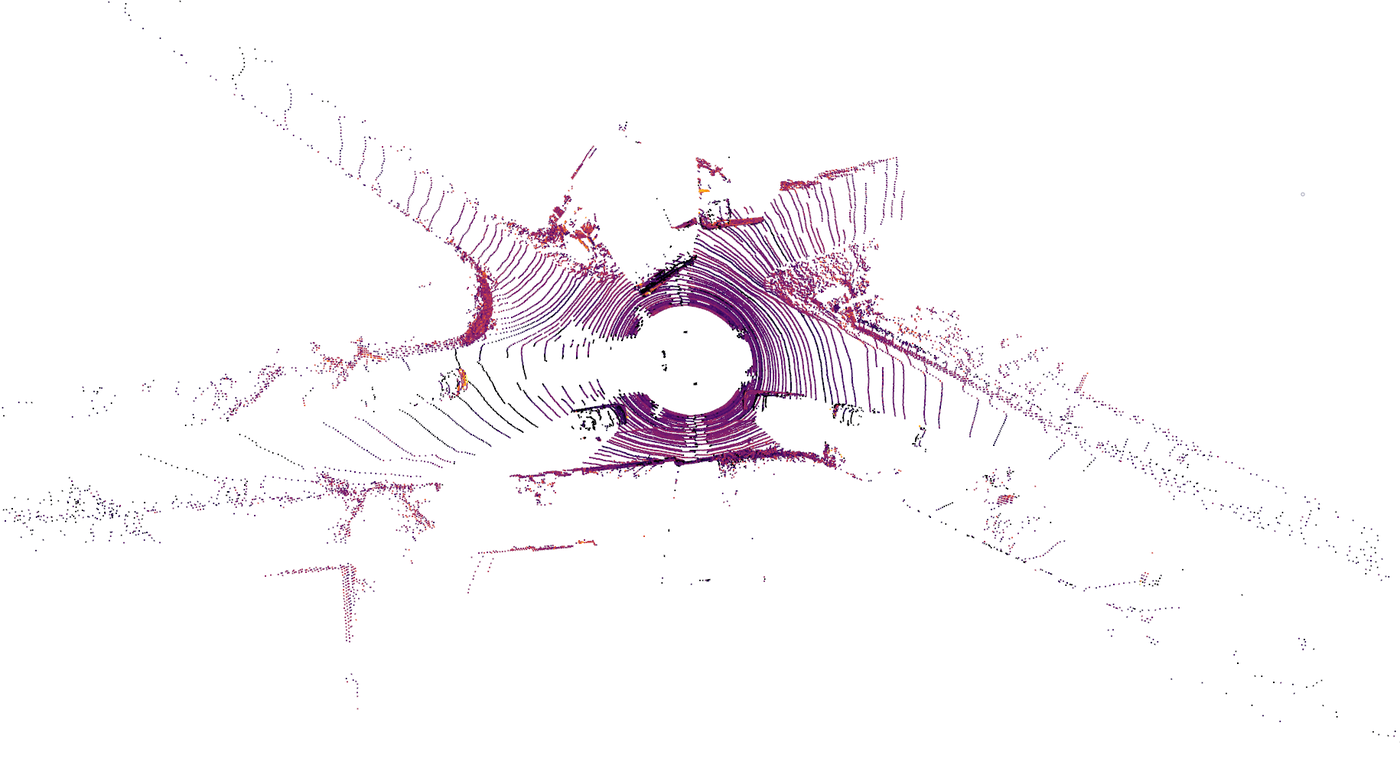} & 
\adjincludegraphics[width=0.32\textwidth, trim={{.2\width} {.3\height} {.2\width} {.3\height}},clip]{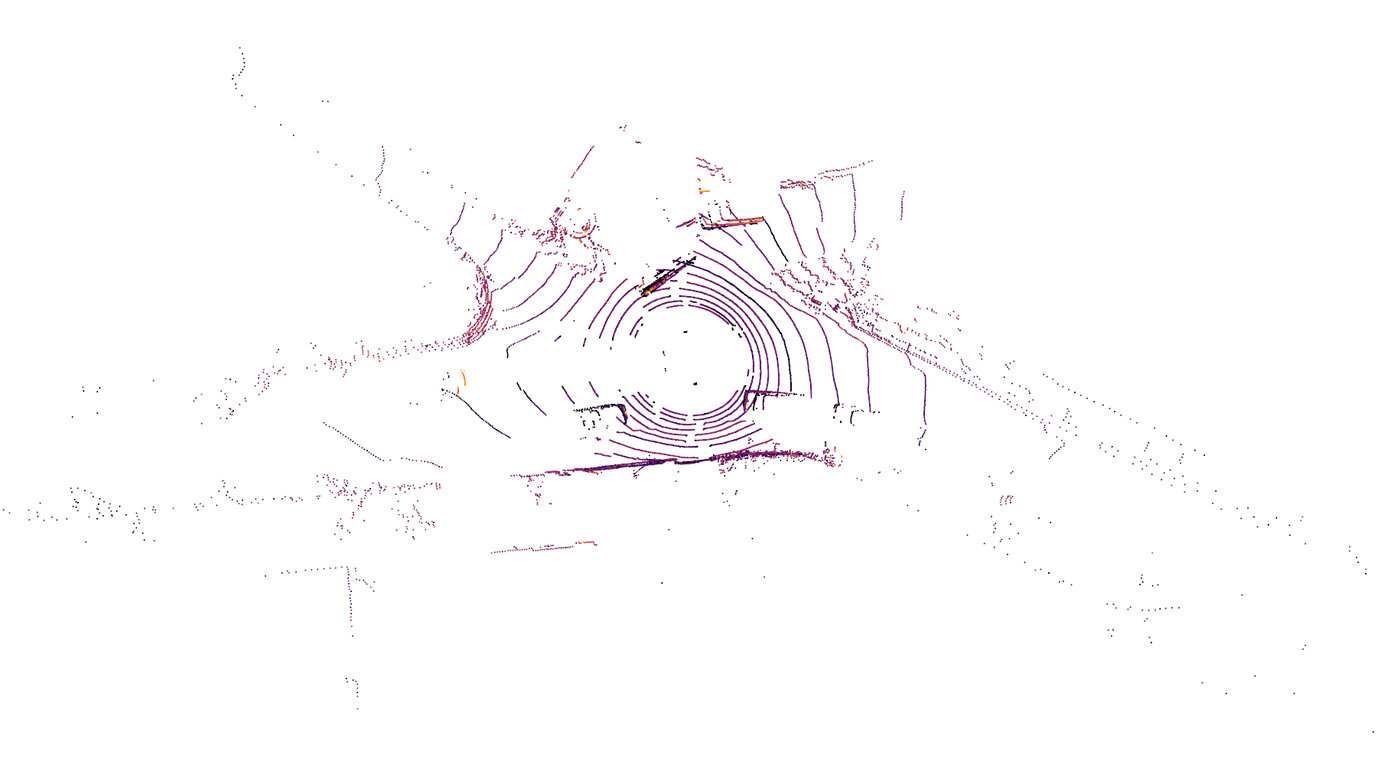} & 
\adjincludegraphics[width=0.32\textwidth, trim={{.2\width} {.3\height} {.2\width} {.3\height}},clip]{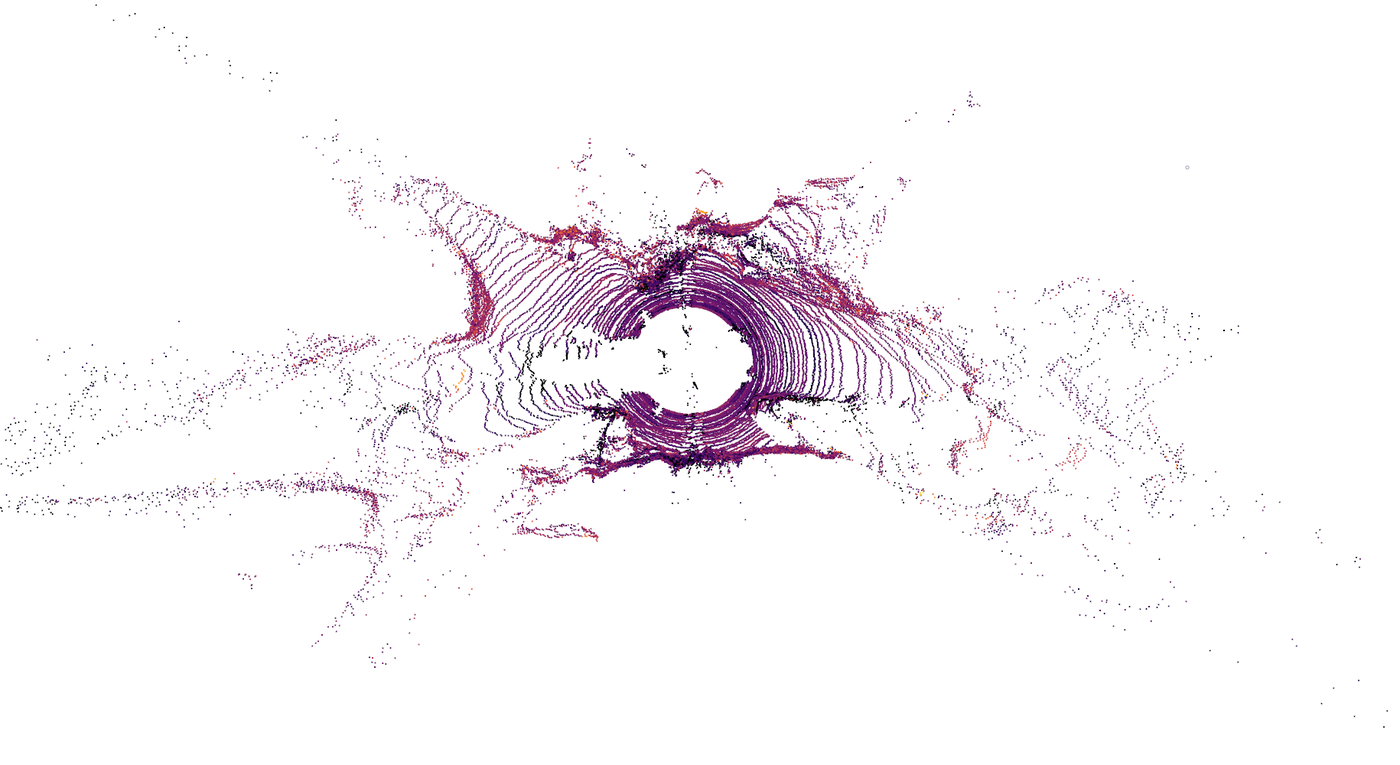} \\
\adjincludegraphics[width=0.32\textwidth, trim={{.2\width} {.3\height} {.2\width} {.3\height}},clip]{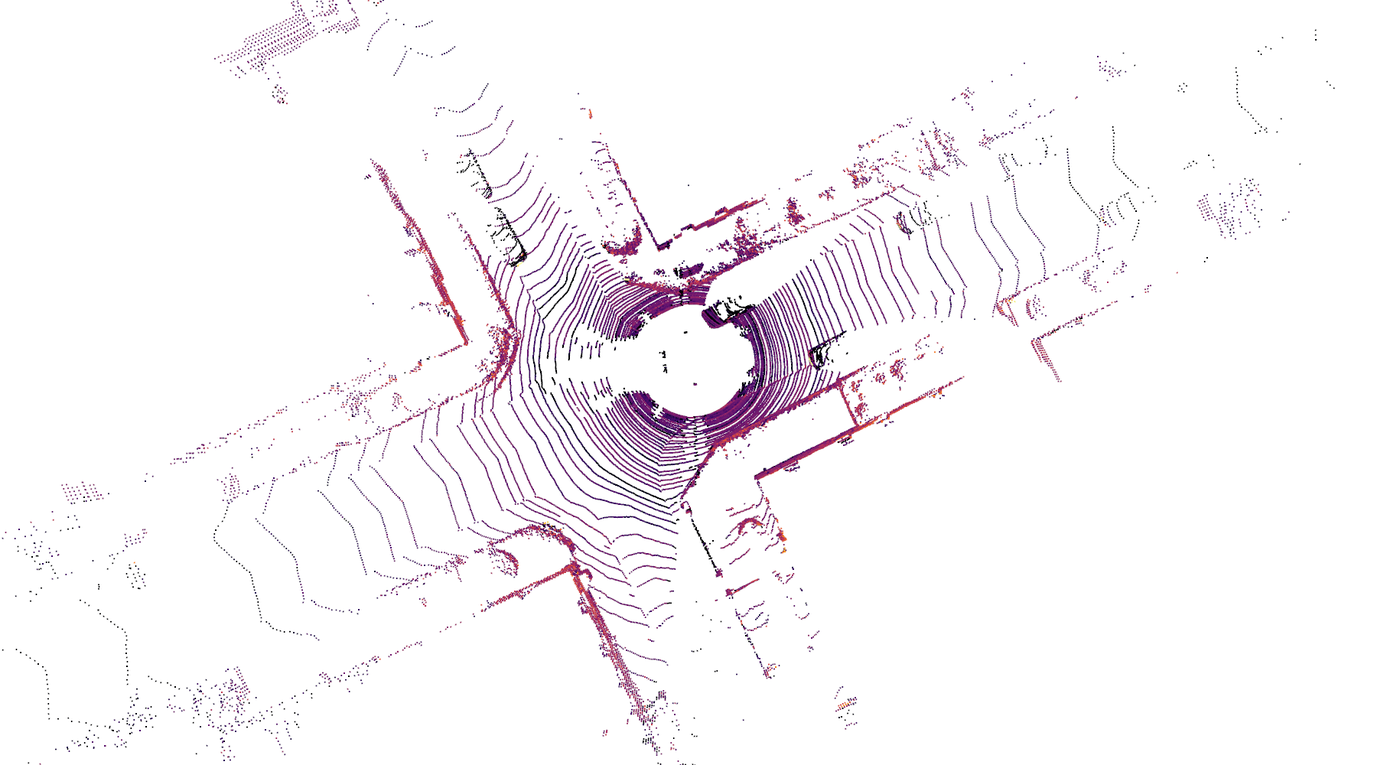} & 
\adjincludegraphics[width=0.32\textwidth, trim={{.2\width} {.3\height} {.2\width} {.3\height}},clip]{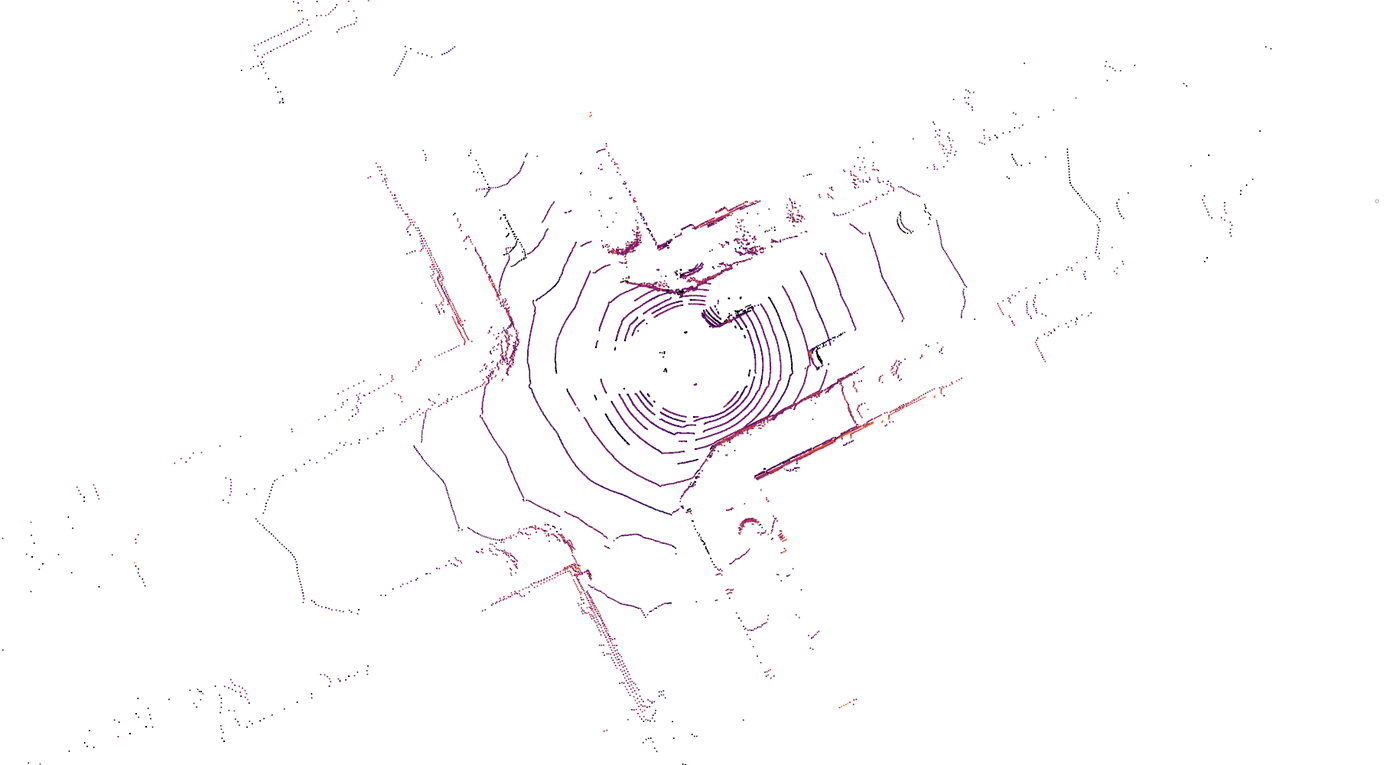} & 
\adjincludegraphics[width=0.32\textwidth, trim={{.2\width} {.3\height} {.2\width} {.3\height}},clip]{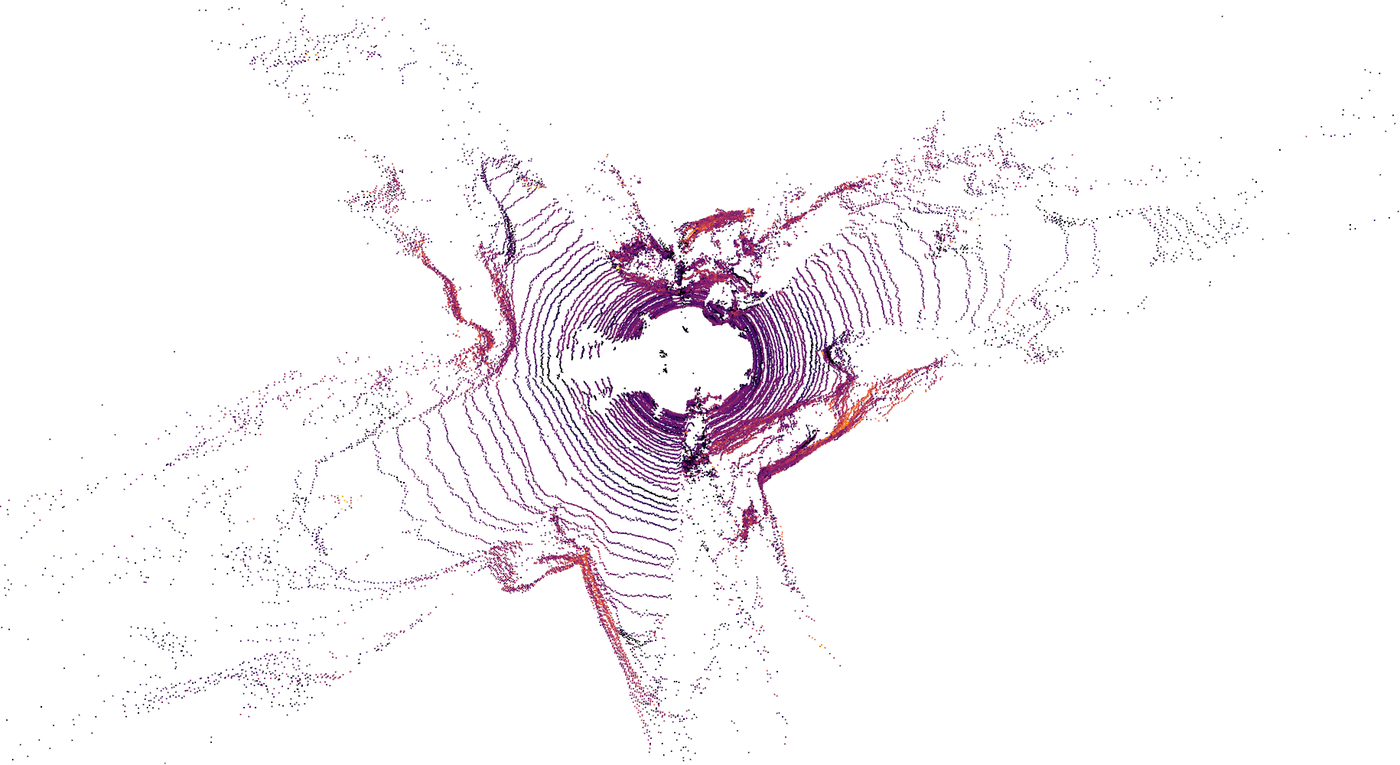} \\
Reference & 4-Beam Input & \textbf{Ours}\\
\adjincludegraphics[width=0.32\textwidth, trim={{.2\width} {.3\height} {.2\width} {.3\height}},clip]{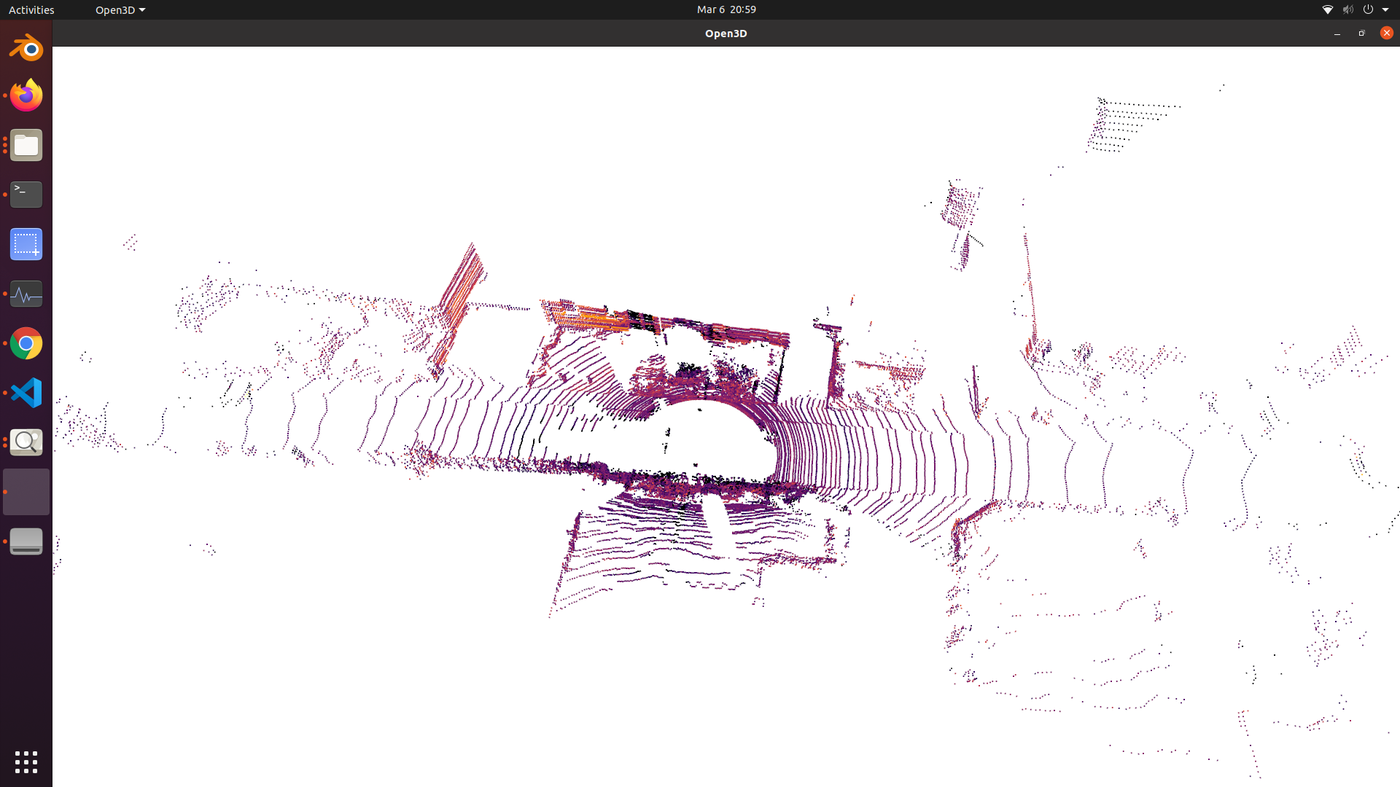} & 
\adjincludegraphics[width=0.32\textwidth, trim={{.2\width} {.3\height} {.2\width} {.3\height}},clip]{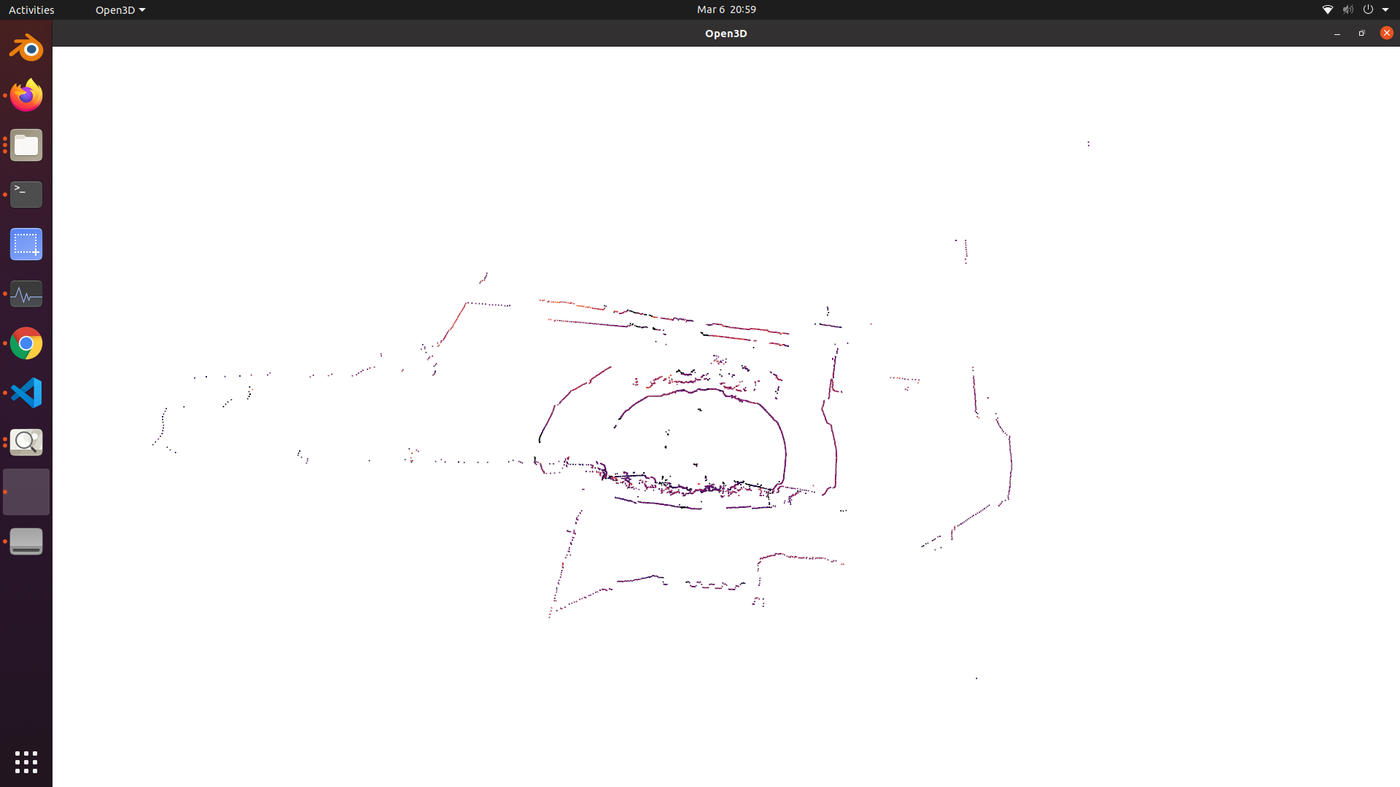} & 
\adjincludegraphics[width=0.32\textwidth, trim={{.2\width} {.3\height} {.2\width} {.3\height}},clip]{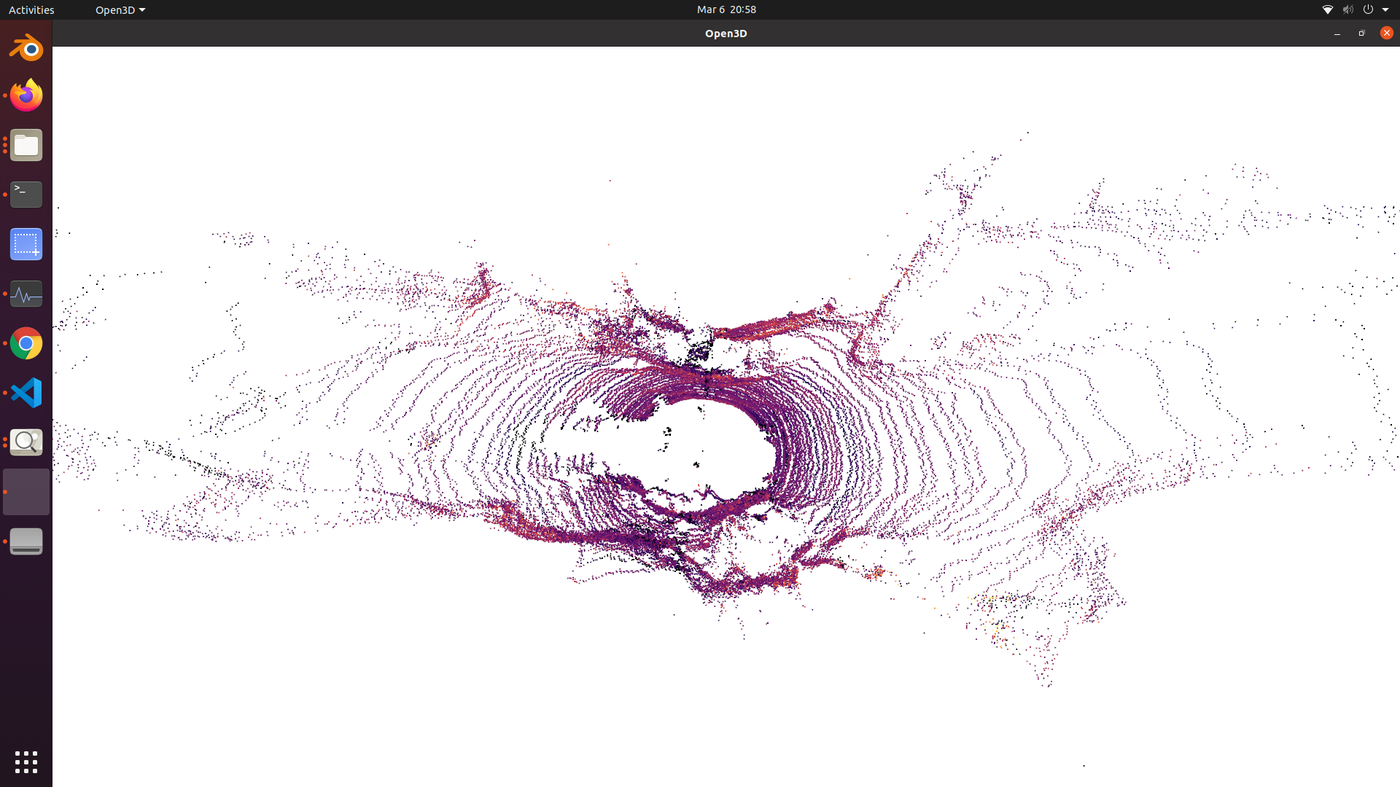} \\

\label{tab:densify}
\end{tabular}
\captionof{figure}{Qualitative Results for Unsupervised LiDAR Densification.}
\label{fig:densification}
\vspace{-25pt}
\end{table}

\section{Experiments}

\subsection{Experimental Setup}

\mypara{Datasets} We train and test our model's performance on the challenging KITTI-360 ~\cite{Kitti360} and \textbf{nuScenes} \cite{nuscenes2019} datasets. KITTI-360 contains 81,106 LiDAR readings from 9 long sequences around the suburbs of Karlsruhe, Germany. The scenes KITTI-360 covers are diverse, consisting of driving on highways and through residential and commercial districts. We split the dataset into two parts, where the first two sequences (30,758 frames) are the testing set, and the rest are used for training and cross-validation. nuScenes contains 297,737 LiDAR sweeps in the training set and 52,423 LiDAR sweeps in the testing and cross-validation set. The LiDAR sweeps were collected in the cities of Boston and Singapore. The two datasets provide different sensors (64 and 32 beams), geographic regions~(EU and NA), and content (suburbs and cities). 

\mypara{Metrics} 
Quantitatively measuring generative models is known to be difficult. 
In our work, we leverage three different metrics for evaluation. \textbf{Maximum Mean Discrepancy (MMD)} is a non-parameteric distance between two sets of samples. It compares the distance between two sets of samples by measuring the mean squared difference of the statistics of the two. MMD could be measured through the kernel trick: \[\textrm{MMD} = \frac{1}{N^2}\sum_i^N \sum_{i^\prime}^N k(\bx_i, \bx_{i^\prime})
- \frac{2}{NM}\sum_i^N \sum_{j}^M k(\bx_i, \bx_{j})
+ \frac{1}{M^2}\sum_j^M \sum_{j^\prime}^N k(\bx_j, \bx_{j^\prime}).\] For each point cloud we compute a $50\times 50$ spatial histogram along the ground plane (x and y coordinates) then use a Gaussian kernel to measure the similarity between the two.  
Additionally, inspired by the FID score for image generation, we evaluate a new \textbf{Frechet Range Distance} (FRD score) on KITTI-360. It evaluates the squared Wasserstein metric between mean and the covariance of a LiDAR perception network's activations from the synthetic samples and true samples. We choose RangeNet++, which is a encoder-decoder based network for segmentation pretrained on KITTI-360. To trade-off between quality and and preserve locality, we randomly choose 4,096 activation from the feature map of its bottleneck layer to fit the Gaussian distribution. 
Finally, we report the \textbf{Jensen–Shannon divergence} (JSD) between the empirical distributions of two sets of point clouds. We approximate the distribution through a birds-eye view 2D histogram at a resolution $100 \times 100$ for both reference sets and generated sets.

\mypara{Baselines} We have 3 baseline comparisons. The first baseline is the range-view based VAE-based LiDAR generation model proposed by Caccia et al.~\cite{caccia}. We have added additional layers and increased the generated range image size to $1024\times64$. We train the models for 165 epochs until convergence. The second baseline is the GAN-based model from Caccia et al.~\cite{caccia}. The GAN was pre-trained at a resolution of $256\times40$ following the original paper\footnote{we did not manage to make training converge in higher resolution}, followed by a upsampling layer to $1024\times40$. The last baseline is Projected GAN~\cite{ProjectedGAN}, one of the state-of-art GAN models for image generation. We adapt ProjectedGAN into our setting and train it for 3,000 epochs. All the generated range image samples are converted to a 3D point cloud in Cartesian coordinates for quality comparison.   

\mypara{Implementation Details} We use a UNet-like model for the score function. It takes in a 64x1024x2 (KITTI) or 32x1024x2 (Nuscenes) tensor as input and outputs the same size, denoting the gradient of log-prob. The U-Net comprises a stack of 6 down-sampling and a stack of 6 upsampling blocks, with skip connections in between. Each block has two convs. Each conv is preceded by InstanceNorm++ and an ELU activation. Number of channels is 32-64-64-64-128-128-128-128-64-64-64-32. Our model was trained with Adam optimizer with a learning rate of 1e-4. For sampling, we use a gradient update step of 2e-6 and 5 iterations per noise level. The initial $\sigma_0$ is 50, the final $\sigma_L$ is 0.01, and the number of levels is 232. To train Caccia et al.'s~\cite{caccia} models, we used a learning rate of 1e-4. All the models are trained and tested with an Nvidia RTX A4000 GPU.

\begin{table}[!t]
\scriptsize
\caption{Quantitative Results on KITTI-360~\cite{Kitti360}.\textbf{Bold} is Best; {\color{blue}Blue} is Second. 
}
\scriptsize
\centering
\begin{tabular}{rrrcr}
\toprule
& MMD$_\textrm{BEV}$ $\downarrow$ & FID$_\textrm{range}$ $\downarrow$ & JSD$_\textrm{BEV}$ $\downarrow$ \\
\midrule
LiDAR GAN \cite{caccia} & $3.06 \times 10^{-3}$    & $3003.8$ & $-$ \\
LiDAR VAE \cite{caccia} & $1.00 \times 10^{-3}$    & $2261.5$  & $0.161$\\ 
Projected GAN \cite{ProjectedGAN} & \boldmath$3.47 \times 10^{-4}$   & {\color{blue} \boldmath$2117.2$} & {\color{blue} \boldmath$0.085$} \\
Ours & {\color{blue} \boldmath$3.87 \times 10^{-4}$}    & \boldmath$2040.1$ & \boldmath$0.067$ \\
\bottomrule
\label{tab:kitti}
\end{tabular}
\label{tab:quant}
\vspace{-20pt}
\end{table}

\subsection{KITTI-360 Evaluation}

\mypara{Quantitative Results}
Tab.~\ref{tab:kitti} shows quantitative results among all the competing algorithms. From the table we could see that our method produces superior performance on the FRD score compared against other methods. In terms of MMD our approach is also ranking high. It is slightly lower than projected GAN, however both match the histogram well with an MMD score smaller than 1e-4. As we mention in metric subsection, every metric is a partial evaluation of the sampling quality, and urge the readers to consider all quantitative metrics, the human study, and the qualitative results as a holistic evaluation.

\mypara{Qualitative Results}
Fig.~\ref{fig:qual} demonstrates some randomly selected samples from all the competing algorithms. We also list the true point cloud samples from KITTI as a reference. From the figure, we could see our approach produces significantly higher quality samples than the competing algorithms. Specifically, LiDARGAN captures the overall layout, but fails in producing high-detailed structures, such as cars, trees, sidewalks, pedestrians, etc. Projected GAN generates reasonable, detailed structures at near range, but brings significant artifacts at far range. Ours excel in terms of both realism in layout and geometry details, as well as diversity in content. Additionally, we provide a zoom-in visualization of our 3D point cloud in Fig.~\ref{fig:details}, highlighting the high quality geometric details our method could offer.

\begin{table}[!t]
\scriptsize
\caption{Human Study Results on KITTI-360}
\small
\centering
\begin{tabular}{lc}
\toprule
Method  & Percent Prefer Ours \\
\midrule
Ours vs. VAE  & 97\% \\
Ours vs. GAN  & 96.6\% \\
Ours vs. ProjectedGAN & 100\%\\
\bottomrule
\label{tab:humanstudy}
\end{tabular}
\vspace{-25pt}
\end{table}

\mypara{Human Study}
To evaluate the perceptual quality, we perform an A/B test on a team of students. Our test system shows a pair of randomly chosen images of two point cloud sampled from two different methods. Human judges then choose which one is more realistic. Participates also have access to real KITTI point clouds for reference. The raw results are shown in Tab.~\ref{tab:humanstudy}. In total, 5 participants labeled 600 image pairs. At a confidence level of $99\%$ the two-sided test p-value is smaller than 1e-4, demonstrating statistical significance.

\subsection{NuScenes Results} Fig.~\ref{fig:nuscenes} depicts the qualitative comparison results. From this figure, we can see that our method still achieves superior results compared to both VAE~\cite{caccia} and projected GAN~\cite{ProjectedGAN}. An AB test on a group of four human subjects suggests that our method is significantly favored over other competing algorithms in $89\%$ of cases.  While achieving superior human study performance, we notice that our method tends to generate point clouds that concentrate their mass closer to the viewpoint. As a result, despite superior visual quality, our MMD score at BEV is worse than VAE and Projected GAN (2e-3 vs. 1.1e-3 and 6e-5). We will leave this shrinking effect for future investigation. 

\begin{table}[!t]
\scriptsize
\centering
\begin{tabular}{ccccc}
Ground-Truth & VAE~\cite{caccia} & ProjectedGAN~\cite{ProjectedGAN} & \textbf{Ours}\\

\adjincludegraphics[width=0.22\textwidth, trim={{.3\width} {.3\height} {.3\width} {.3\height}},clip]{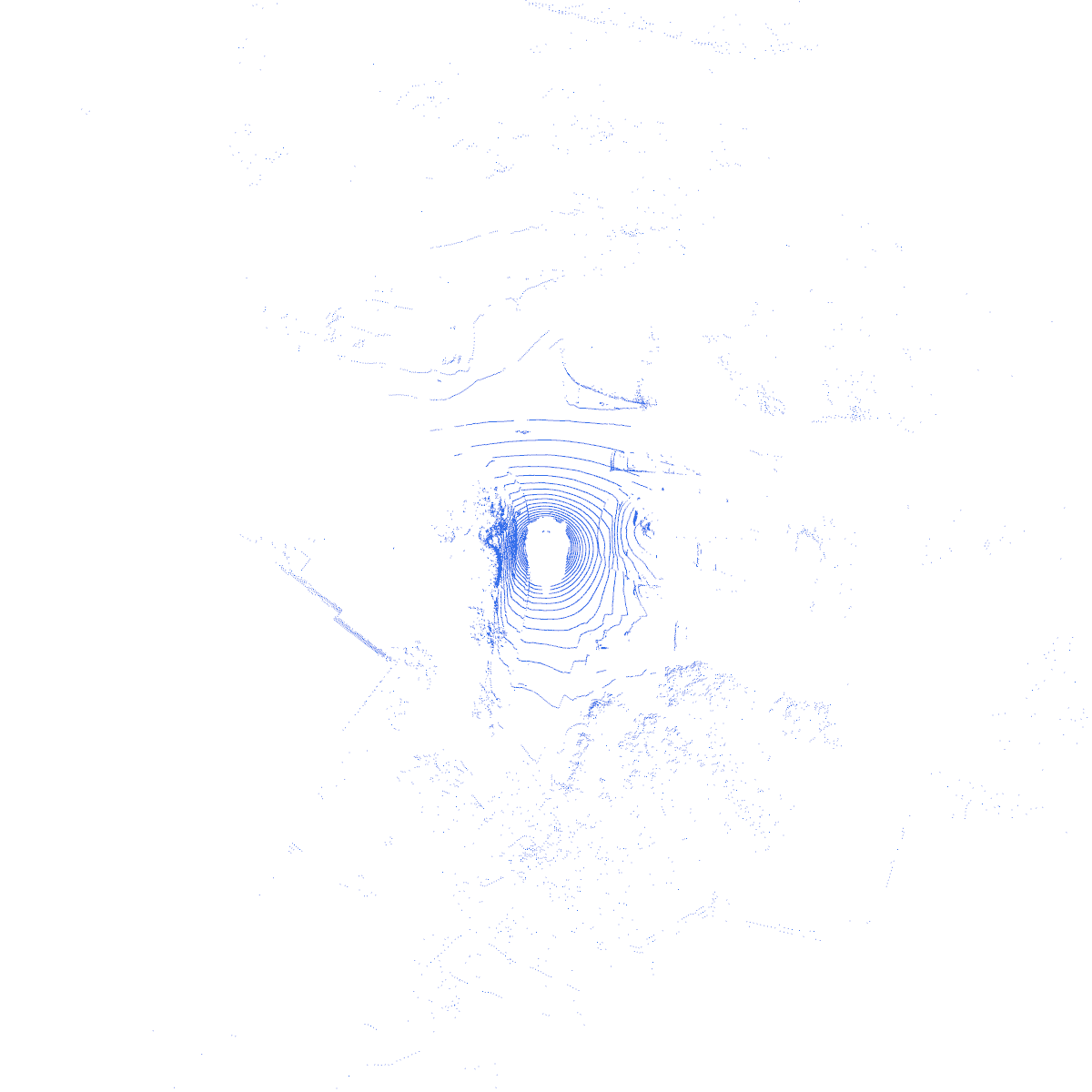} & 
\adjincludegraphics[width=0.22\textwidth, trim={{.3\width} {.3\height} {.3\width} {.3\height}},clip]{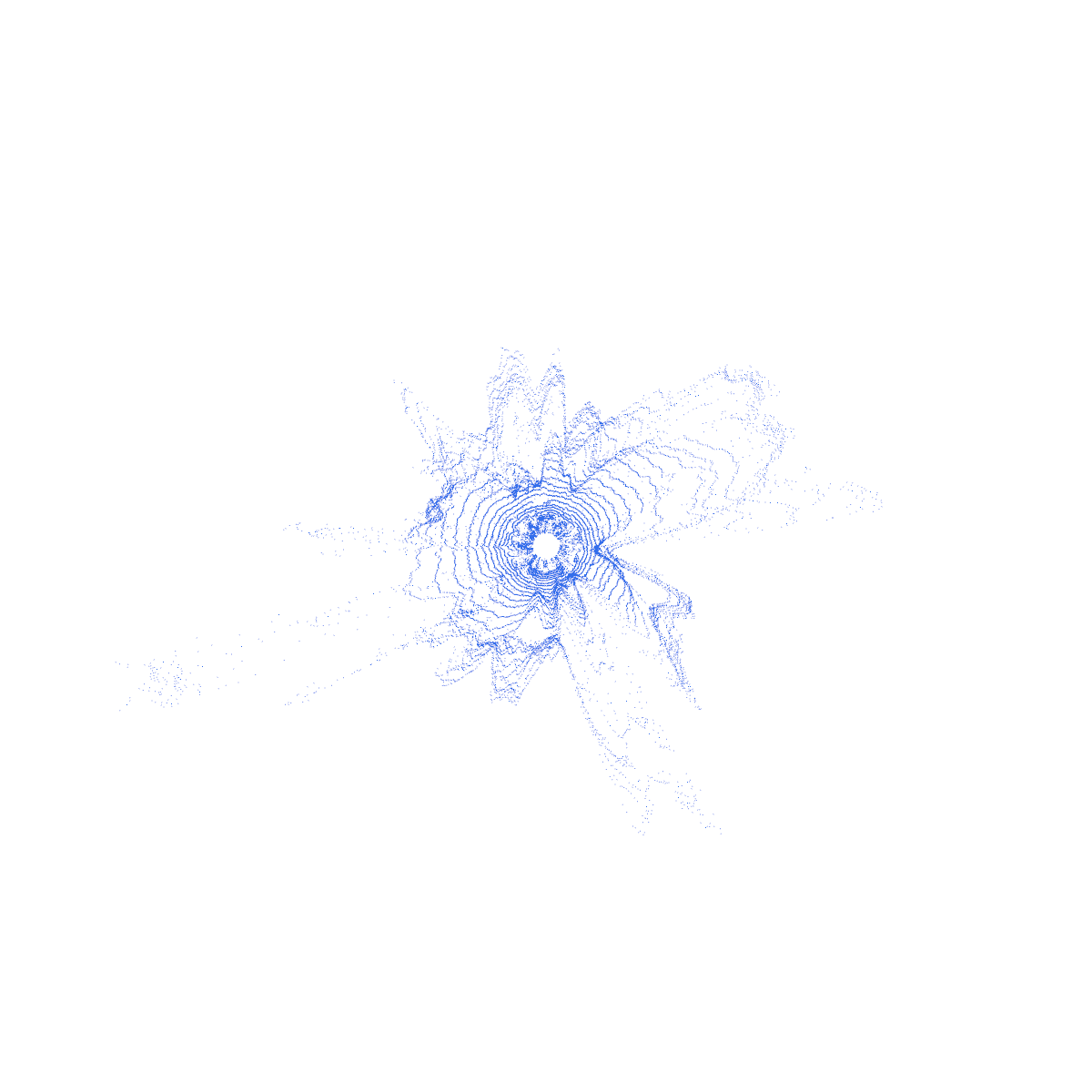} & 
\adjincludegraphics[width=0.22\textwidth, trim={{.3\width} {.3\height} {.3\width} {.3\height}},clip]{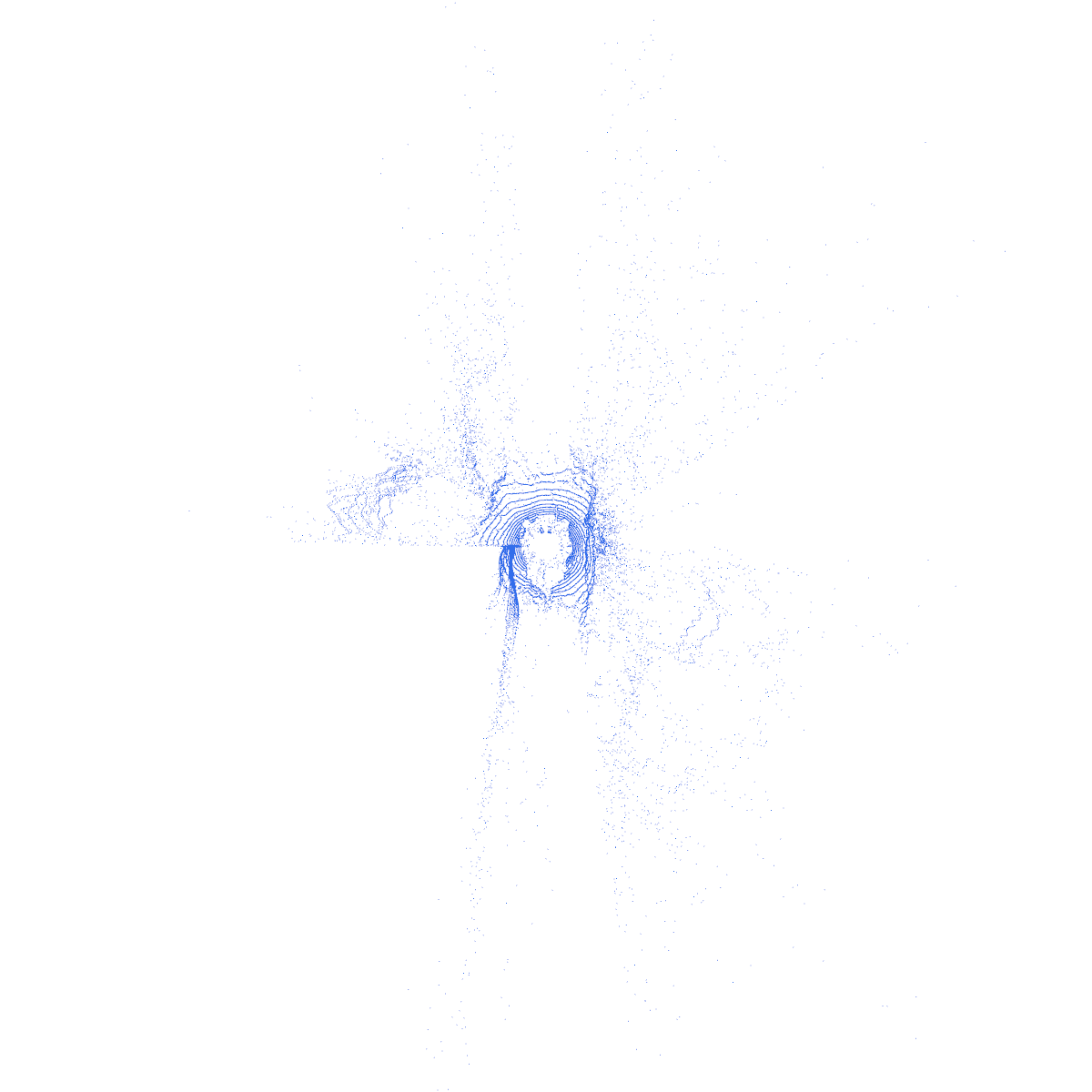}  & 
\adjincludegraphics[width=0.22\textwidth, trim={{.3\width} {.3\height} {.3\width} {.3\height}},clip]{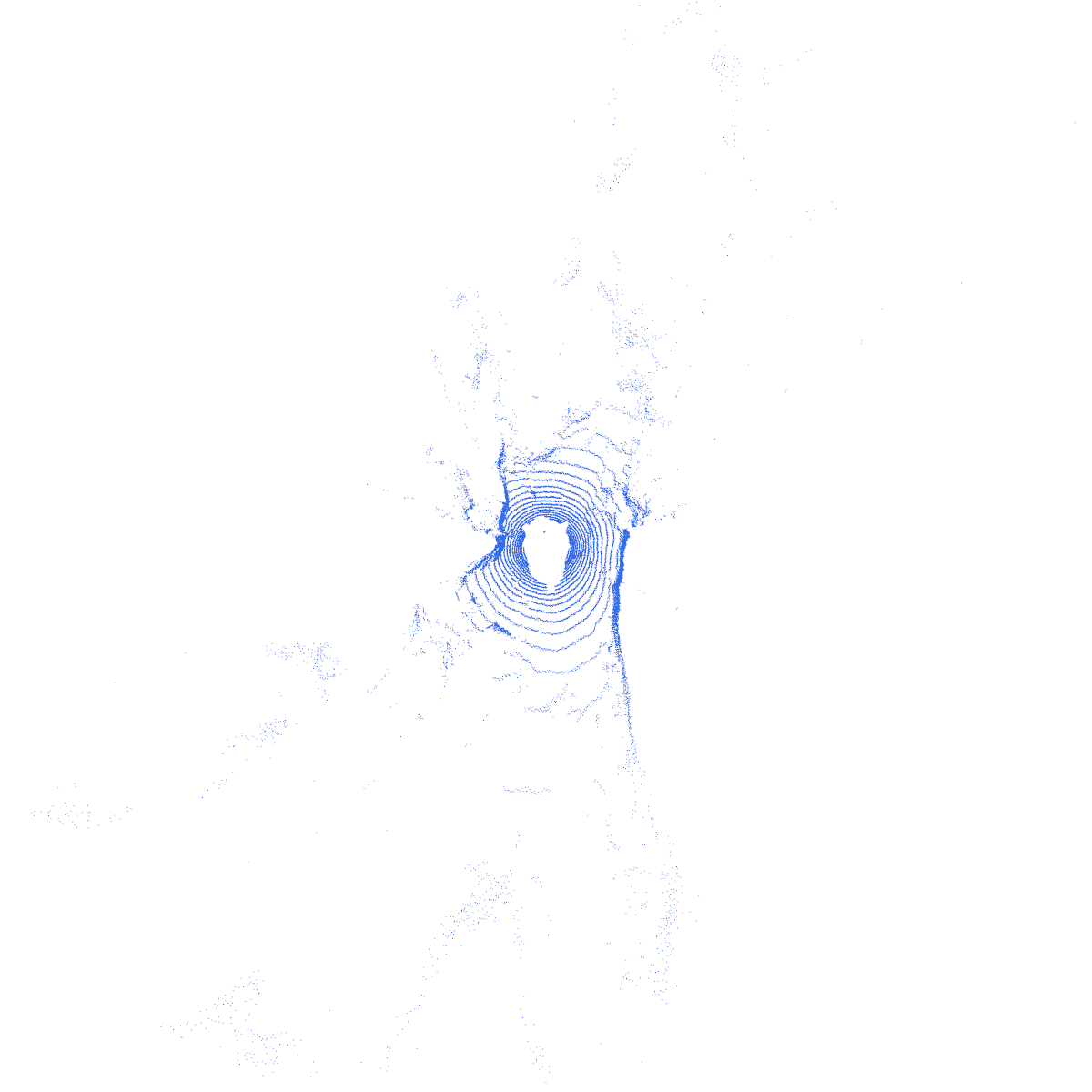}\\
\end{tabular}
\captionof{figure}{Qualitative Results on the nuScenes dataset. }
\label{fig:nuscenes}
\vspace{-8mm}
\end{table}

\subsection{Posterior Sampling}
We also evaluate our LIDAR generation model on the task of LiDAR densification. More specifically, we simulate low-beam LiDAR sensor readings as our sparse input by selecting a subset of the beams from the raw 64-beam sensors. In this example, we create 4-beam and 16-beam input as shown in Fig.~\ref{fig:densification}. Following the posterior sampling procedure described in Eq.~\ref{eq:posterior} and Eq.~\ref{eq:densification}. 
Fig.~\ref{fig:densification} depicts the sparse input, a dense ground-truth reference and our qualitative posterior sampling results. As shown in the figure, the resulting point cloud is realistic and reflects the input guidance. %

\begin{figure}[t!]
\vspace{-3mm}
\centering
\def\arraystretch{0.3}
\setlength{\tabcolsep}{0.5pt}
\resizebox{\linewidth}{!}{
\tiny
\begin{tabular}{ccccccc}
\includegraphics[width=0.11\linewidth]{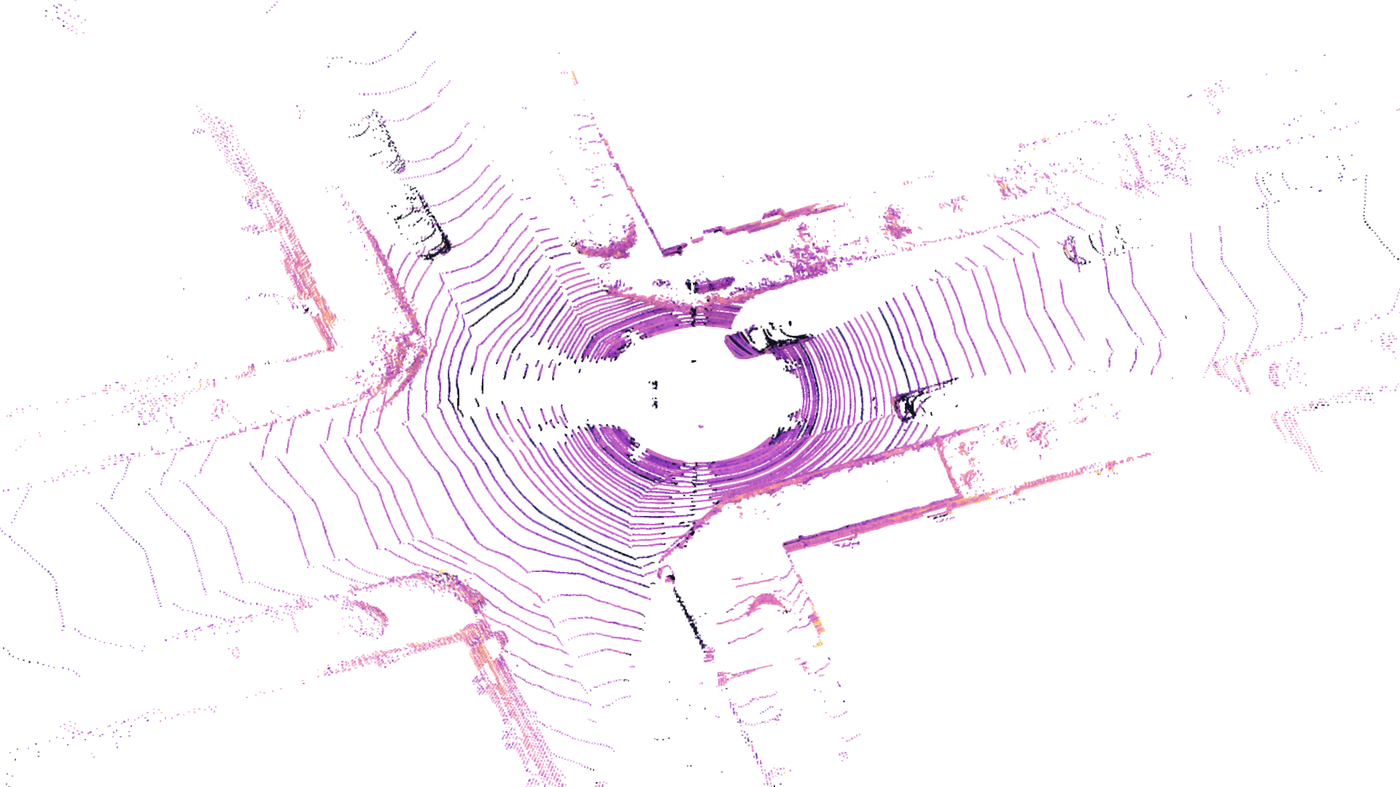}
&\includegraphics[width=0.11\linewidth]{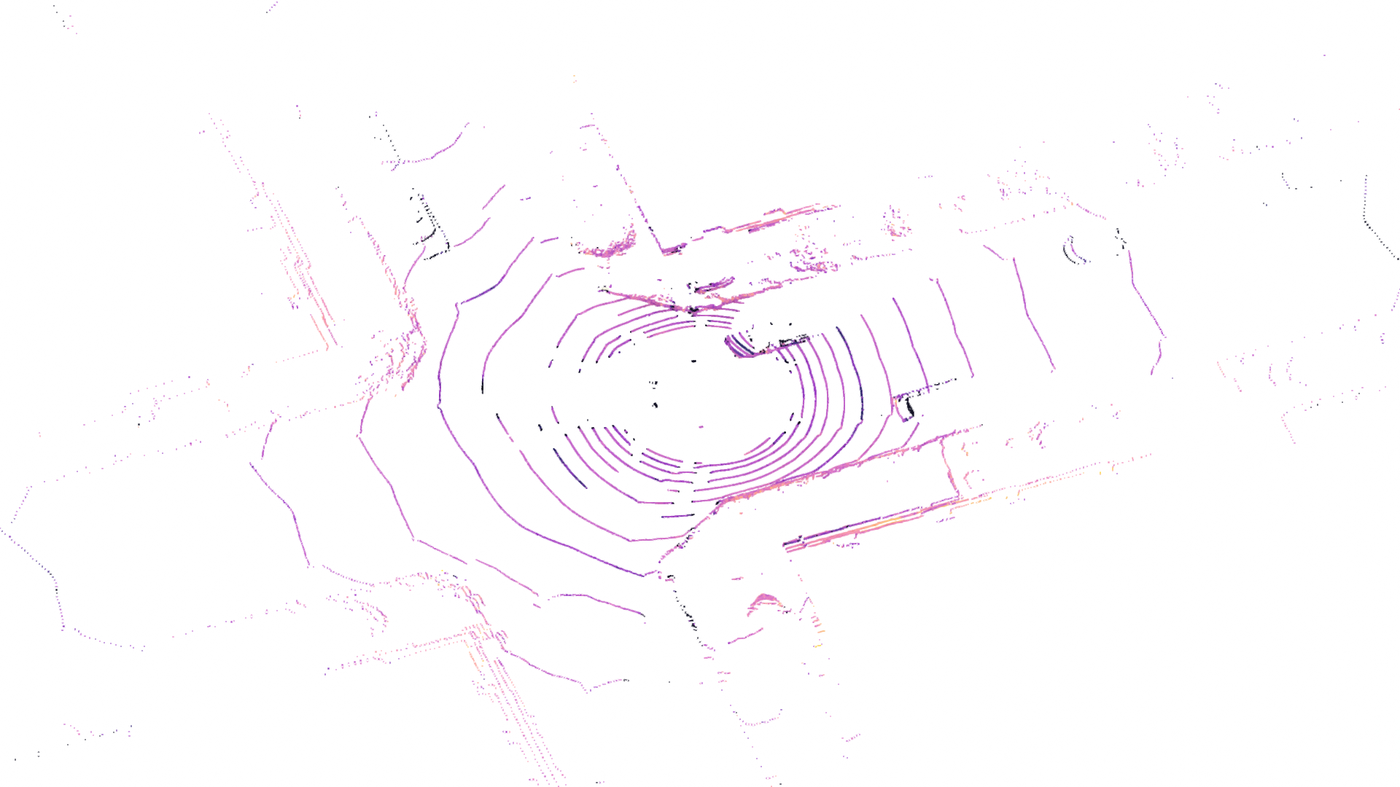}
&\includegraphics[width=0.11\linewidth]{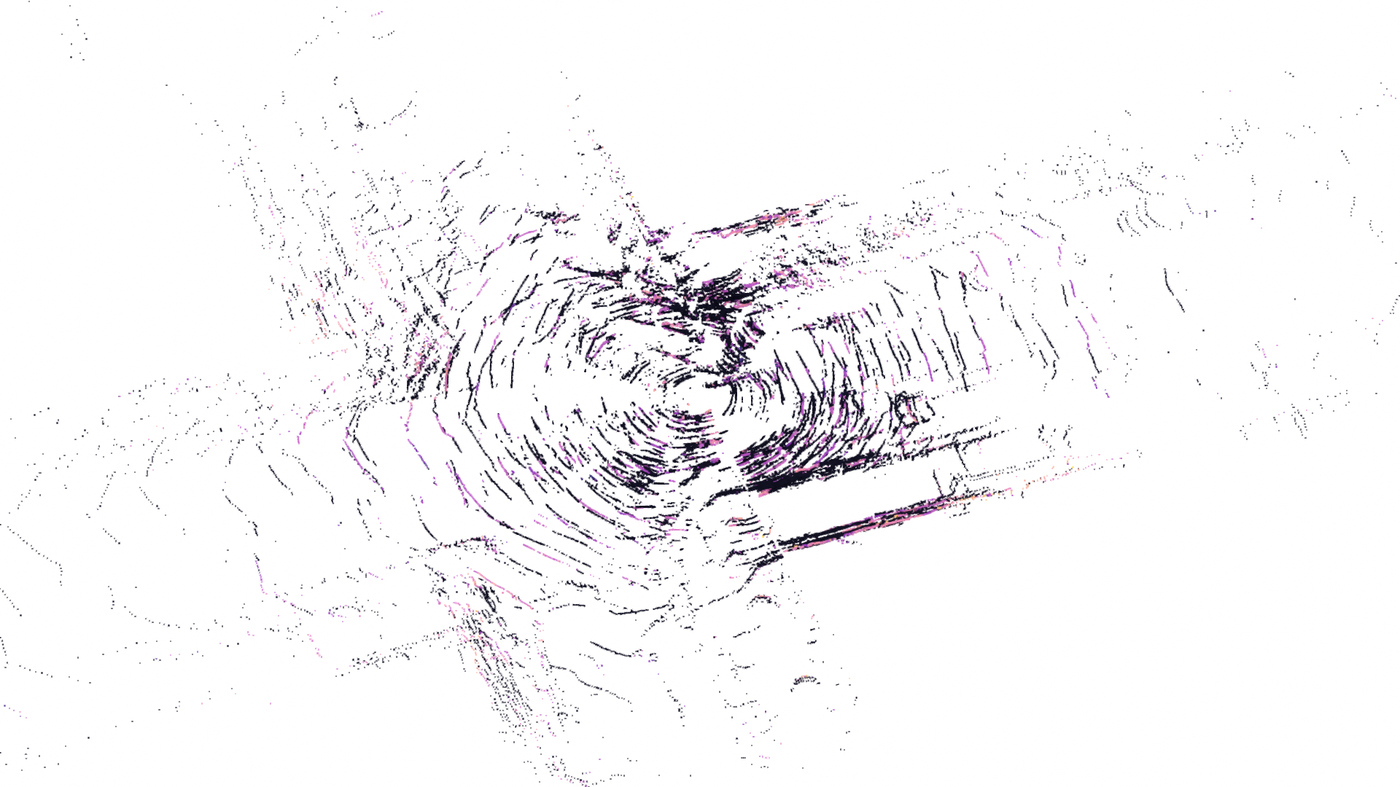}
&\includegraphics[width=0.11\linewidth]{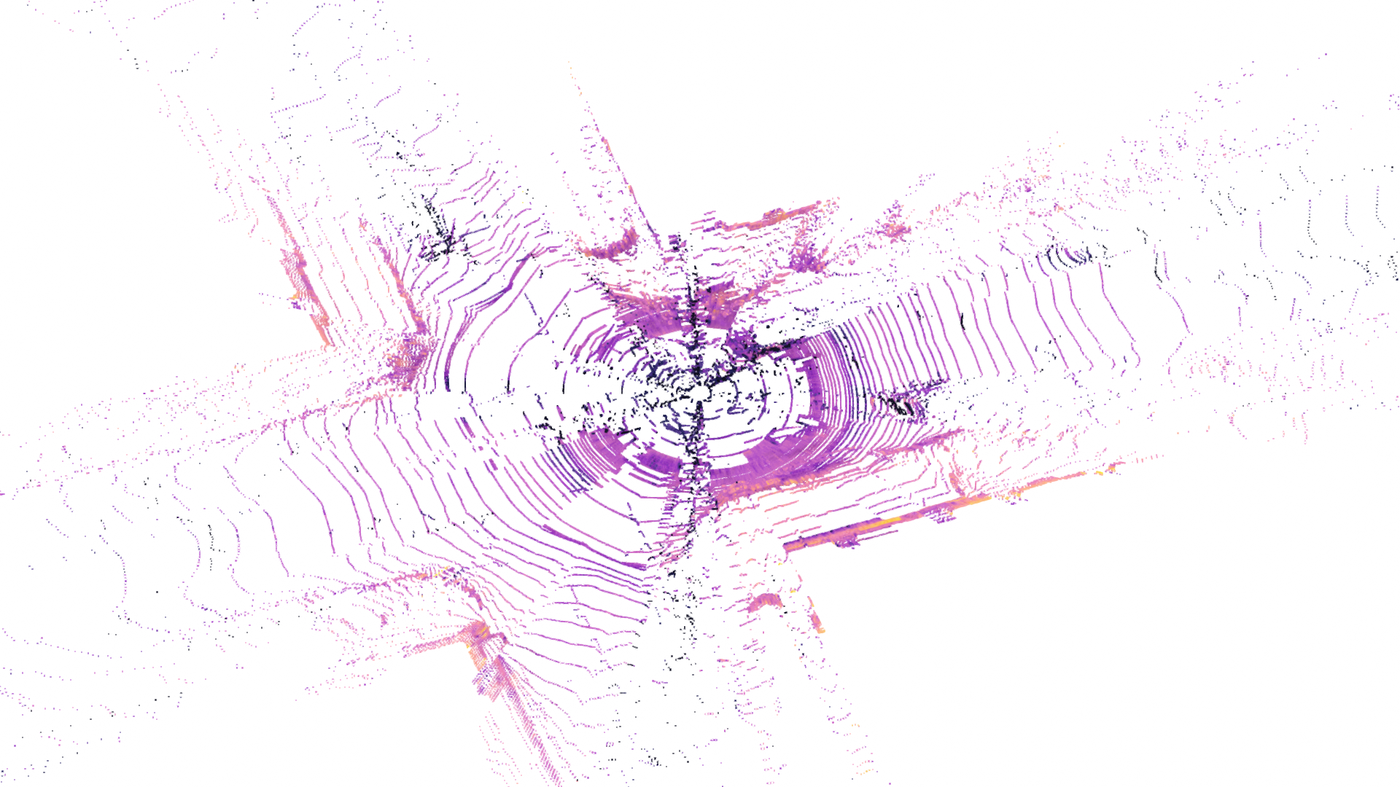}
&\includegraphics[width=0.11\linewidth]{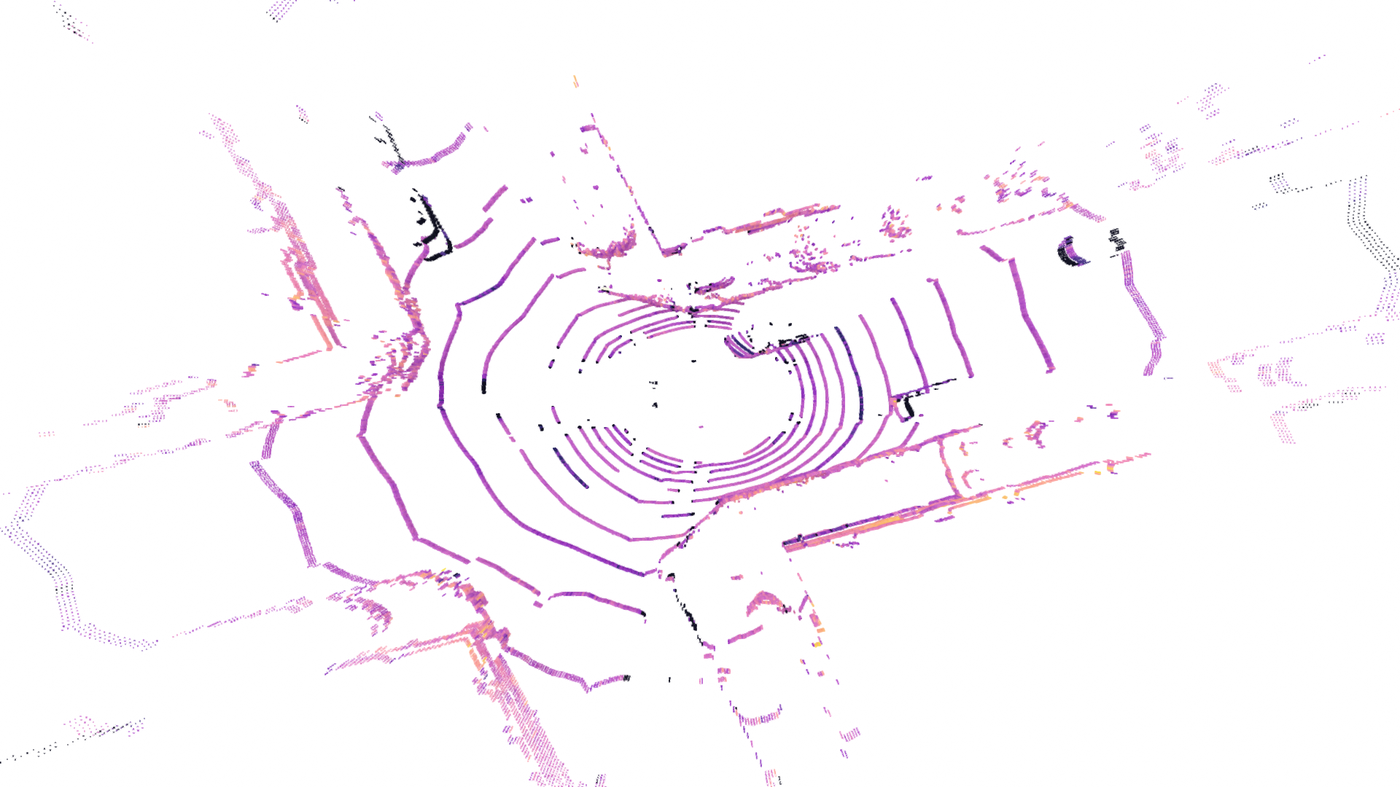}
&\includegraphics[width=0.11\linewidth]{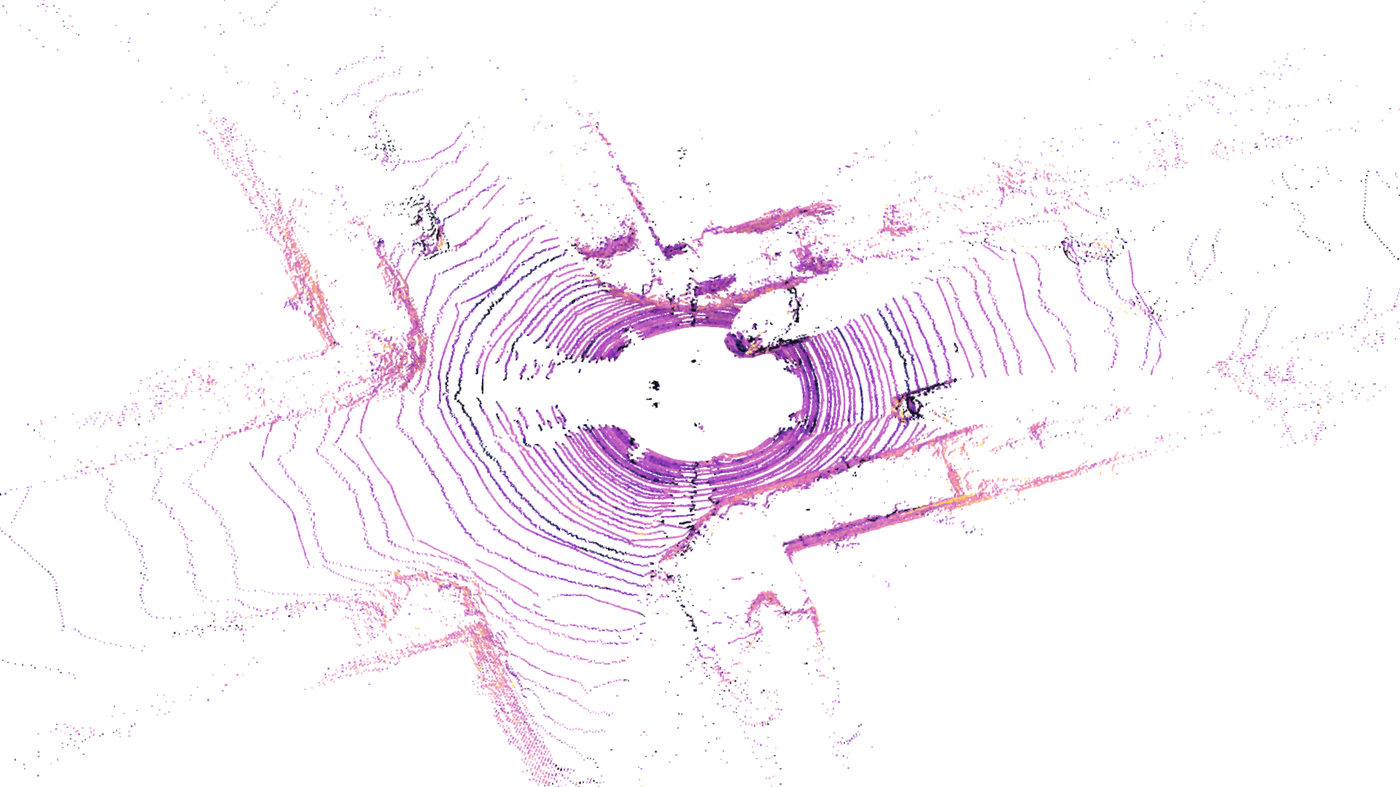}\\
GT & Input & PUNet & Bicubic & NN & Ours
\end{tabular}
}
\vspace{-3mm}
\caption{Densification Results.}
\label{fig:densification_comparison}
\vspace{-5mm}
\end{figure}

\mypara{Qualitative Comparison} We compare PUNet~\cite{PU-net}, bicubic interpolation, and nearest neighbor interpolation with ours on KITTI-360. Quantitative results are shown in Tab.~\ref{tab:densification_quant}. Qualitative results are shown in Fig.~\ref{fig:densification_comparison}. PUNet is B/W as it does not upsample intensity. Our results suggest that the proposed densification method is superior to both learned and interpolation approaches.  %
\begin{table}[!t]
\parbox{.45\linewidth}{
\centering
\caption{LiDAR Densification.}
\begin{tabular}{rrrcr}
\toprule
& MAE $\downarrow$  \\
\midrule
PUNet~\cite{PU-net} & $6.88$ \\
NN & $2.18$ \\
Ours & {$\mathbf{1.23}$} \\
\bottomrule
\label{tab:densification_quant}
\end{tabular}
}
\hfill
\parbox{.54\linewidth}{
\centering
\caption{Ablation Study}
\begin{tabular}{cccc}
\toprule
Coord-aware  &  CircConv & FRD & MMD \\
\midrule
No & No & 2422.3 &  $7.60 \times 10^{-4}$ \\
Yes & No & 2251.1 & $3.94 \times 10^{-4}$ \\
Yes & Yes & \textbf{2040.1} & $\mathbf{3.87 \times 10^{-4}}$   \\
\bottomrule
\label{tab:ablation}
\end{tabular}
}
\vspace{-8mm}
\end{table}

\begin{figure}[t]
\vspace{-3mm}
\centering
\def\arraystretch{0.3}
\setlength{\tabcolsep}{0.5pt}
\resizebox{\linewidth}{!}{
\footnotesize
\begin{tabular}{ccc}
\includegraphics[width=0.28\linewidth]{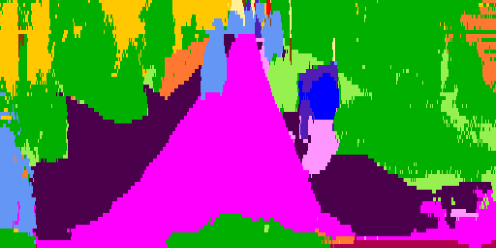}
&\includegraphics[width=0.28\linewidth]{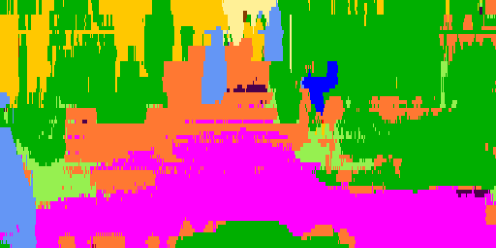}
&\includegraphics[width=0.28\linewidth]{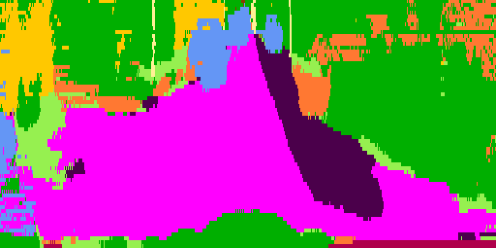}\\
True Reference & Nearest-Neighbor & Ours
\end{tabular}
}
\vspace{-3mm}
\caption{RangeNet++ segmentation on densified LiDAR.}
\label{fig:segmentation}
\vspace{-3mm}
\end{figure}

\mypara{Downstream Applications} We run RangeNet++ semantic seg on densified point cloud without fine-tuning.  
(Fig.~\ref{fig:segmentation}).
Applying LiDARGen to densify a sparse (16 Beam) LiDAR helps RangeNet++ create cleaner results (e.g., the road) and recover lost details (e.g., the cars in the distance). Our method also achieves better quantitative results compared to nearest-neighbour up-sampling. Per-point accuracy is 0.546 (NN) and 0.608 (ours). IOU is 0.394 (NN) and 0.449 (ours). 

\subsection{Discussions}

\mypara{Ablation Studies}
We conduct ablation studies to justify the design choice of our algorithms. We compare the same score function model in three different settings (w/o circular conv and w/o coordinate-encoding). As shown in Tab.~\ref{tab:ablation}, both help improve the performance in terms of FRD and MMD. For more information and qualitative comparison, please refer to the supplementary material.

\mypara{Limitations}
Despite producing superior performance and flexibility, LiDARGen still has several limitations. First, sampling efficiency is one of the major drawback -- LiDARGen takes approximately 12min to sample 36 LiDAR samples in a batch. We leave it as future work and anticipate that leveraging recent acceleration techniques for diffusion-based models is a promising direction to alleviate this issue. In addition, our approach cannot yet pass the Turing test for experienced LiDAR perception researchers. There are a few artifacts: our samples have degraded geometric details at far range and tended to have less straight walls than real samples. We plan to explore multi-modal networks (e.g. hybrid equirectangular view and bird's eye view) in the future.

\section{Conclusion}

We propose LiDARGen, a score-based approach for LiDAR point cloud generation. Our method samples a realistic point cloud by progressively denoising a noisy input. We demonstrate that our unconditional generation model could be directly applied for various conditional generation tasks through posterior sampling. A human study and perceptional similarity evaluation on the challenging KITTI-360 dataset validates the effectiveness of our method. We hope this approach will open up the research to provide easy access to realistic LiDAR sensory data directly from machine learning. We also expect to explore potential applications of LiDARGen in 3D environment generation and self-driving. 

\section{Acknowledgement}
The authors thank Wei-Chiu Ma and Zhijian Liu for their feedback on early drafts and all the participants in the human perceptual quality study. The project is partially funded by the Illinois Smart Transportation Initiative STII-21-07. We also thank Nvidia for the Academic Hardware Grant.

\clearpage
\bibliographystyle{splncs04}
\bibliography{main}
\end{document}


\pagestyle{headings}
\mainmatter
\def\ECCVSubNumber{2118}  %

\title{Learning to Generate Realistic LiDAR Point Clouds -- Supplementary Material } %

\titlerunning{Lidar Generation}

\author{Vlas Zyrianov \and Xiyue Zhu \and Shenlong Wang}
\authorrunning{Zyrianov et al.}
\institute{University of Illinois at Urbana-Champaign, IL, USA
\email{\{vlasz2,xiyuez2,shenlong\}@illinois.edu}}

\maketitle
\begin{abstract}
In the supplementary material, we first provide more qualitative and quantitative results and additional ablation analysis in Sec.~\ref{sec:analysis}. In addition, we report experimental results on the challenging NuScenes dataset in Sec.~\ref{sec:nuscenes}. 
Finally, we provide additional quantitative and qualitative results for posterior sampling in Sec.~\ref{sec:posterior}. The supplementary video “\textit{LiDARGen-intro.mp4}” briefly introduces our method, demonstrates the diffusion process in detail on KITTI-360, and compares qualitative results with other methods.
\keywords{3D generation, self-driving, generative models}
\end{abstract}

\section{Additional Analysis}
\label{sec:analysis}

\mypara{Qualitative Ablation Study} We conduct ablation studies to justify the design choice of our algorithm. We compare the same score function model in three different settings: with circular convolution and without coordinate encoding, with circular convolution and without coordinate encoding, and our final model, which uses circular convolution and a coordinate encoding. Qualitative results are shown in Fig.~\ref{fig:ablation}. 

Without \textbf{circular convolution}, a discontinuity appears in the point cloud representation horizontally, starting from the origin. This discontinuity is most clearly seen in the sixth row of the second column in Fig. \ref{fig:ablation}. This discontinuity is caused by the left and right edges of the range image lacking a receptive field between each other when using normal convolutions. To address this issue, we use Circular Convolutions. Qualitatively, this discontinuity is fixed with this change. 

With the help of \textbf{coordinate encoding}, our approach generates more straight road layouts that appropriately reflect the real-world layout distribution in the urban driving environment. 

\mypara{Comparison to Point-based Backbone} We also compare our model against the point-based score-matching model proposed in ShapeGF~\cite{cai2020learning}. The original ShapeGF model was trained and tested in ShapeNet. We adapt their model on LiDAR generation by changing the point cloud size to be 50k points and changing the noise level schedule by setting the number of noise steps to be 15 and end noise sigma to be 0.001. We train its stage 1 autoencoding and stage 2 GAN model from scratch on the KITTI training set until the validation loss converges. As shown in Fig.~\ref{fig:shapegf}, ShapeGF cannot provide physically feasible results like equirectangular-based approaches. Further, despite working very well for ShapeNet-like objects, we find it has a strong mode collapse issue on complicated urban driving scenes.  

\begin{table}[!t]
\small
\centering
\begin{tabular}{ccc}
 \adjincludegraphics[width=0.33\textwidth, trim={{.2\width} {.3\height} {.2\width} {.3\height}},clip]{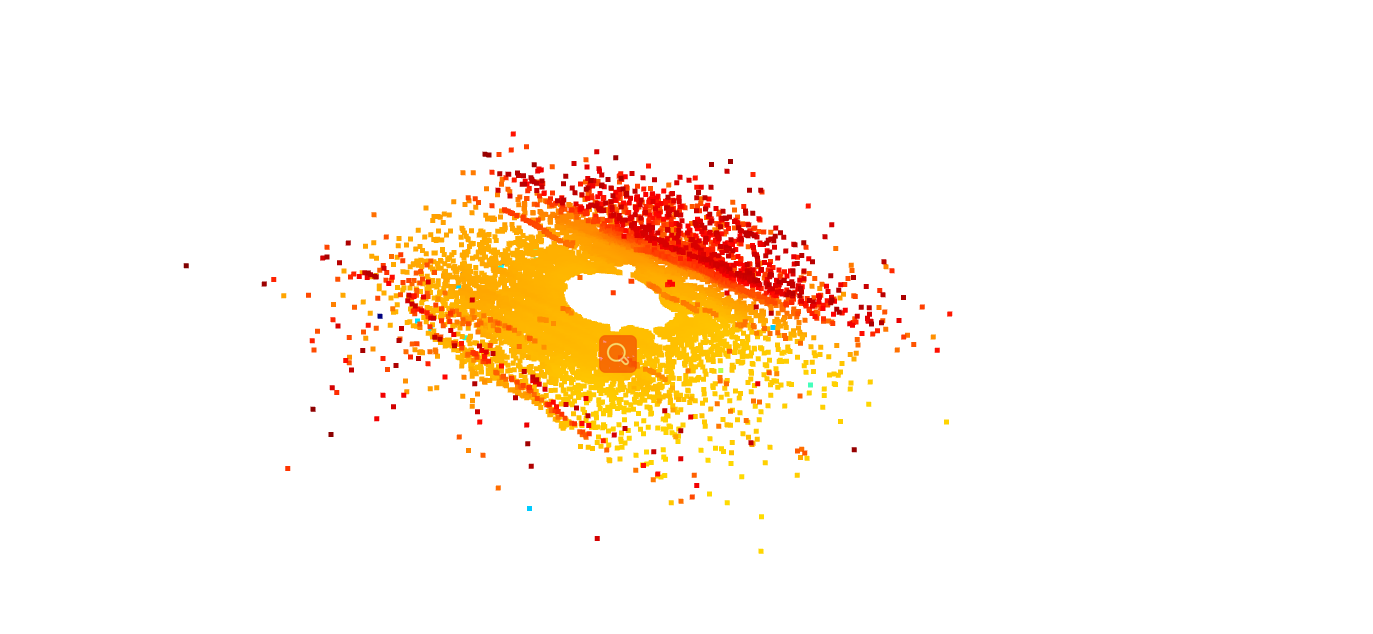} & 
 \adjincludegraphics[width=0.33\textwidth, trim={{.2\width} {.3\height} {.2\width} {.3\height}},clip]{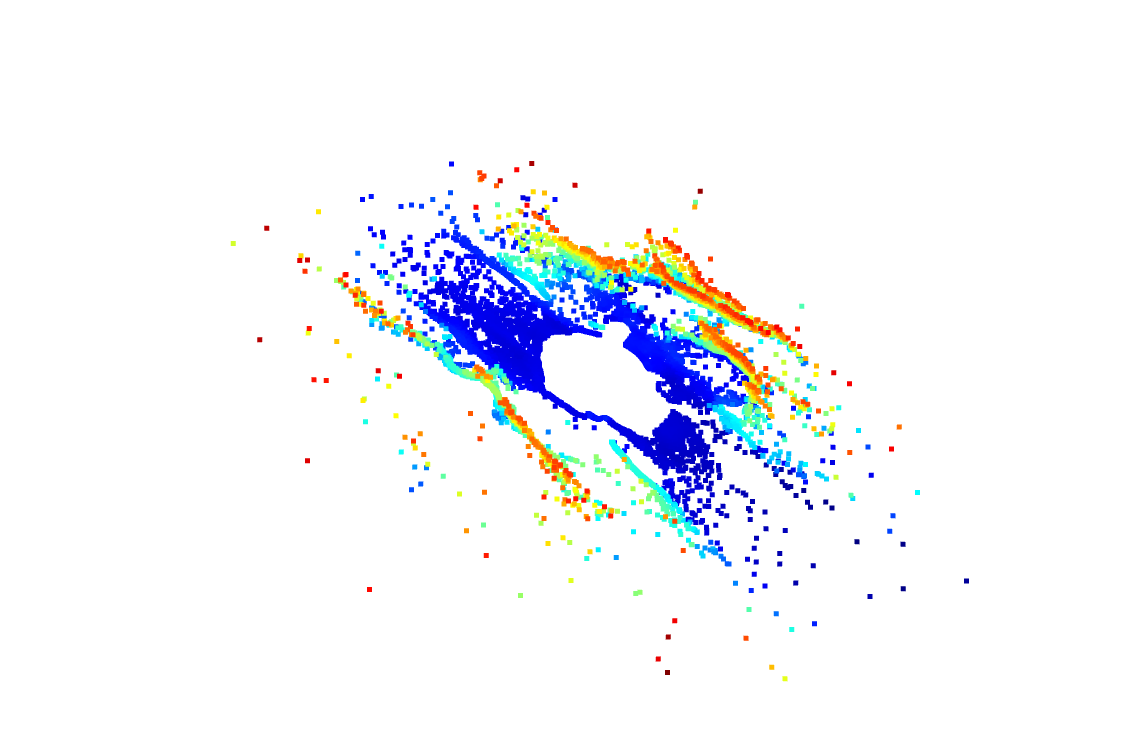} & 
\adjincludegraphics[width=0.33\textwidth, trim={{.2\width} {.3\height} {.2\width} {.3\height}},clip]{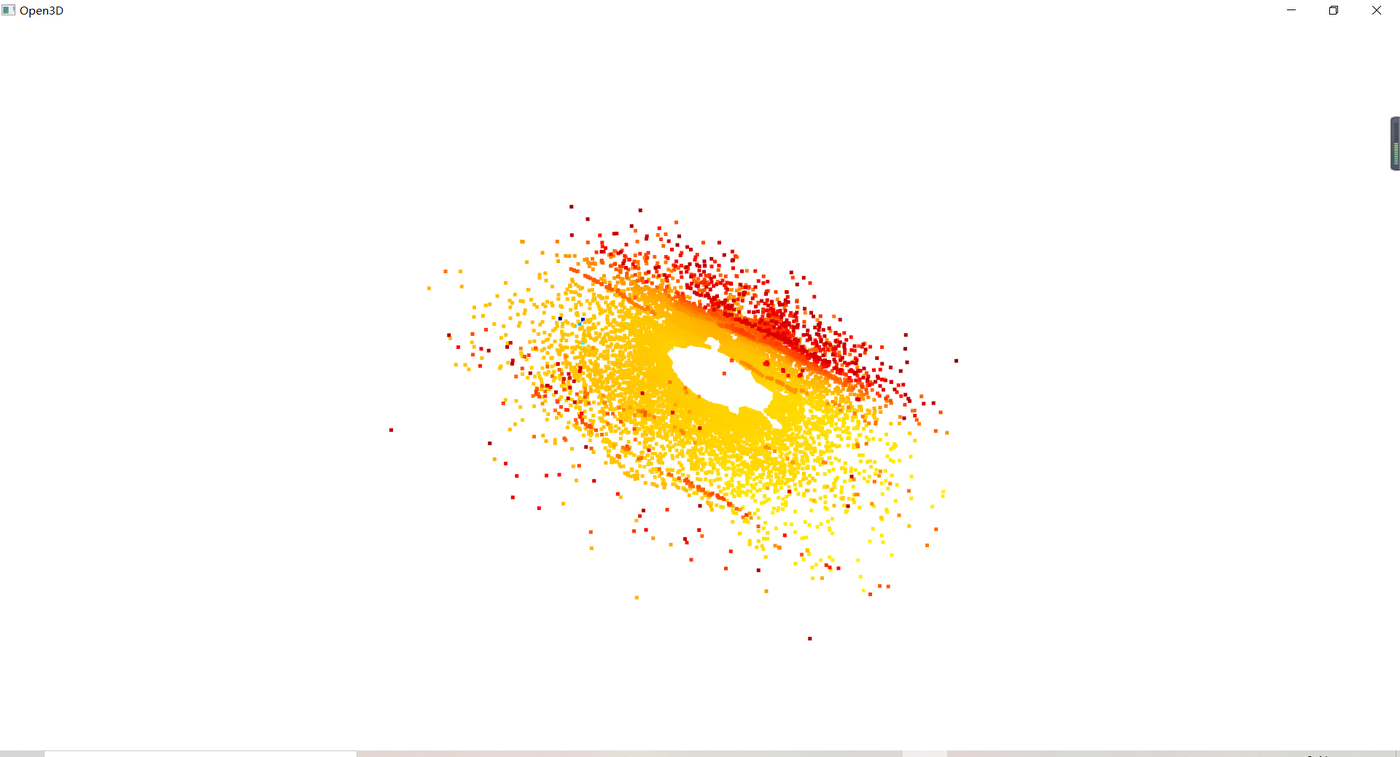} \\
\end{tabular}
\captionof{figure}{Qualitative Results for ShapeGF~\cite{cai2020learning} on KITTI.}
\label{fig:shapegf}
\end{table}

\begin{table}[!t]
\small
\centering
\begin{tabular}{ccc}
No Coord & No Coord & With Coord \\
No Circular & With Circular & With Circular \\

 \adjincludegraphics[width=0.33\textwidth, trim={{.2\width} {.3\height} {.2\width} {.3\height}},clip]{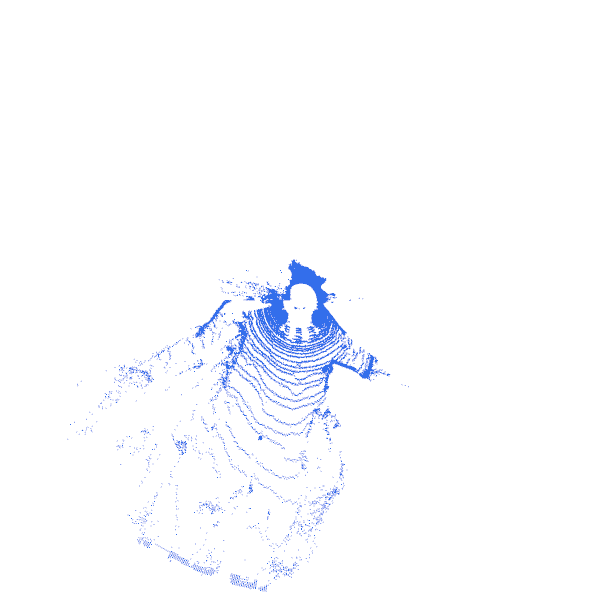} & 
 \adjincludegraphics[width=0.33\textwidth, trim={{.2\width} {.3\height} {.2\width} {.3\height}},clip]{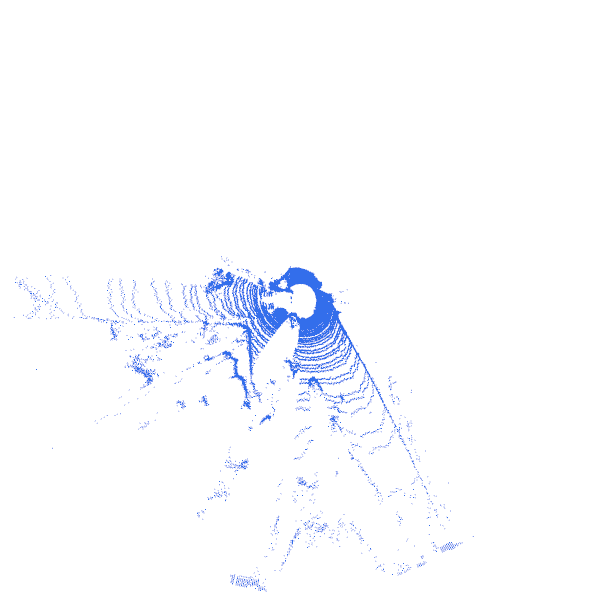} & 
\adjincludegraphics[width=0.33\textwidth, trim={{.2\width} {.3\height} {.2\width} {.3\height}},clip]{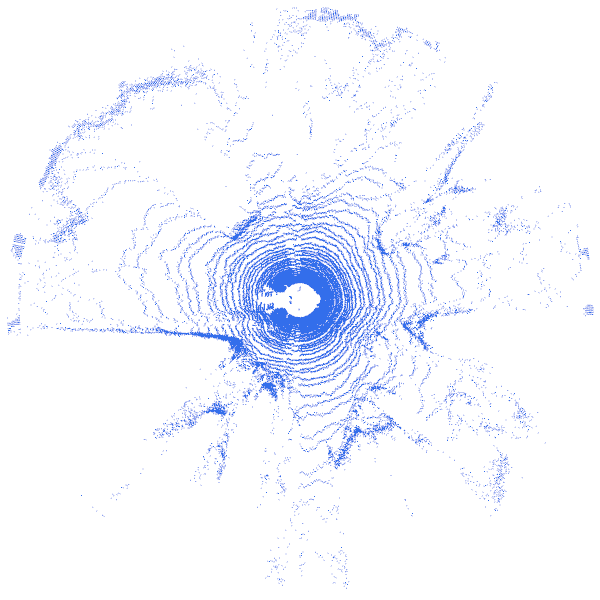} \\

 \adjincludegraphics[width=0.33\textwidth, trim={{.2\width} {.3\height} {.2\width} {.3\height}},clip]{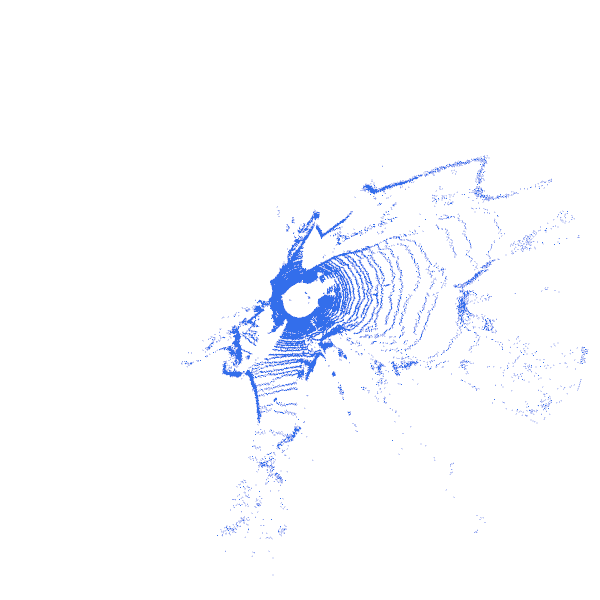} & 
 \adjincludegraphics[width=0.33\textwidth, trim={{.2\width} {.3\height} {.2\width} {.3\height}},clip]{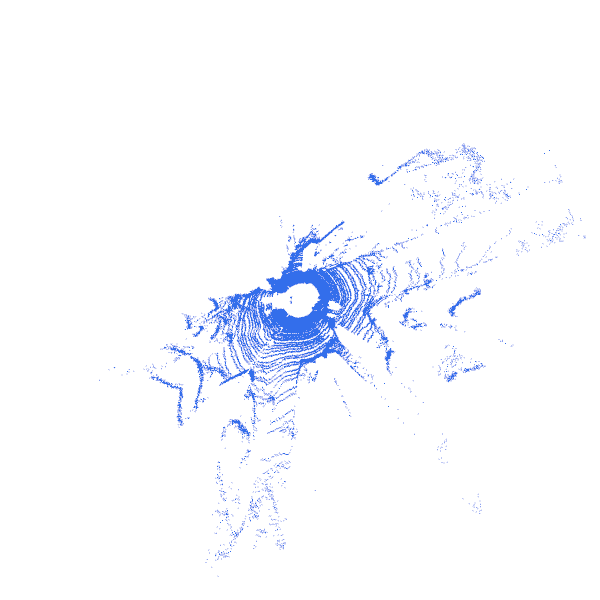} & 
\adjincludegraphics[width=0.33\textwidth, trim={{.2\width} {.3\height} {.2\width} {.3\height}},clip]{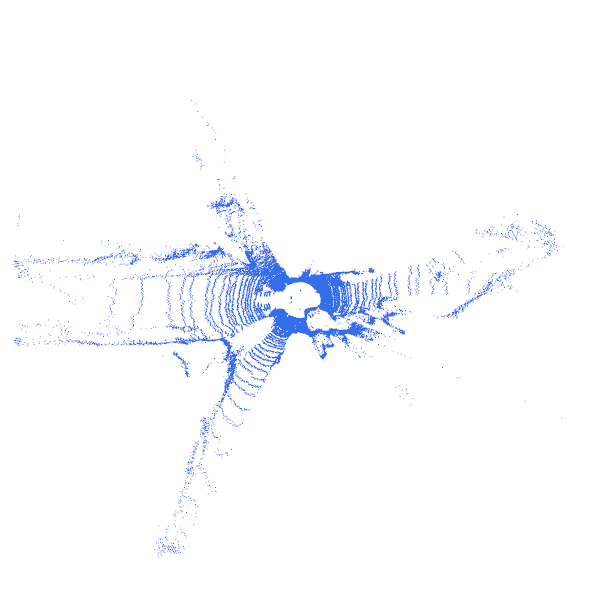} \\

 \adjincludegraphics[width=0.33\textwidth, trim={{.2\width} {.3\height} {.2\width} {.3\height}},clip]{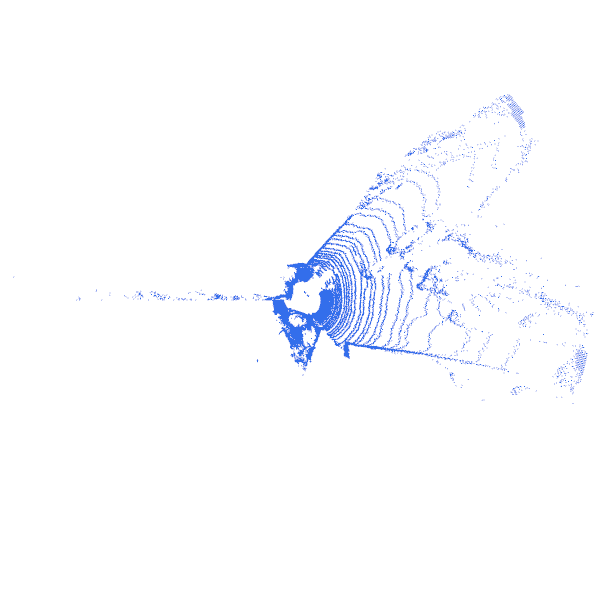} & 
 \adjincludegraphics[width=0.33\textwidth, trim={{.2\width} {.3\height} {.2\width} {.3\height}},clip]{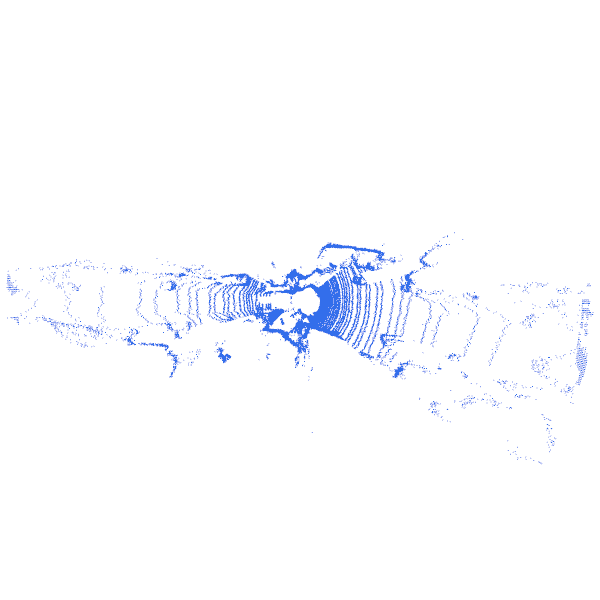} & 
\adjincludegraphics[width=0.33\textwidth, trim={{.2\width} {.3\height} {.2\width} {.3\height}},clip]{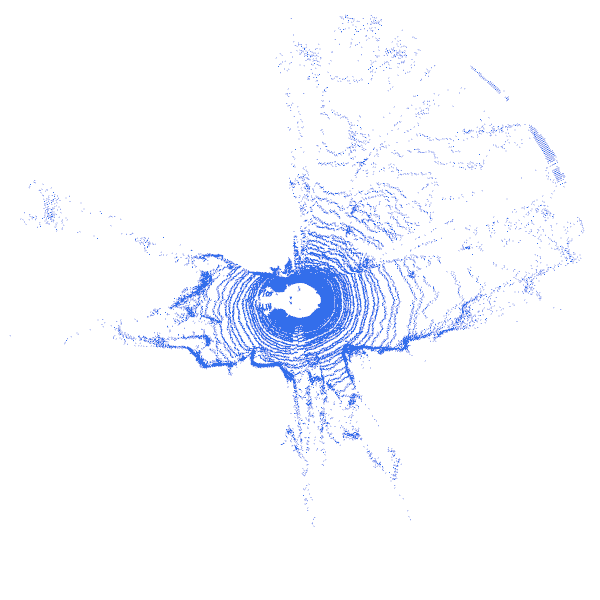} \\

 \adjincludegraphics[width=0.33\textwidth, trim={{.2\width} {.3\height} {.2\width} {.3\height}},clip]{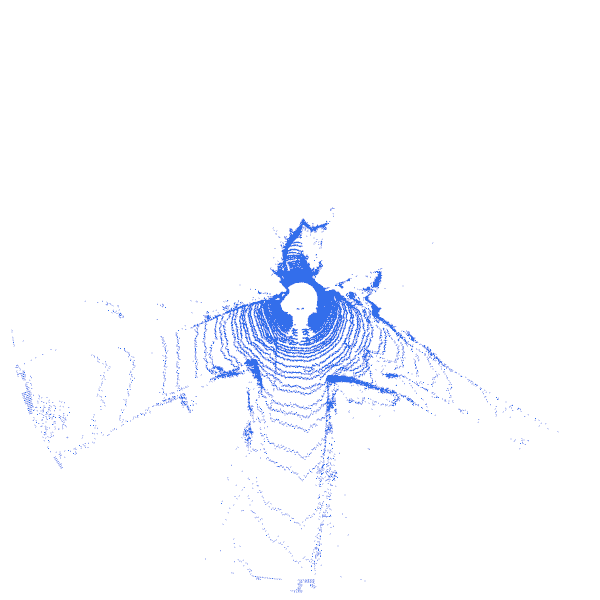} & 
 \adjincludegraphics[width=0.33\textwidth, trim={{.2\width} {.3\height} {.2\width} {.3\height}},clip]{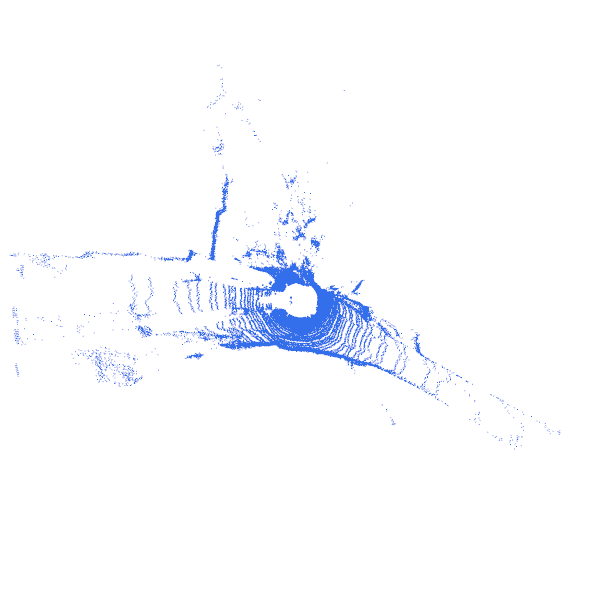} & 
\adjincludegraphics[width=0.33\textwidth, trim={{.2\width} {.3\height} {.2\width} {.3\height}},clip]{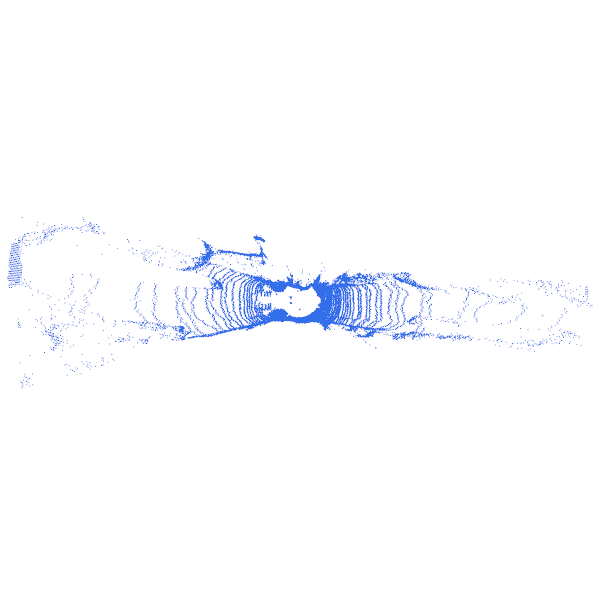} \\

 \adjincludegraphics[width=0.33\textwidth, trim={{.2\width} {.3\height} {.2\width} {.3\height}},clip]{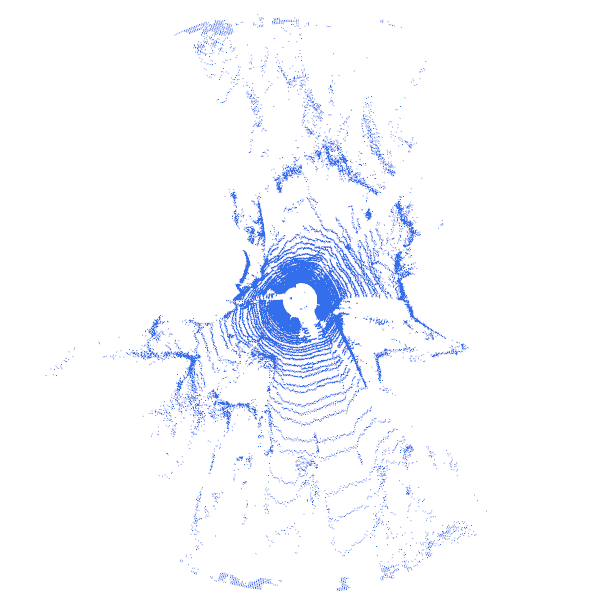} & 
 \adjincludegraphics[width=0.33\textwidth, trim={{.2\width} {.3\height} {.2\width} {.3\height}},clip]{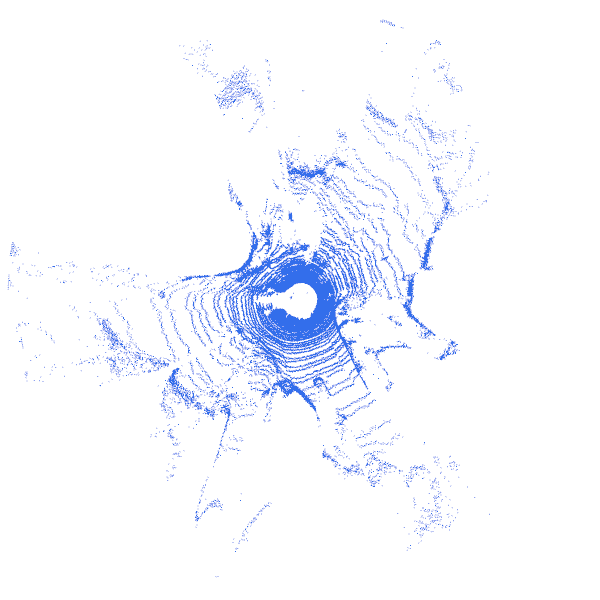} & 
\adjincludegraphics[width=0.33\textwidth, trim={{.2\width} {.3\height} {.2\width} {.3\height}},clip]{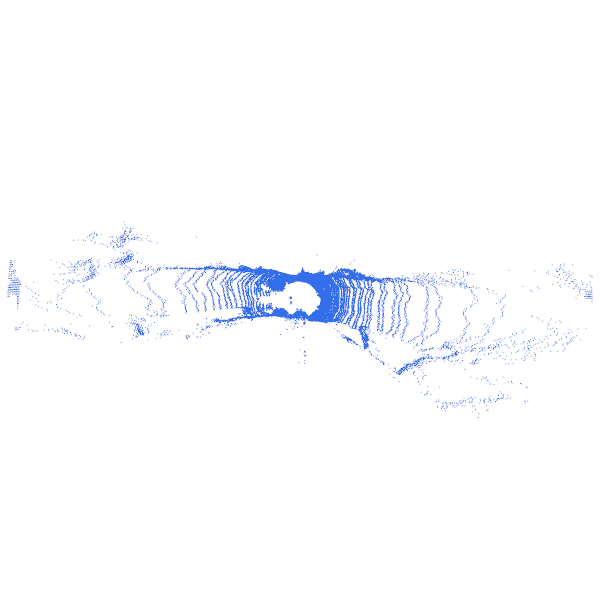} \\
\end{tabular}
\captionof{figure}{Qualitative Ablation Comparison. Circular convolutions prevents a discontinuity that happens on a vector that starts at the origin and points left. The coordinate encoding encourages the network to generate straighter and more realistic roads. }
\label{fig:ablation}
\end{table}

\section{nuScenes}
\label{sec:nuscenes}

\mypara{Datasets}
The proposed LiDAR generation model is applicable across different datasets, geolocations and LiDAR sensors. To demonstrate this, we train and test our model on the \textbf{nuScenes dataset} \cite{nuscenes2019}. nuScenes contains 297,737 LiDAR sweeps in the training set and 52,423 LiDAR sweeps in the testing and cross-validation set. The LiDAR sweeps were collected in the cities of Boston and Singapore. These locations present readings that are uniquely different compared to readings from the KITTI dataset \cite{Kitti360}, which have mostly been collected in the suburbs of Karlsruhe, Germany.  In addition to the different environment, the LiDAR sensor used in nuScenes is different. The data was collected with a Velodyne HDL32E, which has 32 Beams and a $+10^{\circ}$ to $-30^{\circ}$ vertical Field of View. Compared to the sensor used in the KITTI dataset (Velodyne HDL-64E), the one used in nuScenes has lower vertical resolution, however has a higher vertical field of view. 

\mypara{Implementation Details}
We encode the raw nuScenes point cloud into an equirectangular view. 
Specifically, our range image resolution is set to be $32\times1024$, tailored for nuScenes LiDAR sensor's spatial resolution. And our Cartesian-to-range encoding is changed to the following: 
$\bz_i=(\theta_i, \phi_i, d_i), r_i$:
$\mathcal{I}(\lfloor\theta_i / s_\theta\rfloor, \lfloor\phi_j / s_\phi\rfloor) = \left(\frac{1}{6.5} \log_2(d_i + 1), \frac{1}{31} r_i \right)$, ensuring the full range to be normalized to $[0, 1]$,

\mypara{Baselines} Following the main paper's experiments on KITTI-360. We also leverage two baselines for comparison. The first is ProjectedGAN \cite{ProjectedGAN}. This was the second-best performing model in the KITTI evaluation, so we include it in this evaluation too. The second baseline is Caccia et al.'s LiDAR VAE \cite{caccia}. All models were trained with the same settings as for KITTI. Note that the GAN model described in Caccia et al.~\cite{caccia} does not converge after our hyper-parameter tuning, hence we omit it from this study.

\mypara{Experimental Results} Fig.~\ref{fig:nuscenes} depicts the qualitative comparison results. From this figure, we can see that our method still achieves superior results compared to both VAE~\cite{caccia} and projected GAN~\cite{ProjectedGAN}. An AB test on a group of four human subjects suggests that our method is still significantly favored over other competing algorithms in $89\%$ of cases. 

While achieving superior human performance, we notice that our current nuScenes model has a noticeable weakness on the nuScenes dataset. In particular, our method tends to generate point clouds that concentrate their mass closer to the viewpoint. As a result, despite superior visual quality, our MMD score at BEV is worse than VAE and Projected GAN (2e-3 vs. 1.1e-3 and 6e-5). In the nuScenes experiment, we directly adopt the same starting and ending noise level (150 and 0.01) as in KITTI, which might be too large for nuScenes. We believe better-tuned noise parameters will resolve this issue.  

\begin{table}[!t]
\small
\centering
\begin{tabular}{rcccc}
Ground-Truth & VAE~\cite{caccia} & ProjectedGAN~\cite{ProjectedGAN} & \textbf{Ours}\\
\adjincludegraphics[width=0.24\textwidth, trim={{.3\width} {.3\height} {.3\width} {.3\height}},clip]{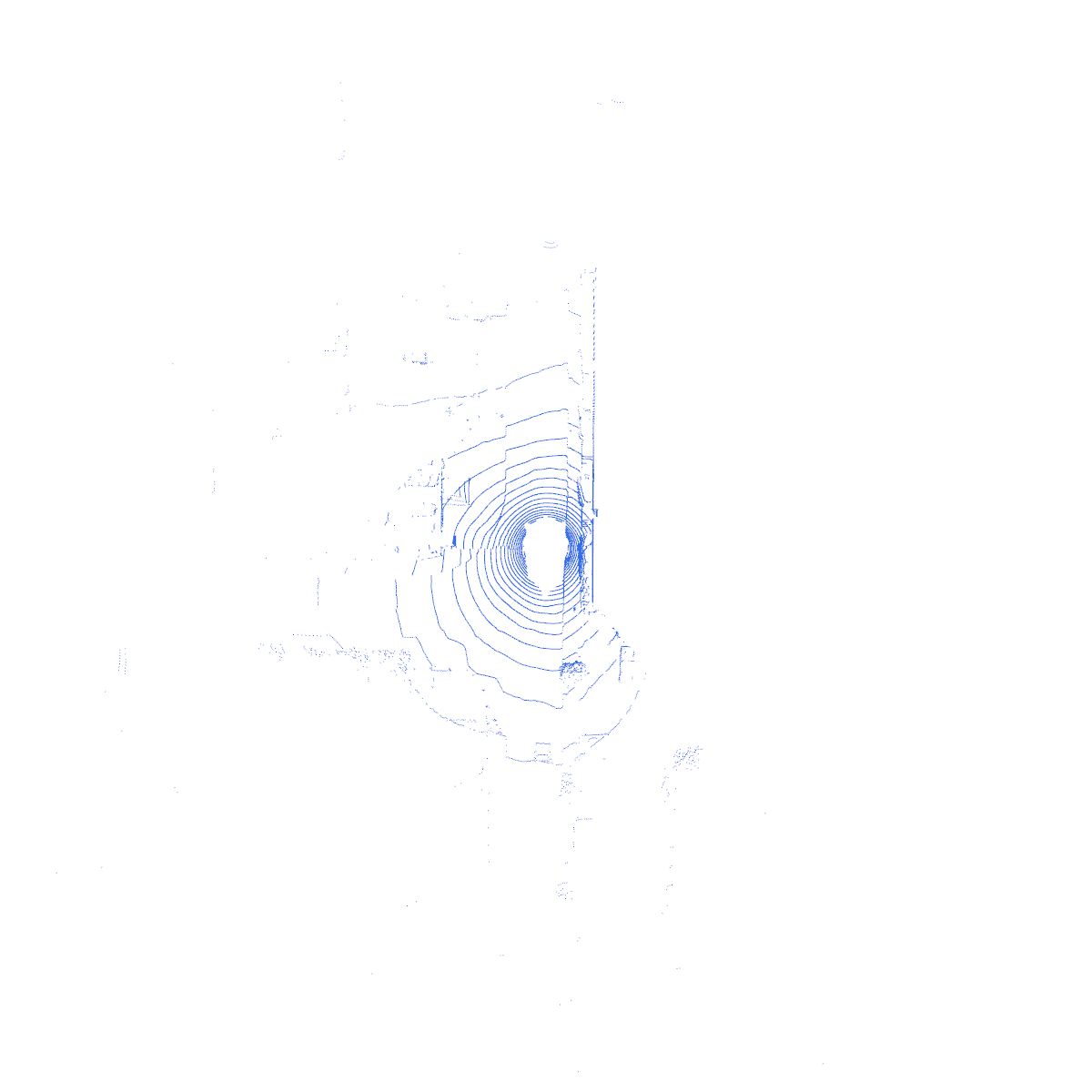} & 
\adjincludegraphics[width=0.24\textwidth, trim={{.3\width} {.3\height} {.3\width} {.3\height}},clip]{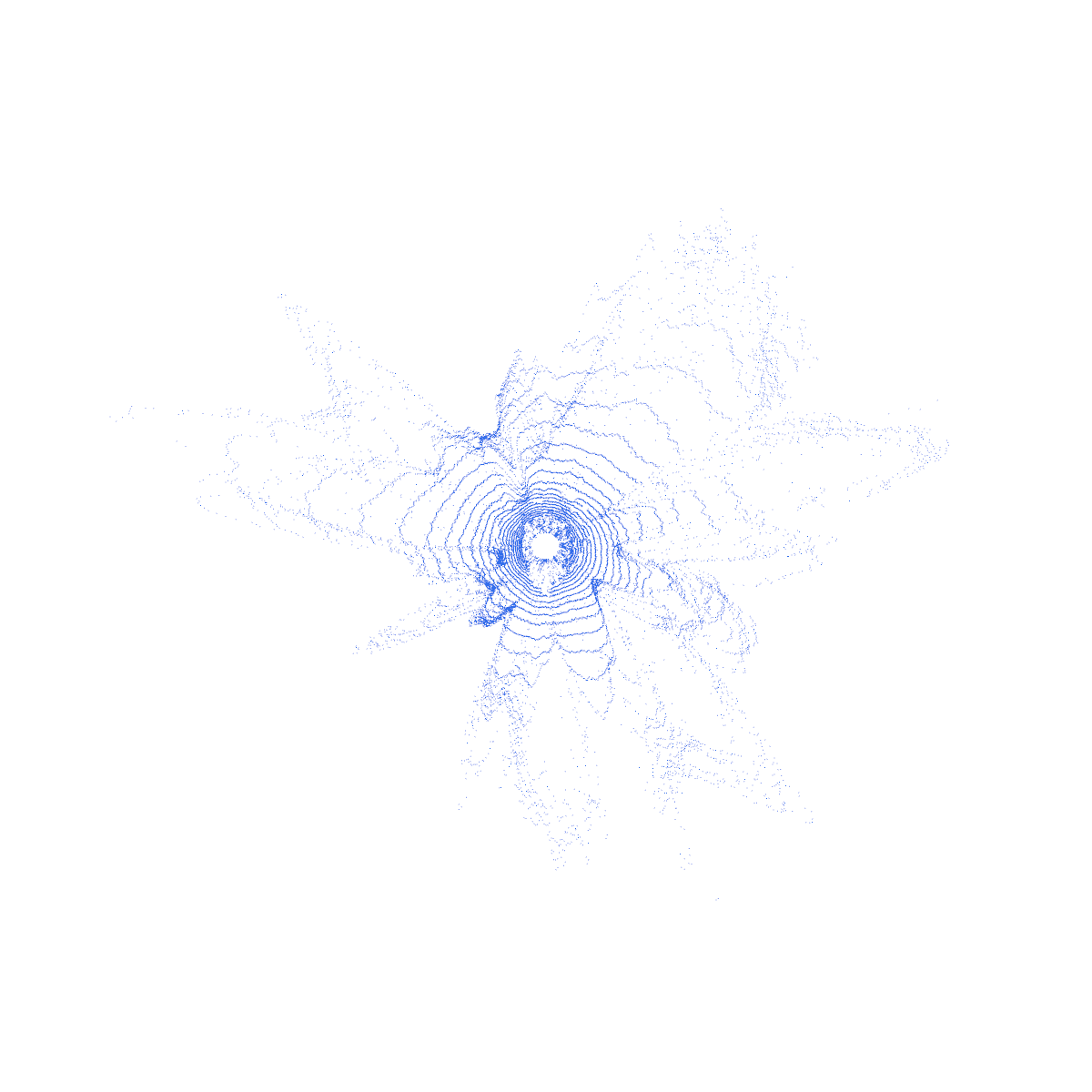} & 
\adjincludegraphics[width=0.24\textwidth, trim={{.3\width} {.3\height} {.3\width} {.3\height}},clip]{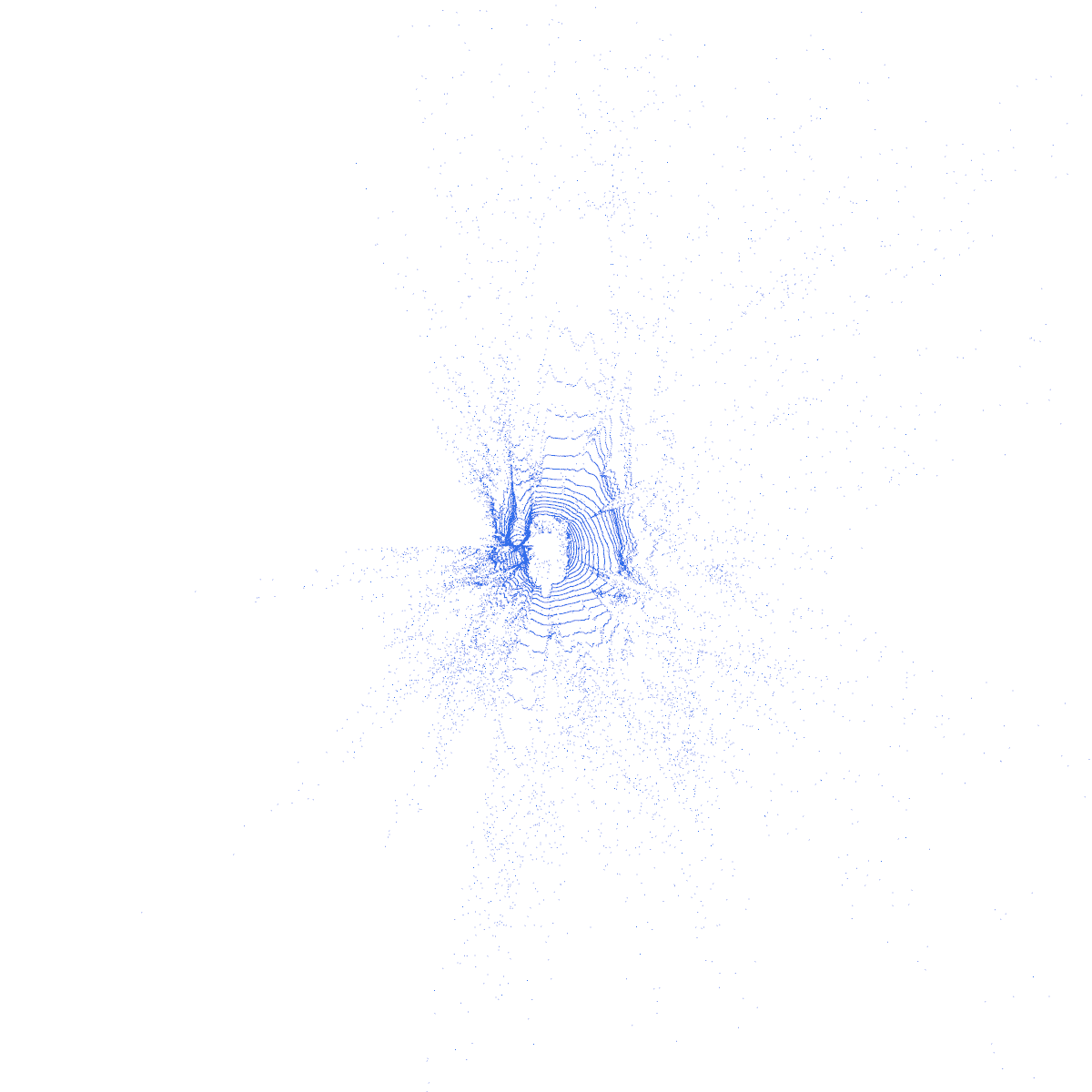}  & 
\adjincludegraphics[width=0.24\textwidth, trim={{.3\width} {.3\height} {.3\width} {.3\height}},clip]{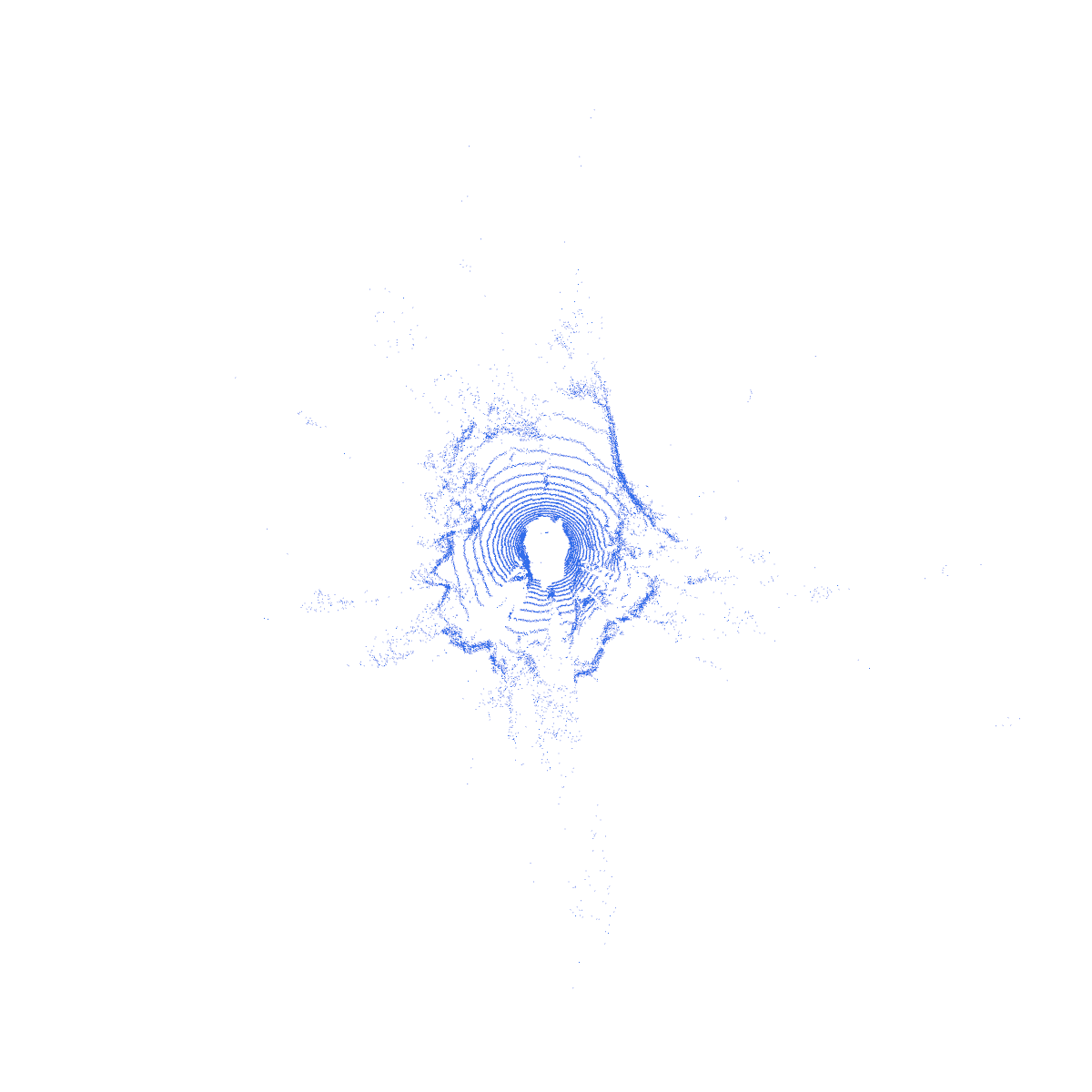}\\

\adjincludegraphics[width=0.24\textwidth, trim={{.3\width} {.3\height} {.3\width} {.3\height}},clip]{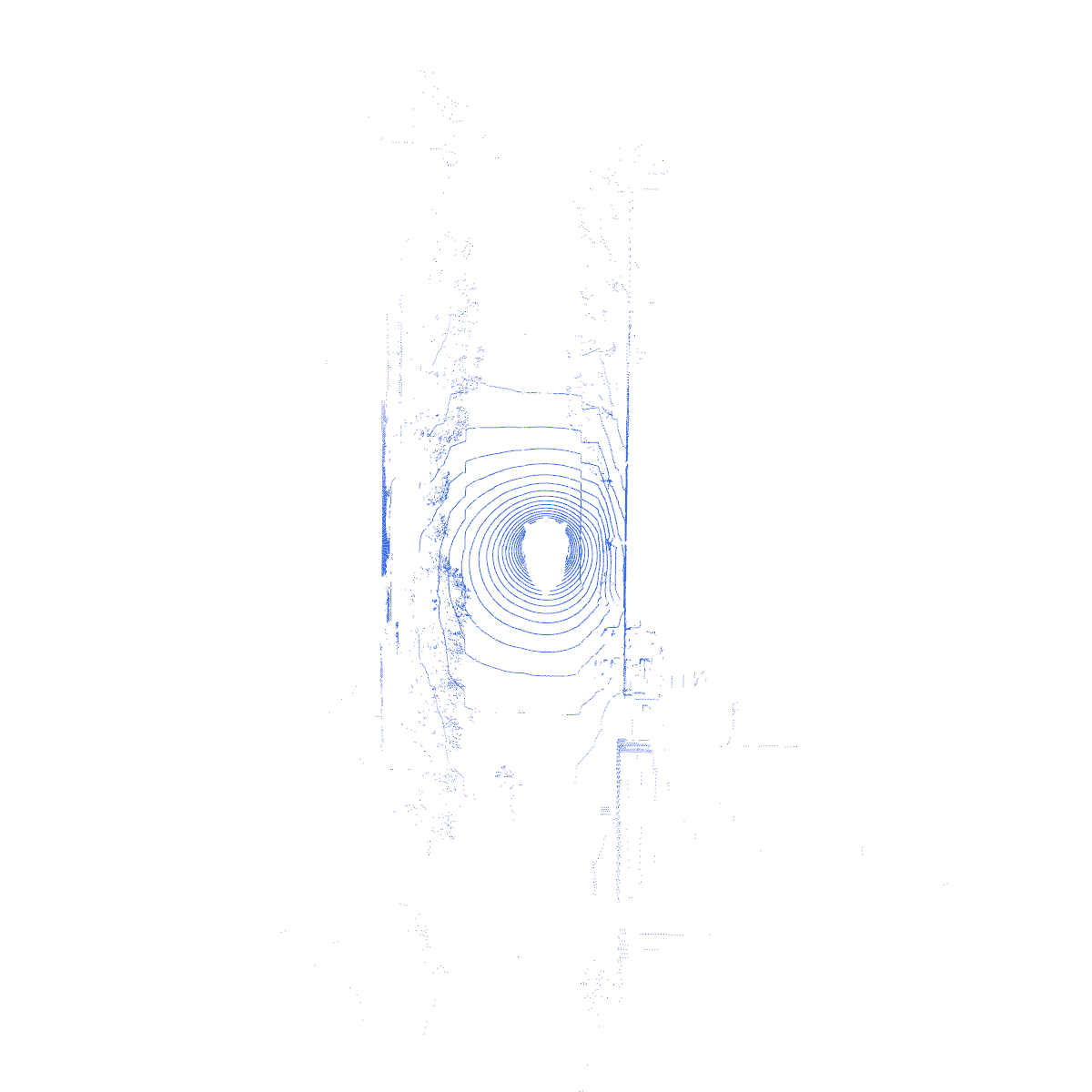} & 
\adjincludegraphics[width=0.24\textwidth, trim={{.3\width} {.3\height} {.3\width} {.3\height}},clip]{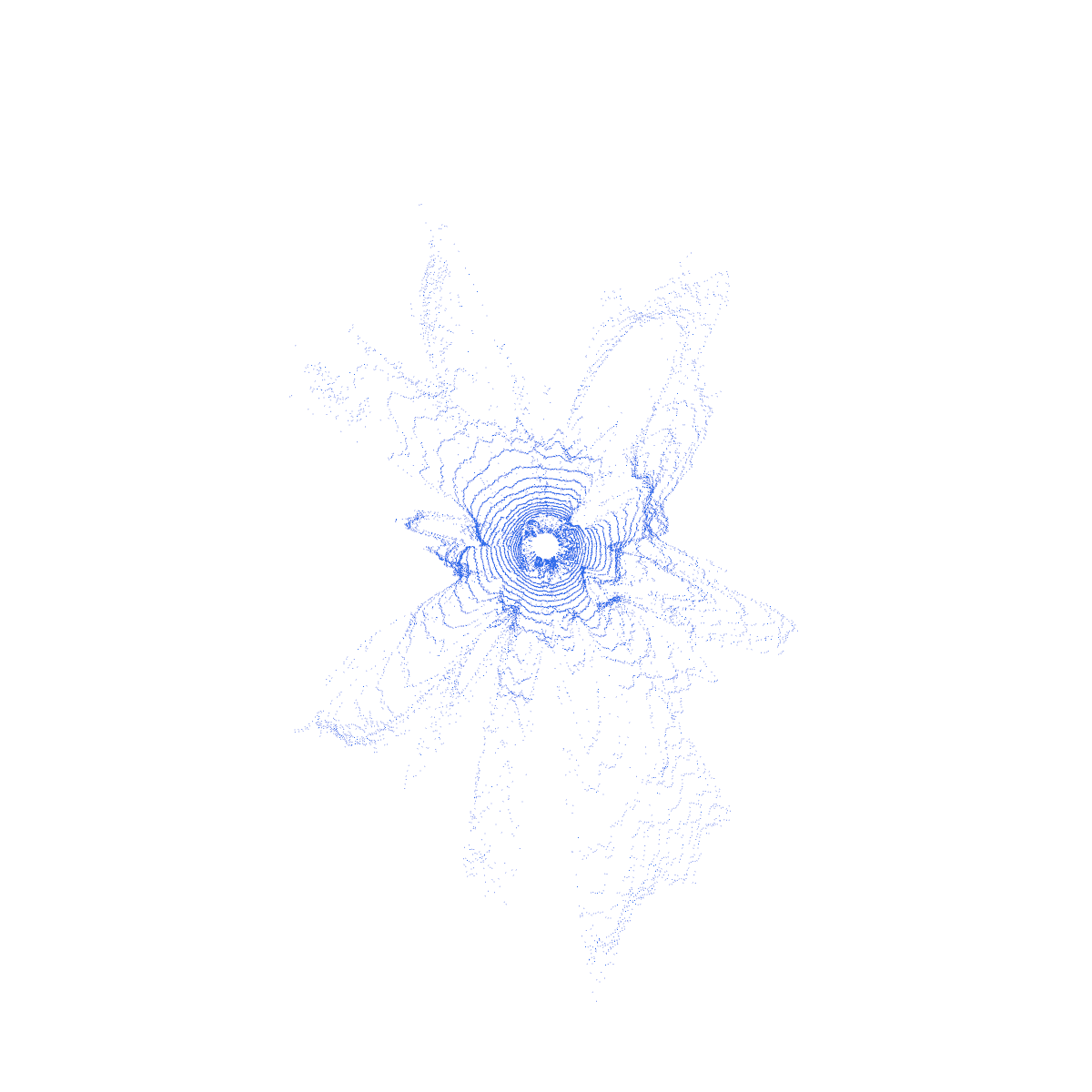} & 
\adjincludegraphics[width=0.24\textwidth, trim={{.3\width} {.3\height} {.3\width} {.3\height}},clip]{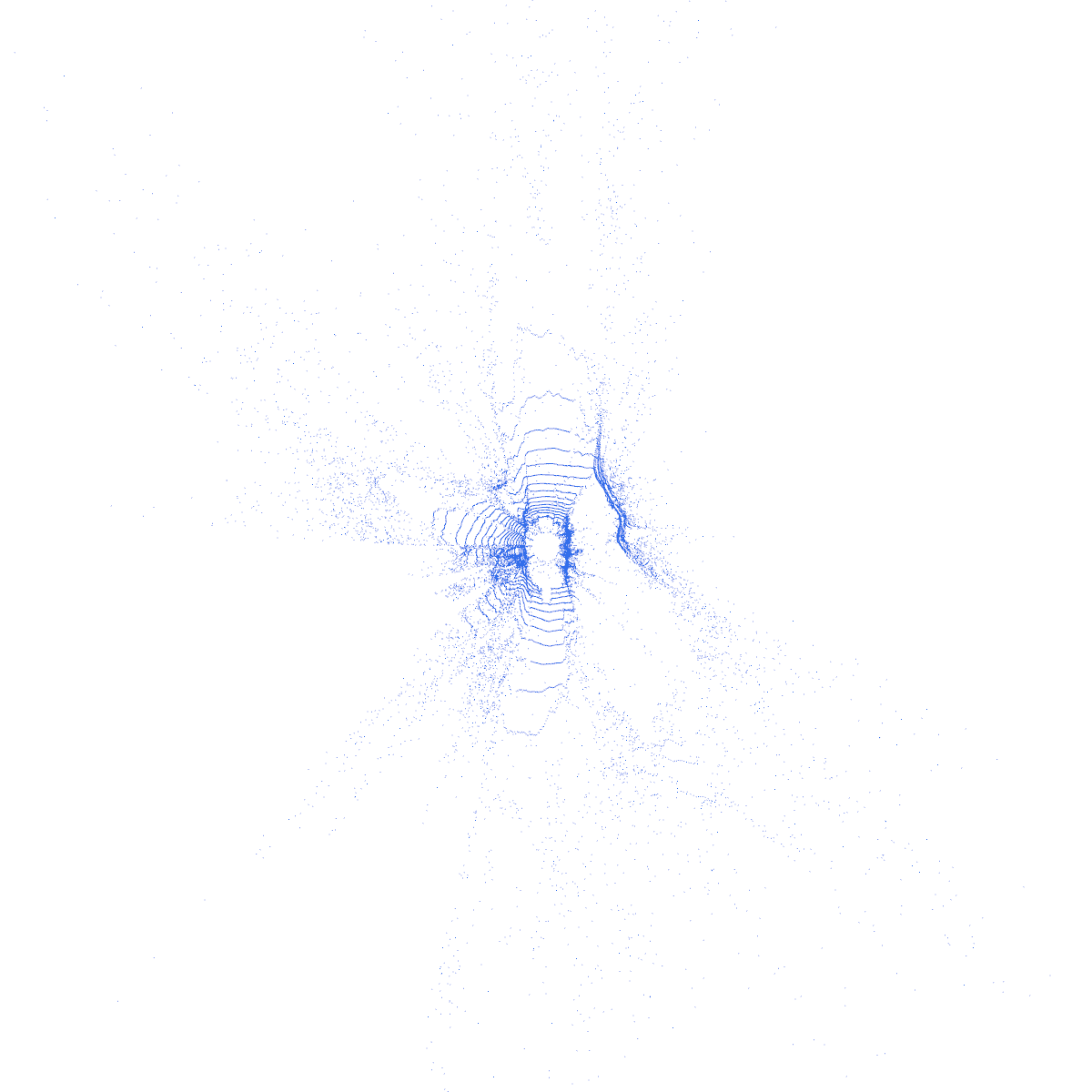}  & 
\adjincludegraphics[width=0.24\textwidth, trim={{.3\width} {.3\height} {.3\width} {.3\height}},clip]{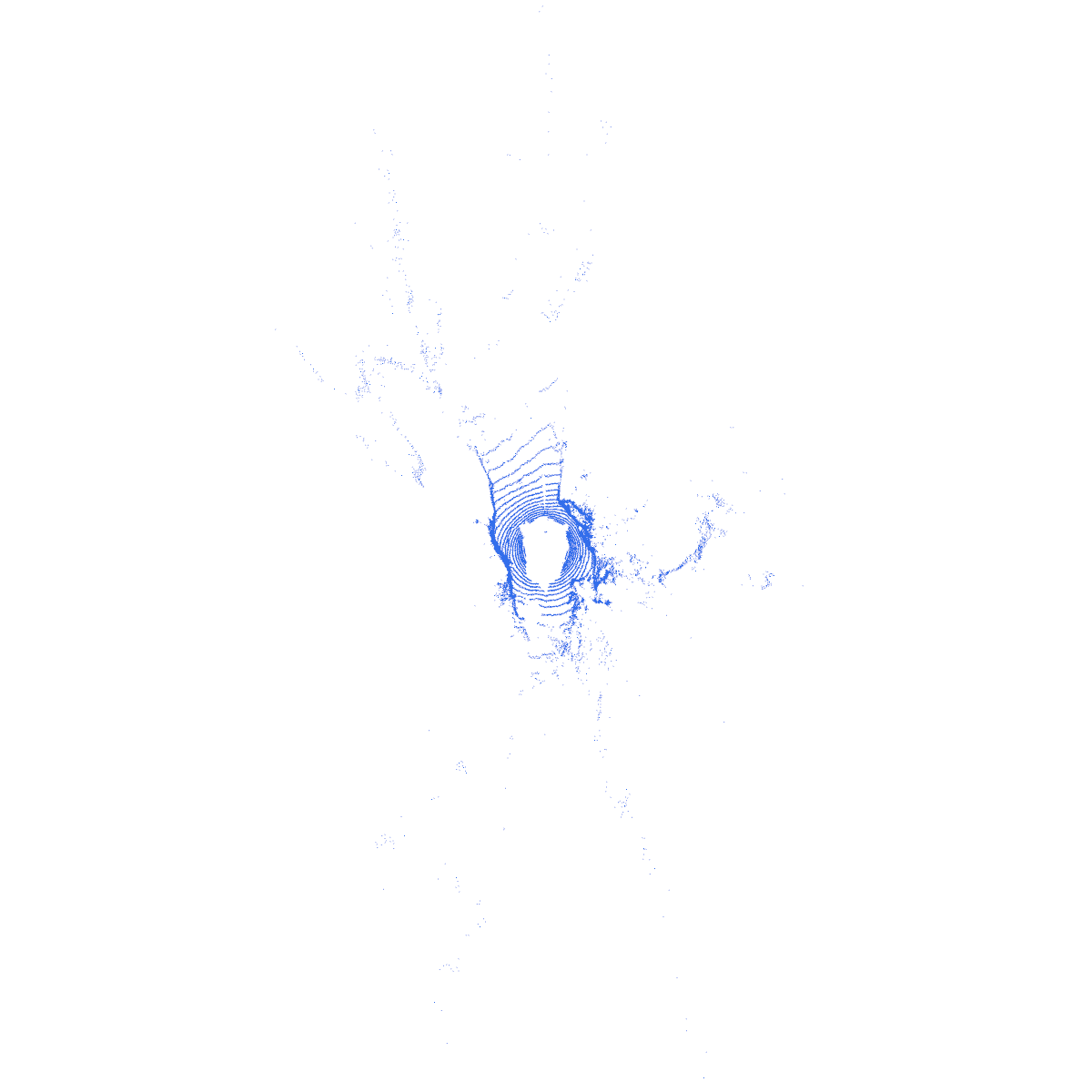}\\

\adjincludegraphics[width=0.24\textwidth, trim={{.3\width} {.3\height} {.3\width} {.3\height}},clip]{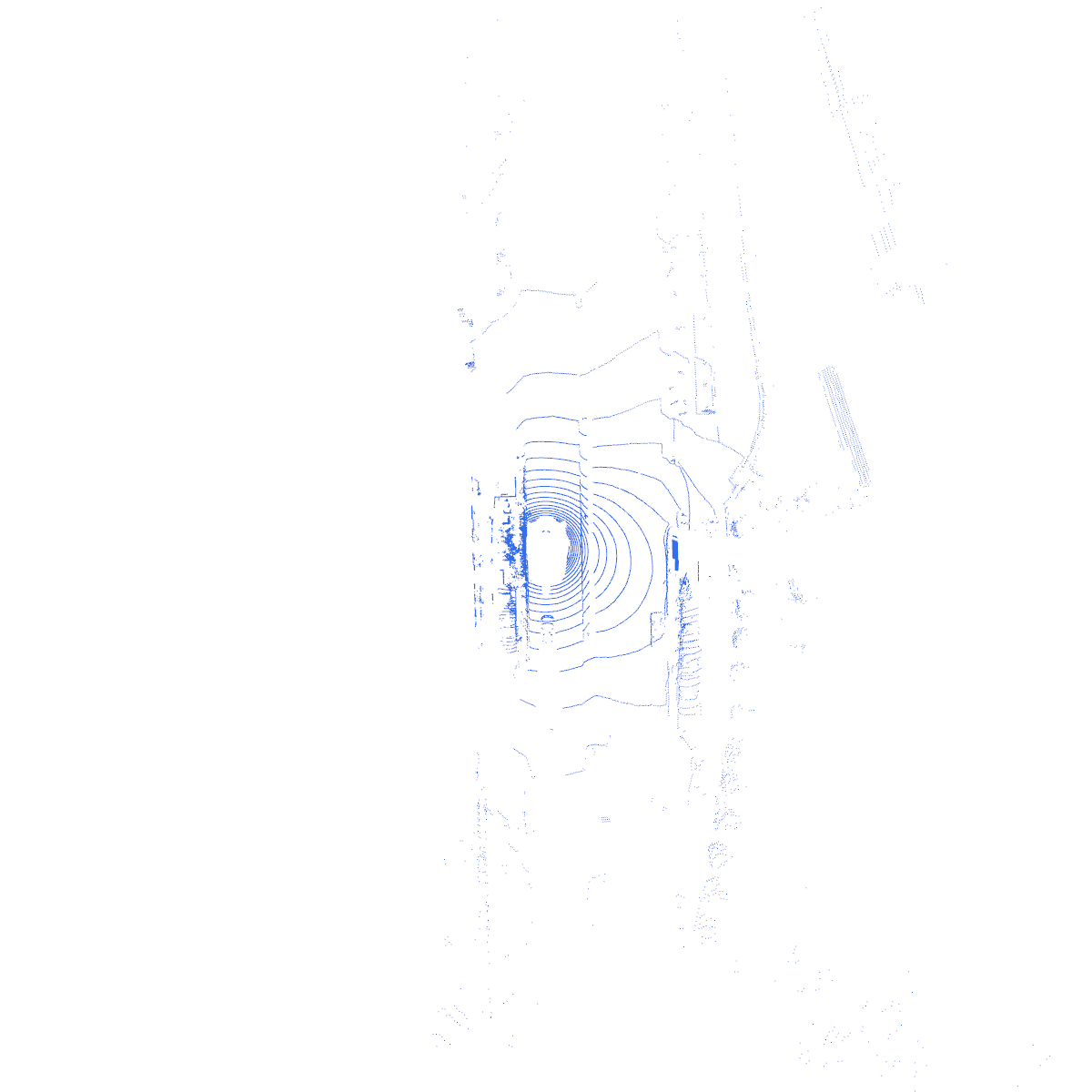} & 
\adjincludegraphics[width=0.24\textwidth, trim={{.3\width} {.3\height} {.3\width} {.3\height}},clip]{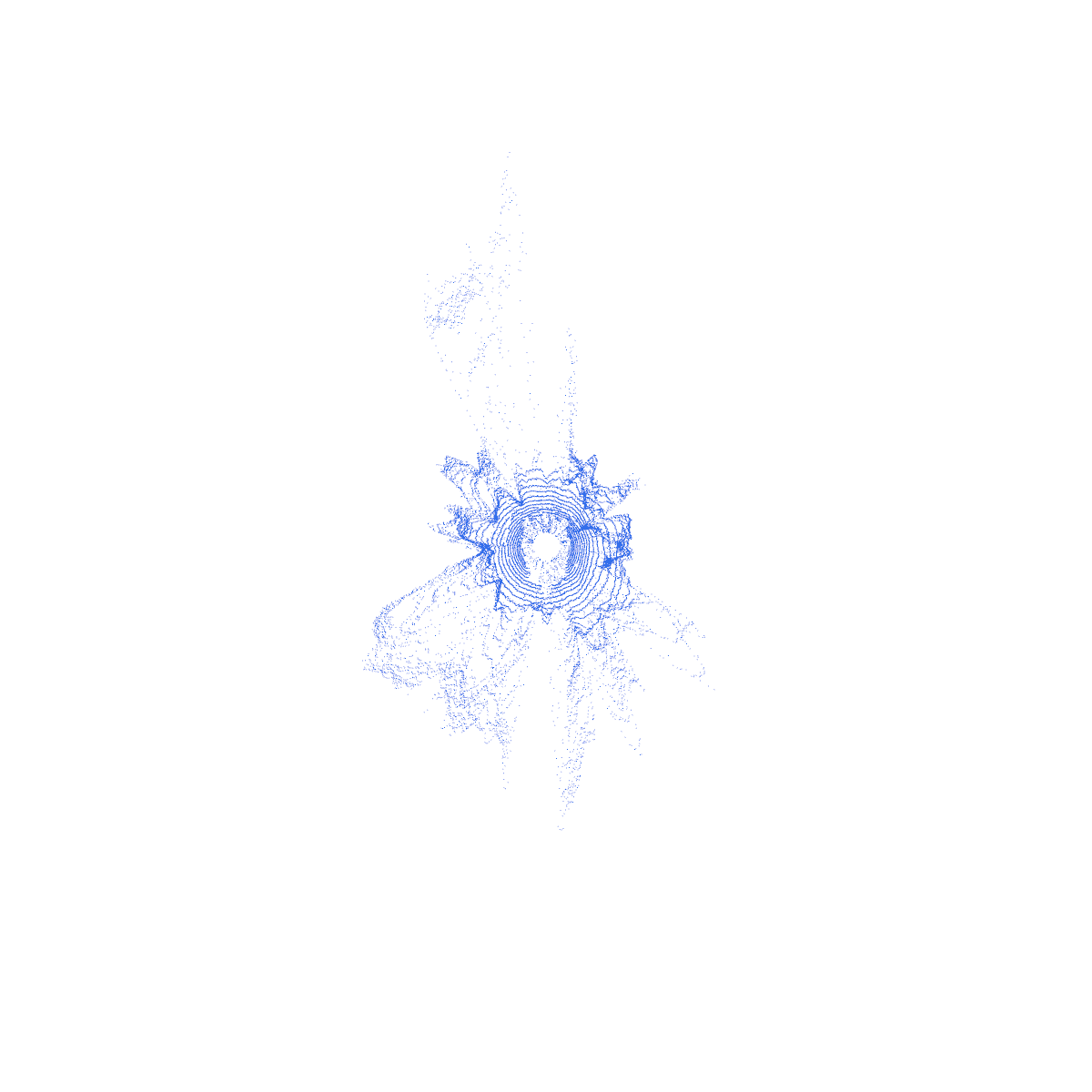} & 
\adjincludegraphics[width=0.24\textwidth, trim={{.3\width} {.3\height} {.3\width} {.3\height}},clip]{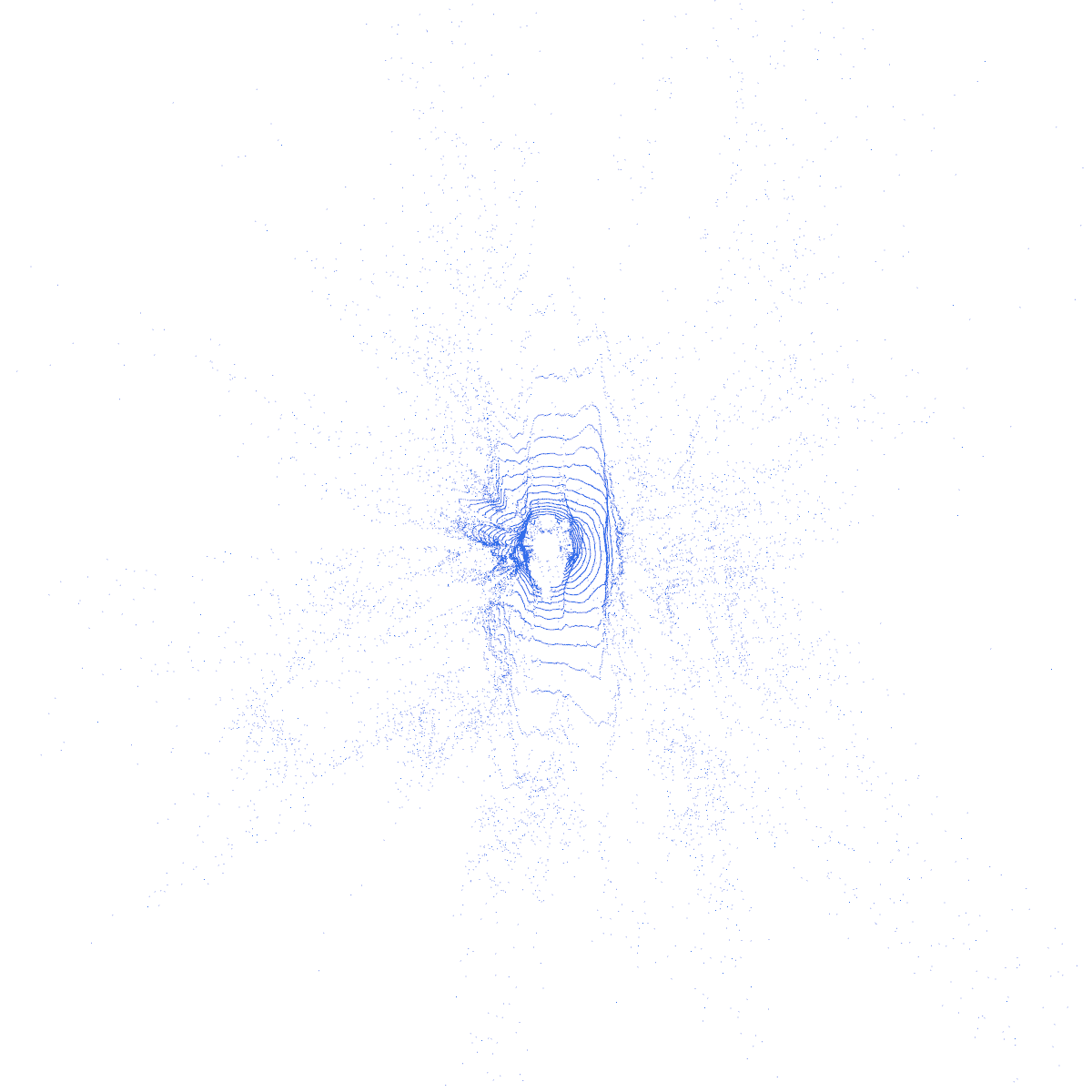}  & 
\adjincludegraphics[width=0.24\textwidth, trim={{.3\width} {.3\height} {.3\width} {.3\height}},clip]{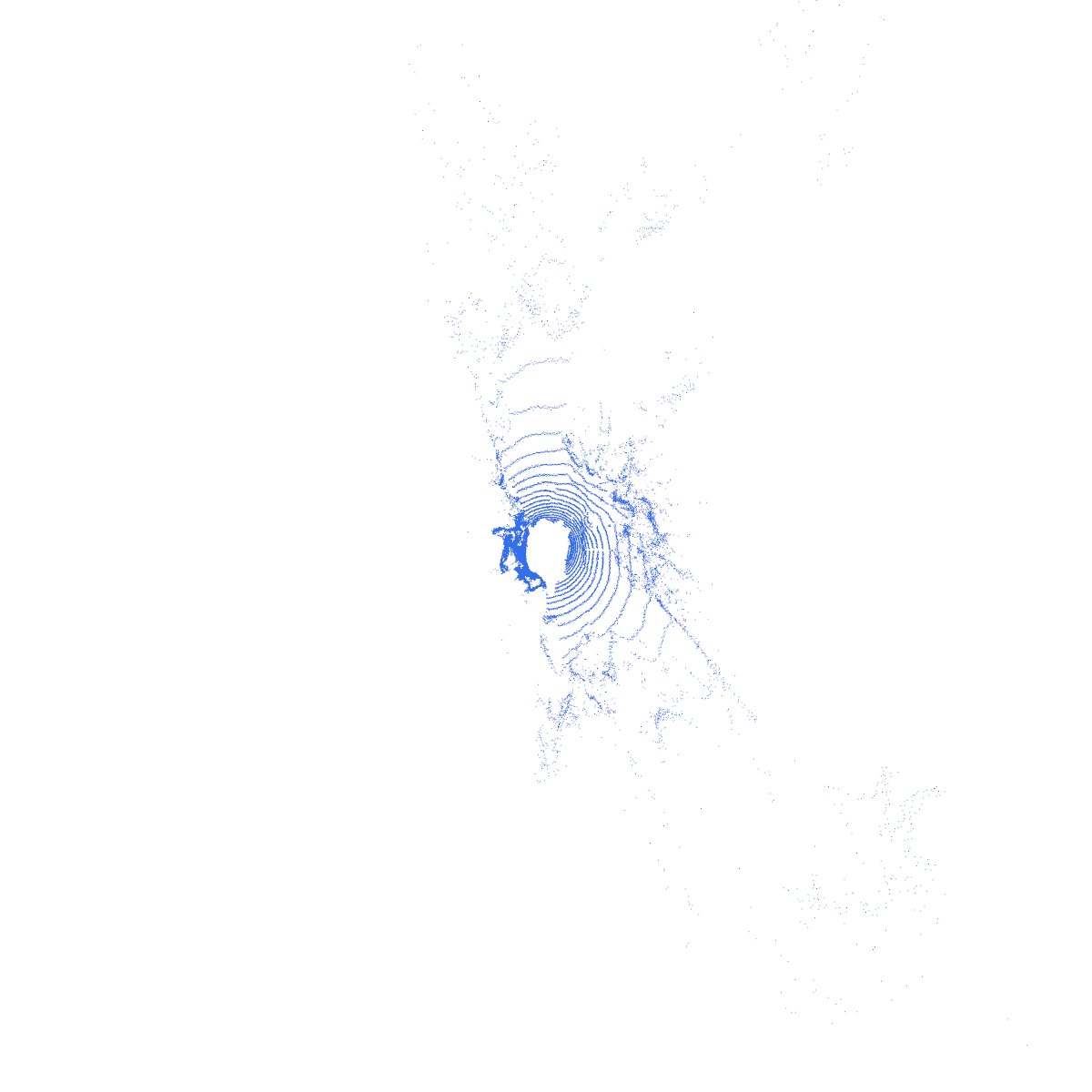}\\

\adjincludegraphics[width=0.24\textwidth, trim={{.3\width} {.3\height} {.3\width} {.3\height}},clip]{nuscenes/nusc_nusc/out_5.png} & 
\adjincludegraphics[width=0.24\textwidth, trim={{.3\width} {.3\height} {.3\width} {.3\height}},clip]{nuscenes/nusc_vae/out_5.png} & 
\adjincludegraphics[width=0.24\textwidth, trim={{.3\width} {.3\height} {.3\width} {.3\height}},clip]{nuscenes/nusc_pgan/out_5.png}  & 
\adjincludegraphics[width=0.24\textwidth, trim={{.3\width} {.3\height} {.3\width} {.3\height}},clip]{nuscenes/nusc_ncsn/out_5.png}\\
\end{tabular}
\captionof{figure}{Qualitative Results on the nuScenes dataset. }
\label{fig:nuscenes}
\end{table}

\section{Additional Posterior Sampling Results}
\label{sec:posterior}

\mypara{Densification} 
We demonstrate additional densification results in Fig.~\ref{fig:more_densification} and Fig.~\ref{fig:more_densification2}. From the figure, we can see our produced diversified point clouds are both highly-realistic and have high fidelity and consistency to the input. 

Fig~\ref{fig:diverse} provides additional results demonstrating multiple samples given the same sparse point input. The figure shows that our posterior sampling approach produces multiple plausible resulting point clouds, further demonstrating the advantage of tackling such a task in a probabilistic fashion.

\begin{table}[!t]
\small
\centering
\begin{tabular}{cccc}
GT & Sample 1  & Sample 2  & Sample 3 \\
\adjincludegraphics[width=0.24\textwidth, trim={{.2\width} {.2\height} {.2\width} {.2\height}},clip]{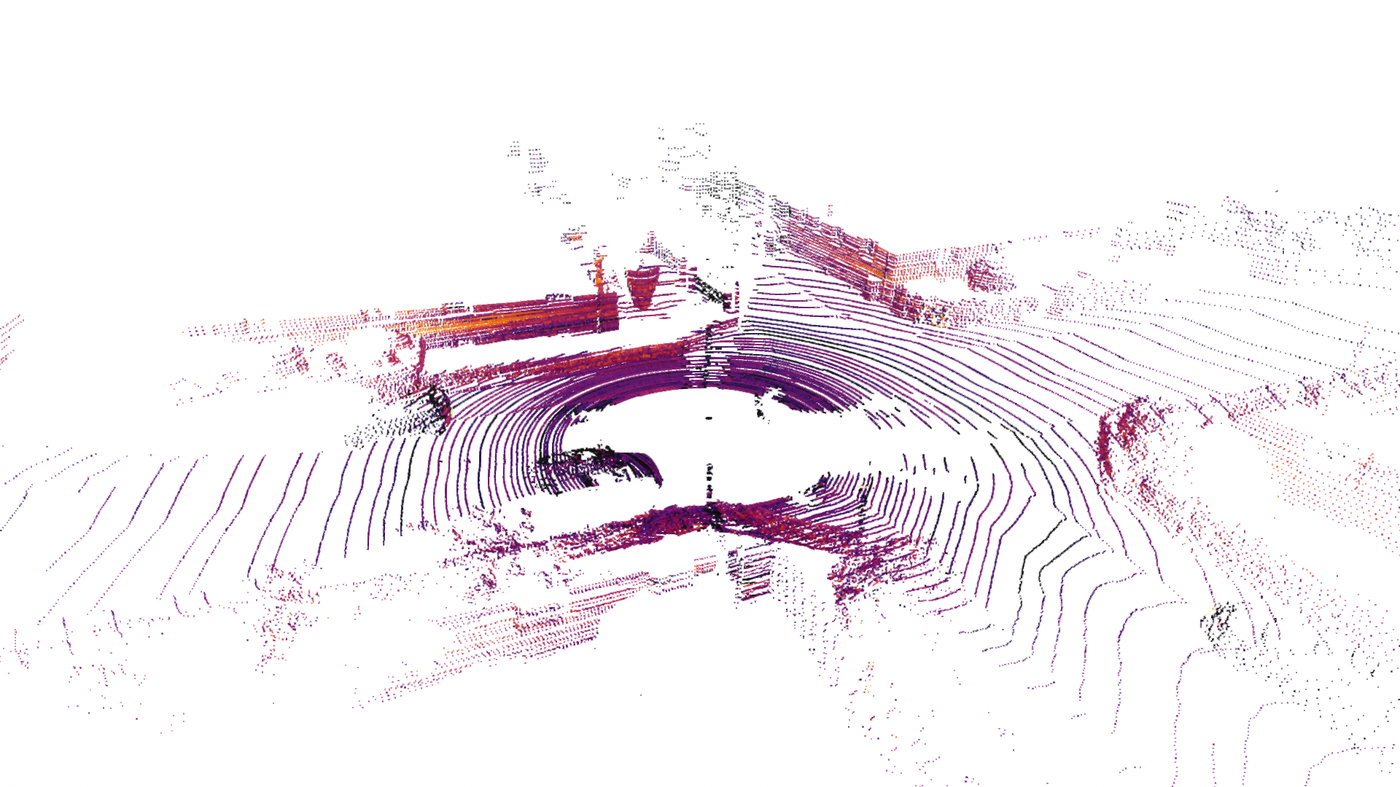} & 
\adjincludegraphics[width=0.24\textwidth, trim={{.2\width} {.2\height} {.2\width} {.2\height}},clip]{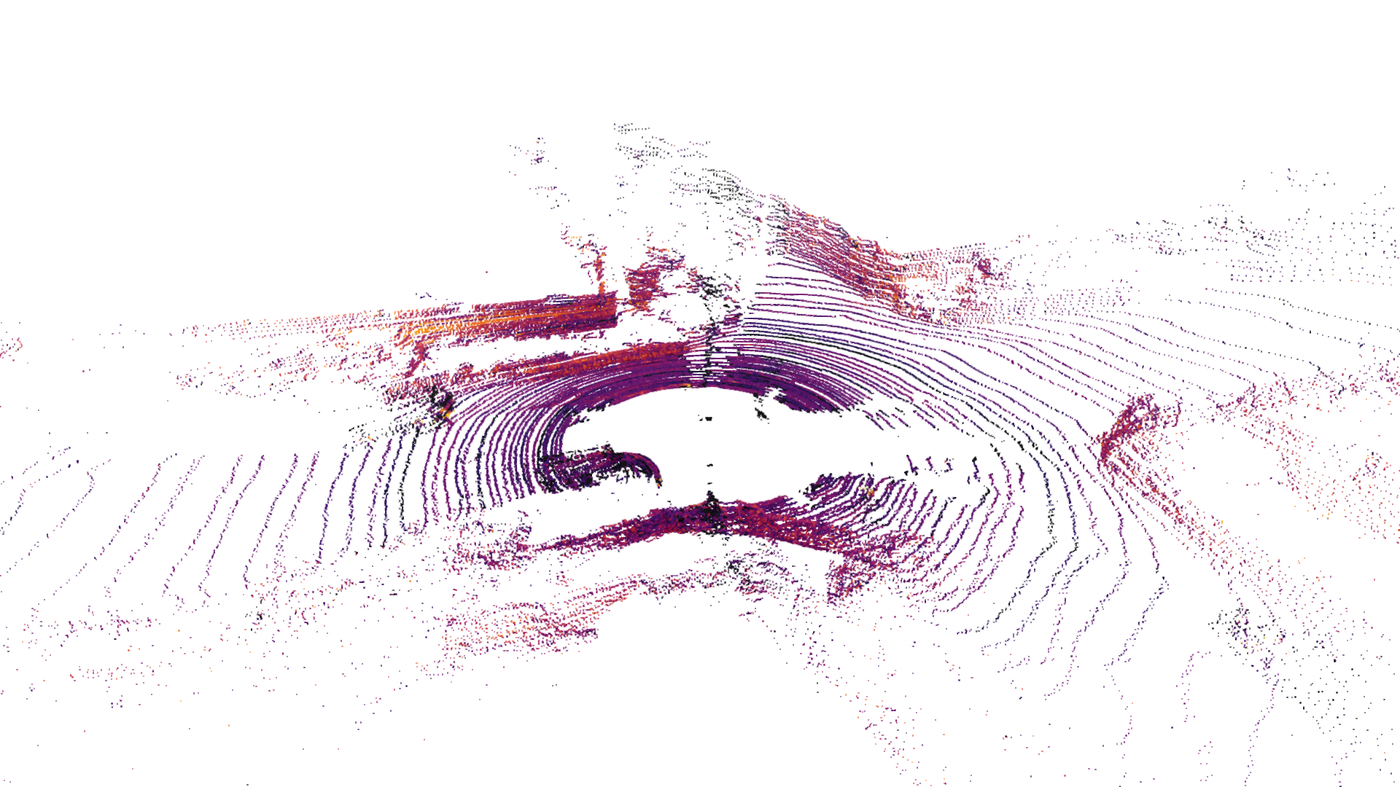} & 
\adjincludegraphics[width=0.24\textwidth, trim={{.2\width} {.2\height} {.2\width} {.2\height}},clip]{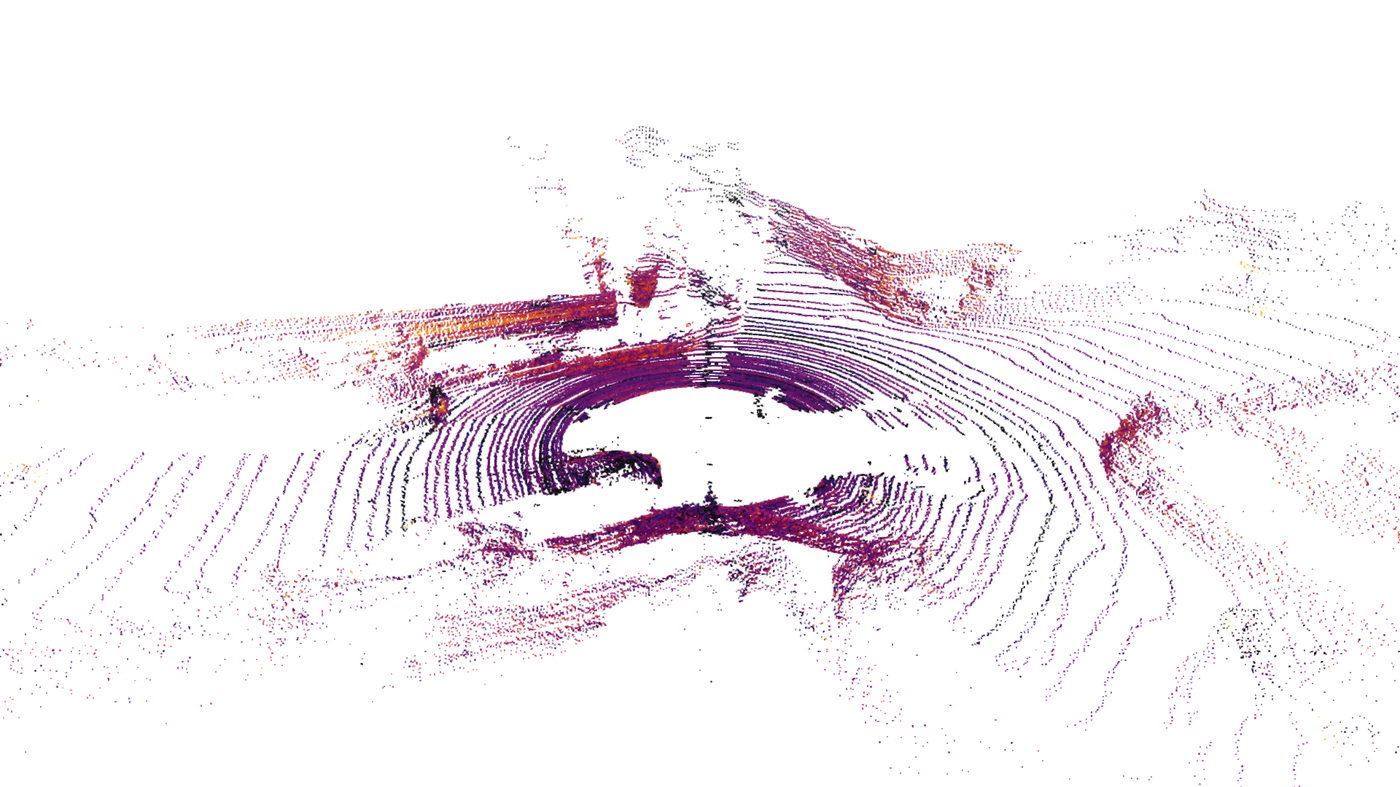} & 
\adjincludegraphics[width=0.24\textwidth, trim={{.2\width} {.2\height} {.2\width} {.2\height}},clip]{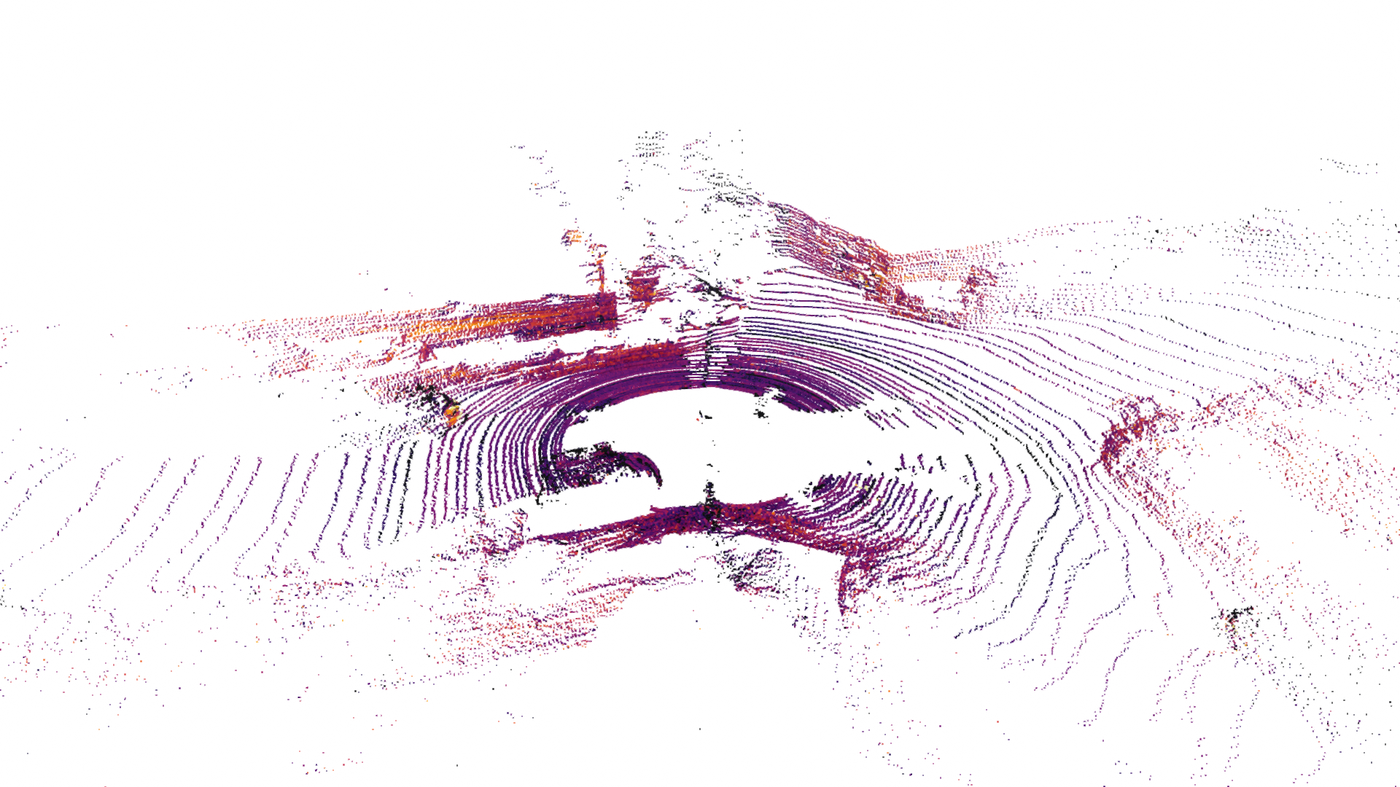} \\
Input & Sample 4  & Sample 5  & Sample 6 \\
\adjincludegraphics[width=0.24\textwidth, trim={{.2\width} {.2\height} {.2\width} {.2\height}},clip]{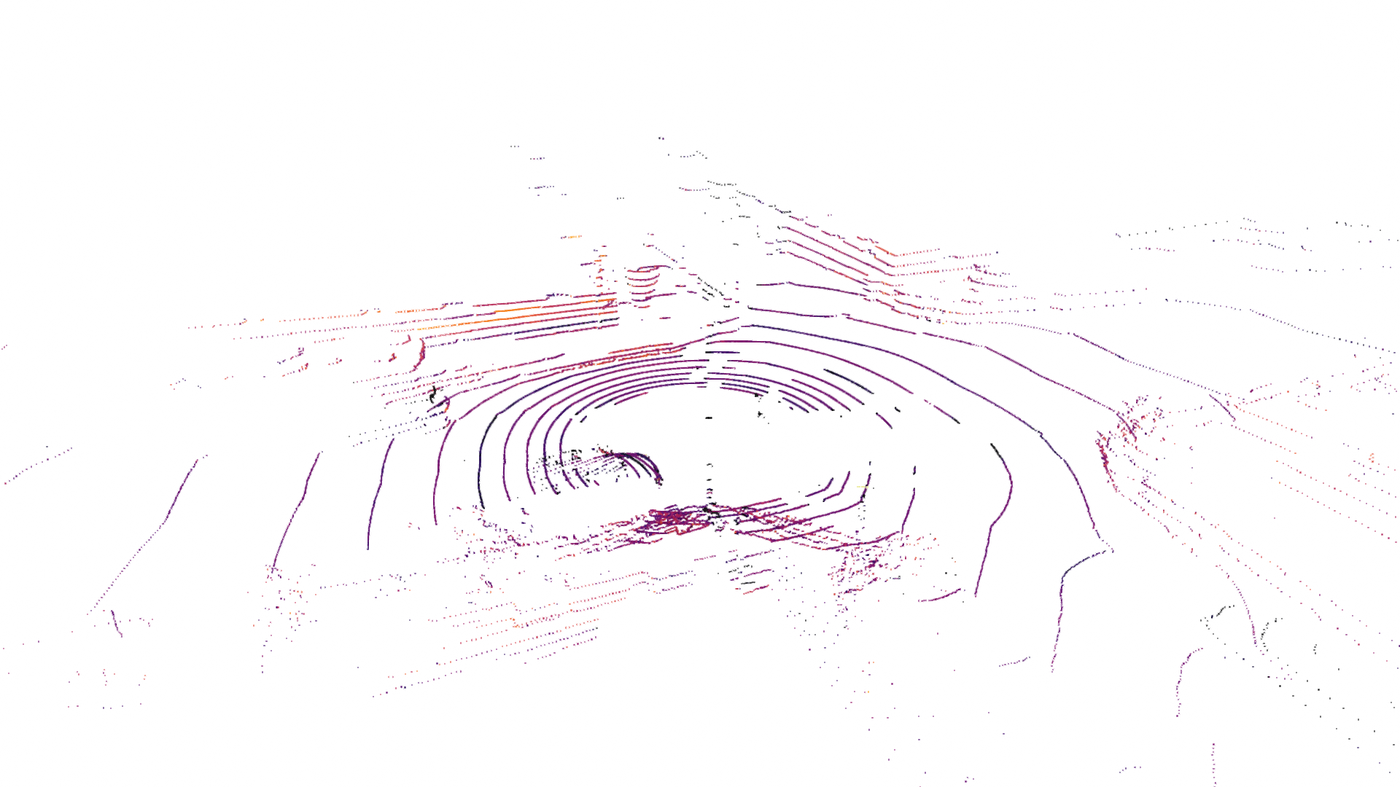} & 
\adjincludegraphics[width=0.24\textwidth, trim={{.2\width} {.2\height} {.2\width} {.2\height}},clip]{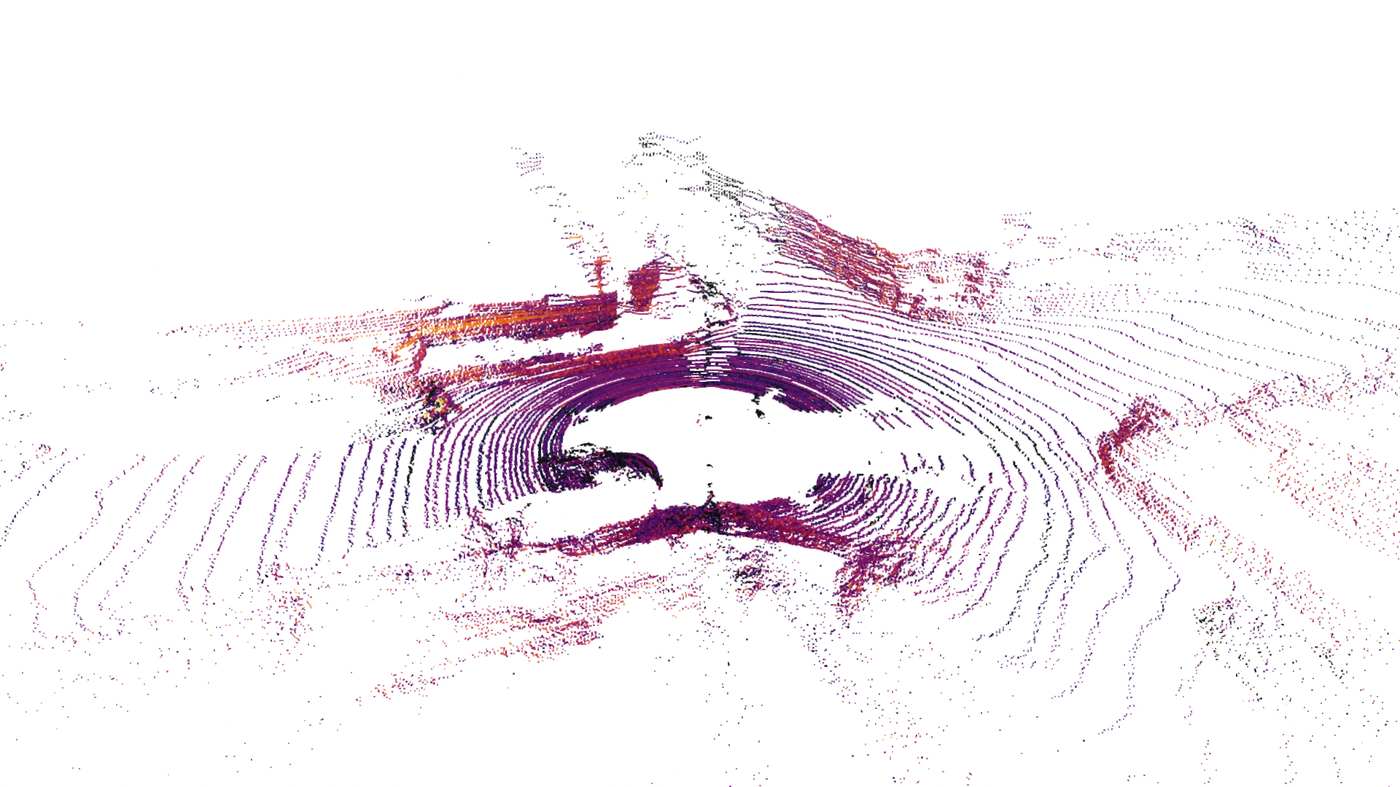} & 
\adjincludegraphics[width=0.24\textwidth, trim={{.2\width} {.2\height} {.2\width} {.2\height}},clip]{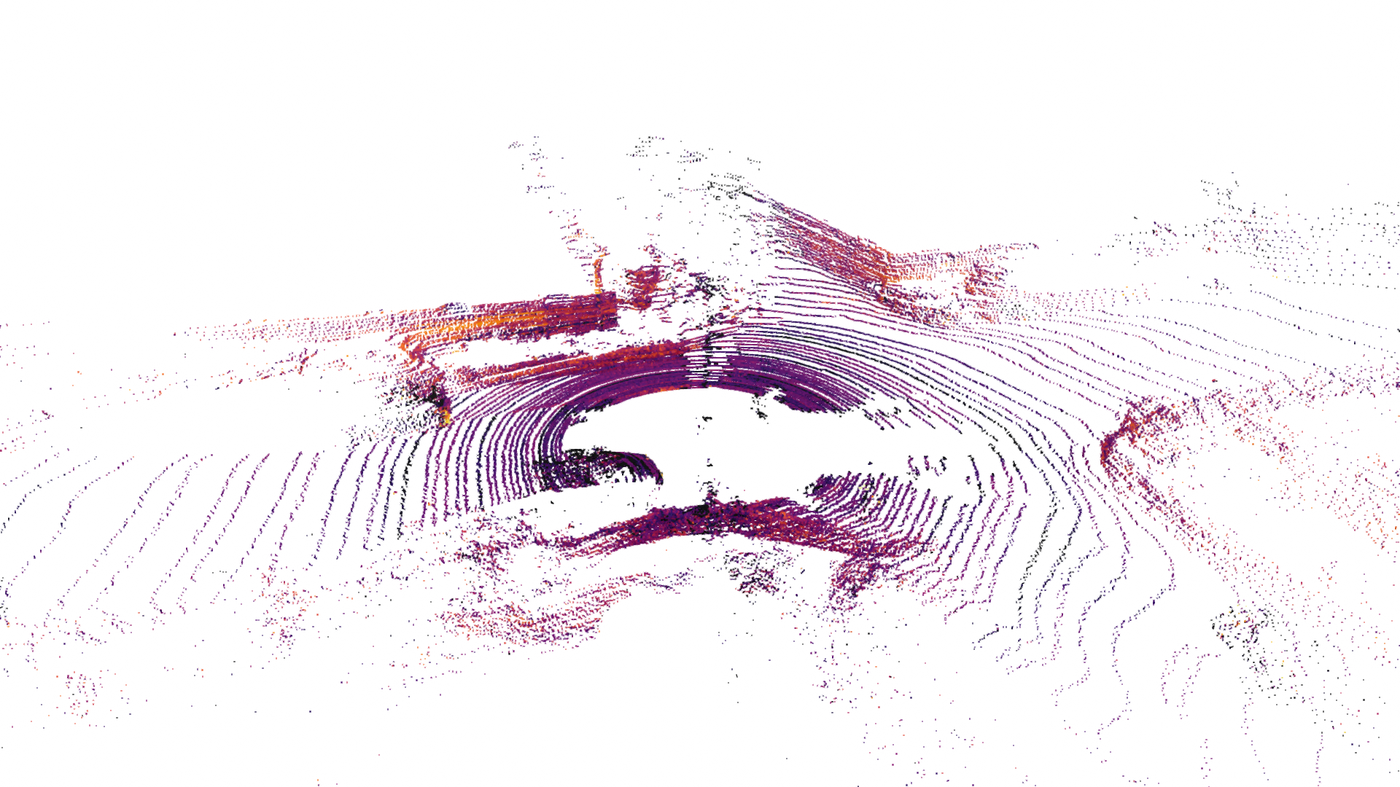} & 
\adjincludegraphics[width=0.24\textwidth, trim={{.2\width} {.2\height} {.2\width} {.2\height}},clip]{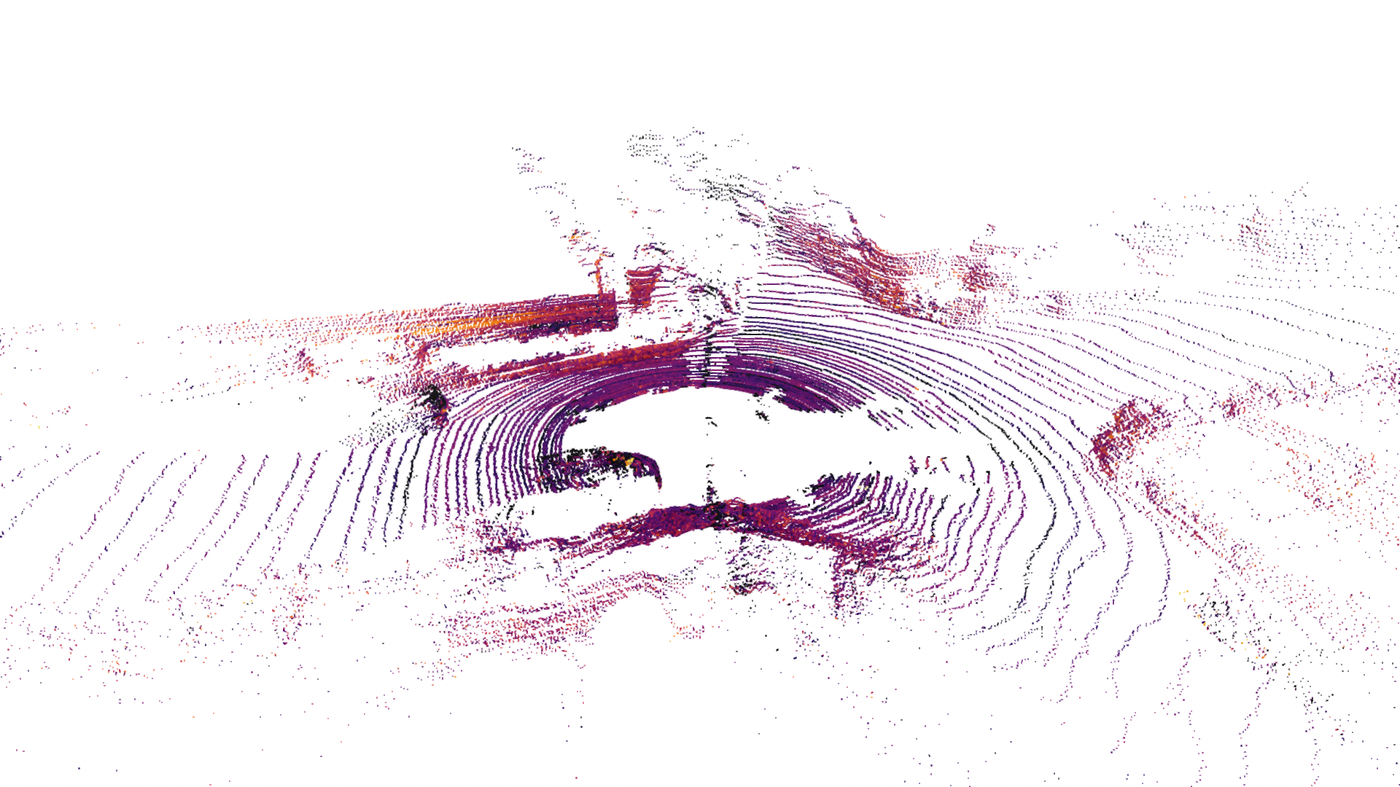} \\
\end{tabular}
\captionof{figure}{Multiple samples of posterior sampling conditioned on the same sparse input. Notice the diversity of the shapes and intensity values of the car on the left, as well as the structure of the wall.}
\label{fig:diverse}
\end{table}

\begin{table}[!t]
\small
\centering
\begin{tabular}{ccc}
Reference & 16-Beam Input & \textbf{Ours} \\
\adjincludegraphics[width=0.33\textwidth, trim={{.2\width} {.2\height} {.2\width} {.2\height}},clip]{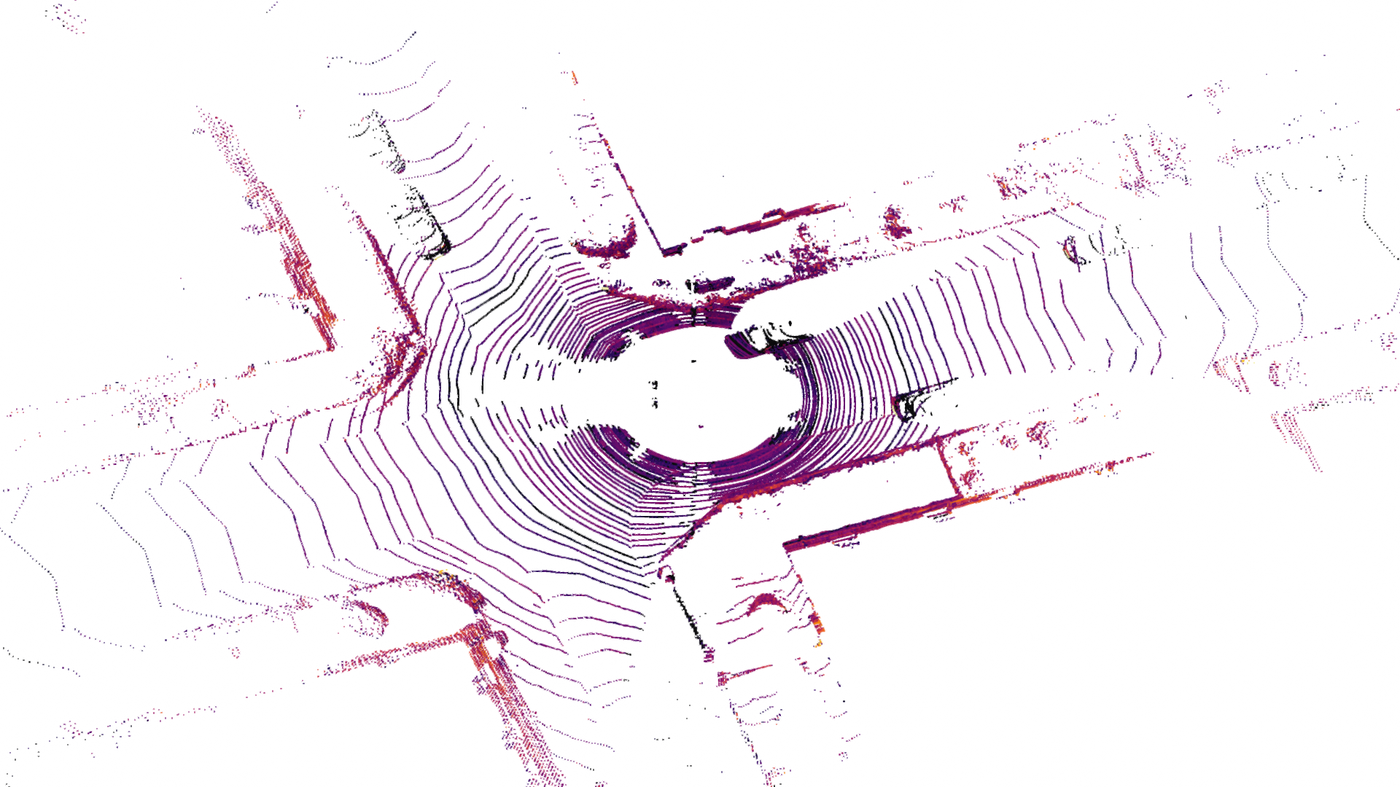} & 
\adjincludegraphics[width=0.33\textwidth, trim={{.2\width} {.2\height} {.2\width} {.2\height}},clip]{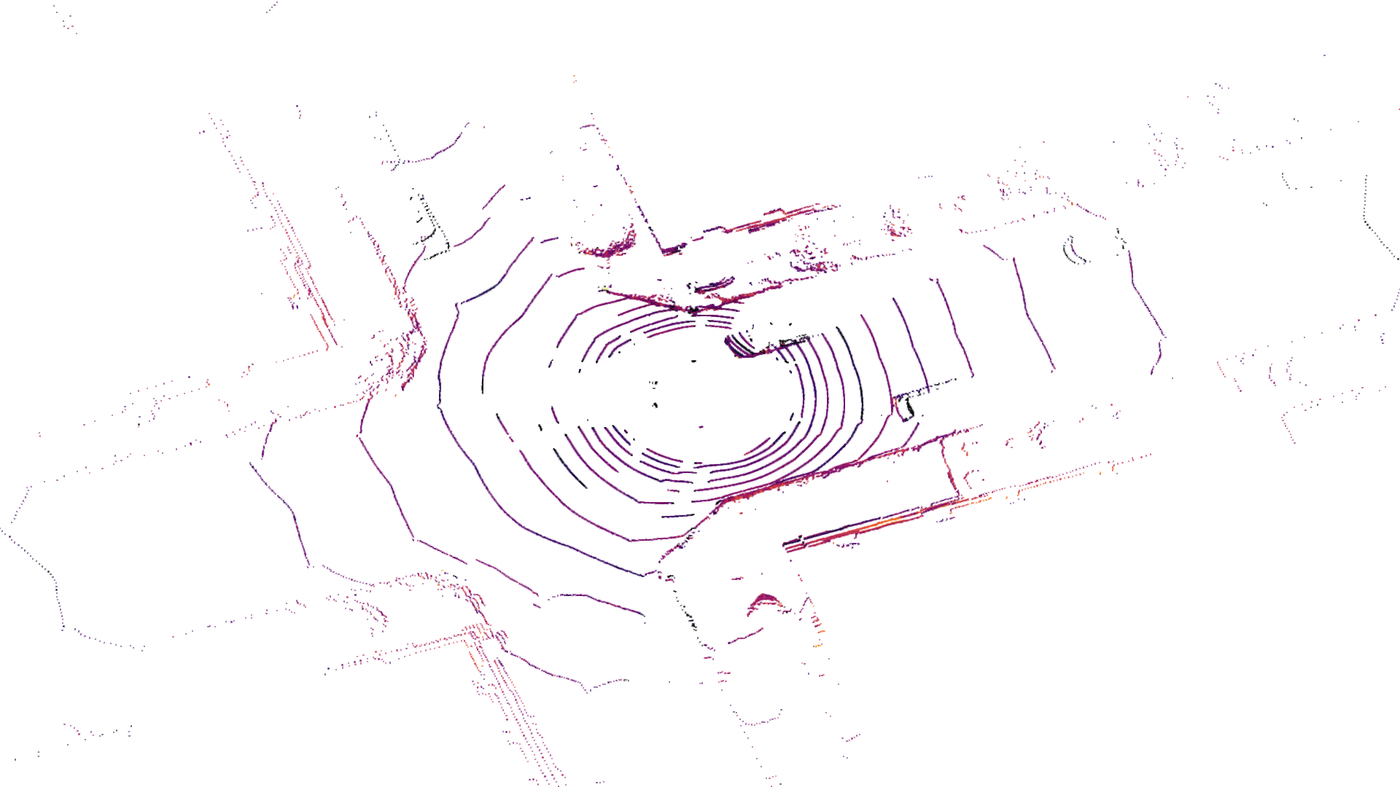} & 
\adjincludegraphics[width=0.33\textwidth, trim={{.2\width} {.2\height} {.2\width} {.2\height}},clip]{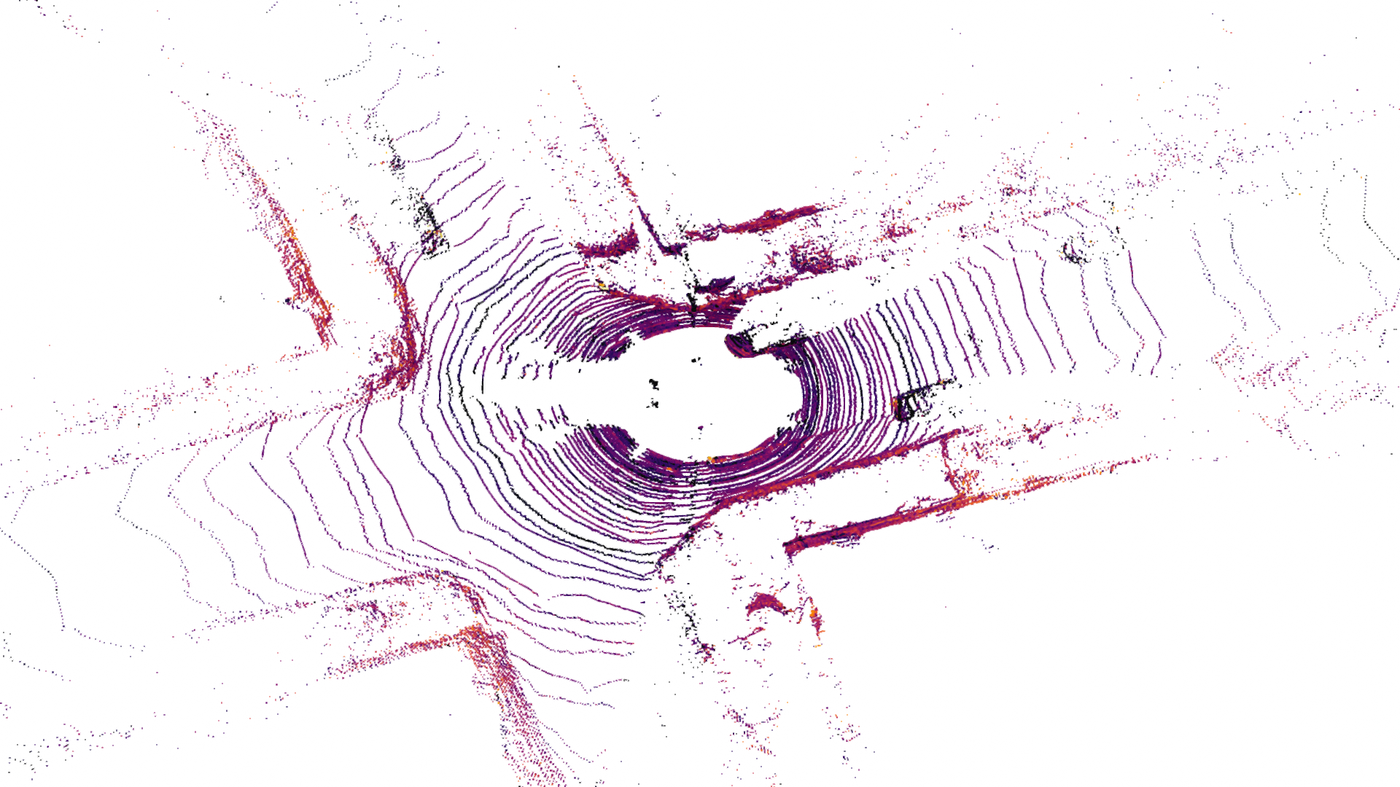} \\
\adjincludegraphics[width=0.33\textwidth, trim={{.2\width} {.2\height} {.2\width} {.2\height}},clip]{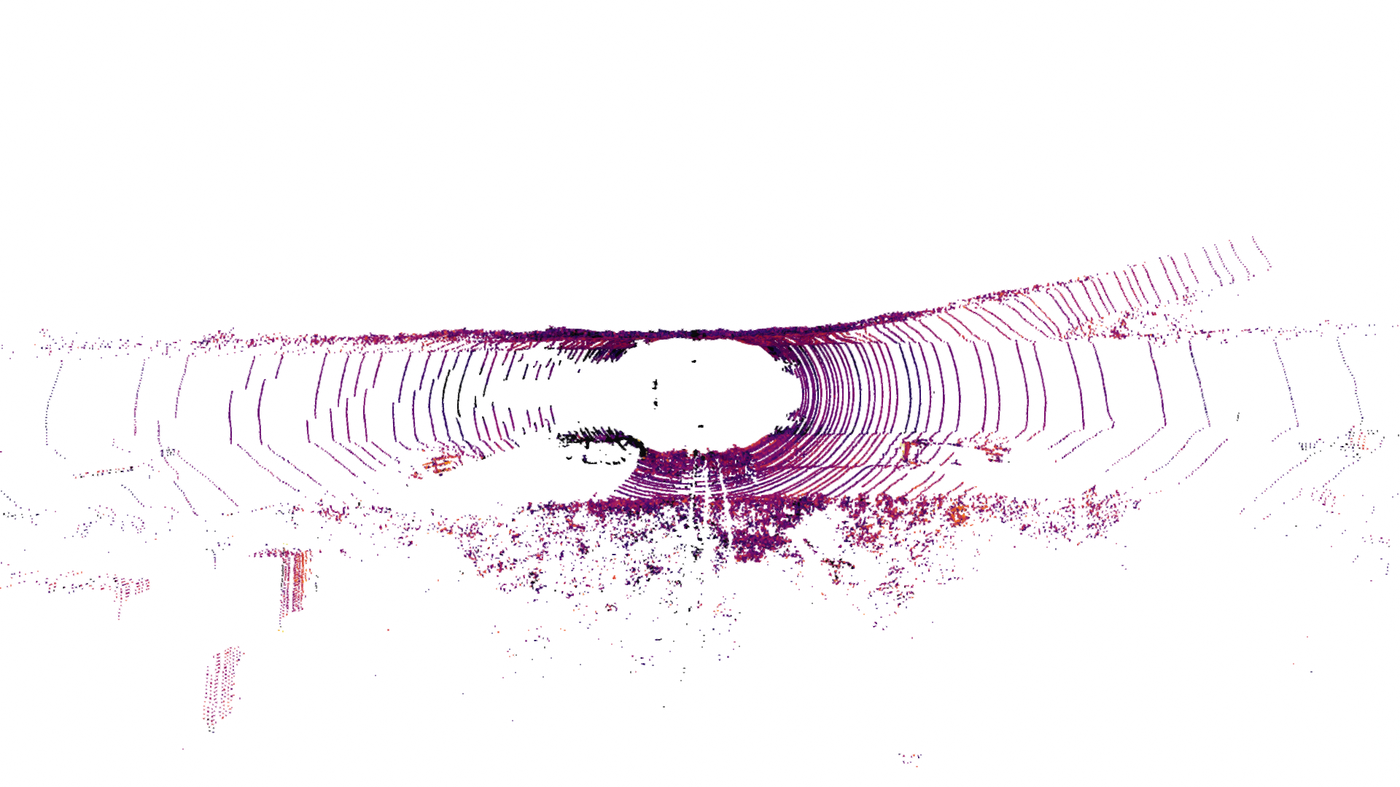} & 
\adjincludegraphics[width=0.33\textwidth, trim={{.2\width} {.2\height} {.2\width} {.2\height}},clip]{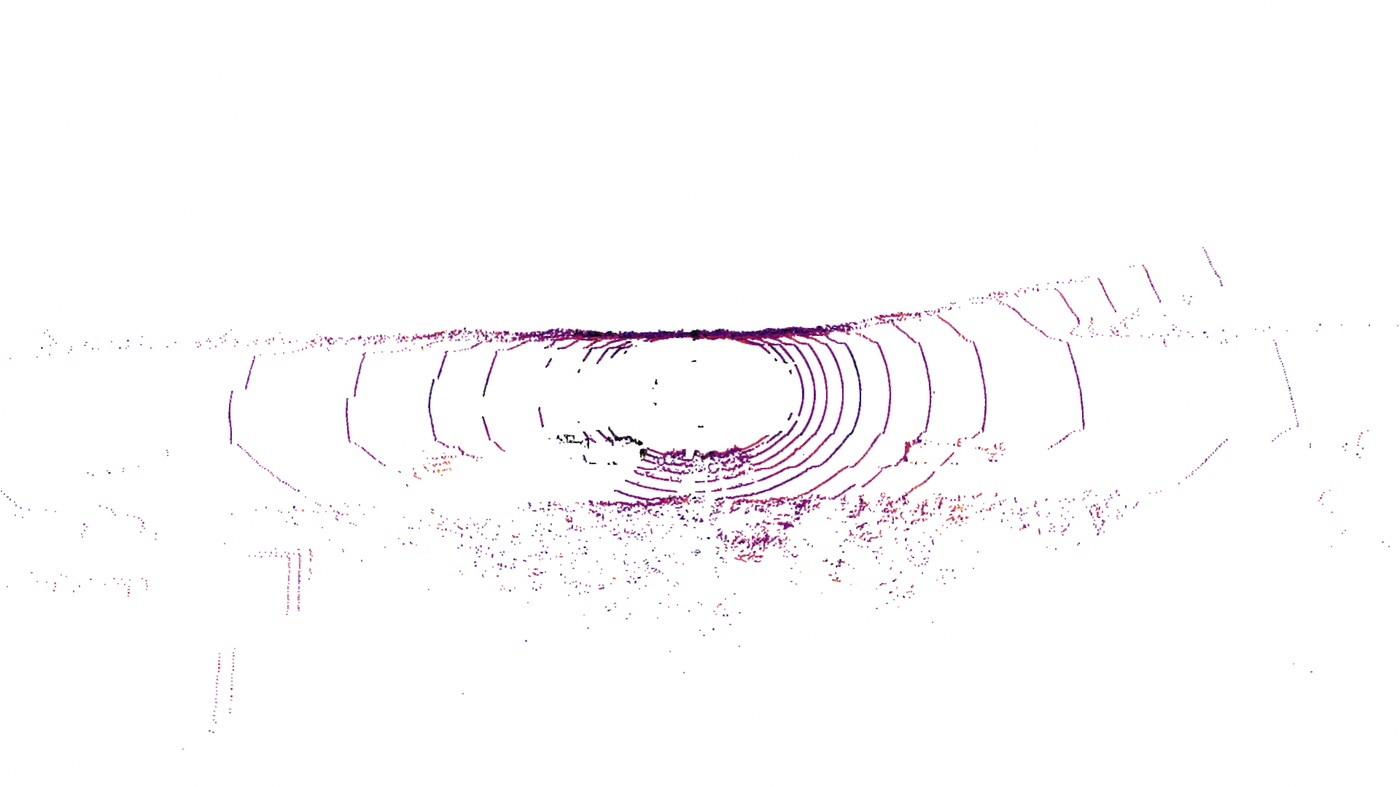} & 
\adjincludegraphics[width=0.33\textwidth, trim={{.2\width} {.2\height} {.2\width} {.2\height}},clip]{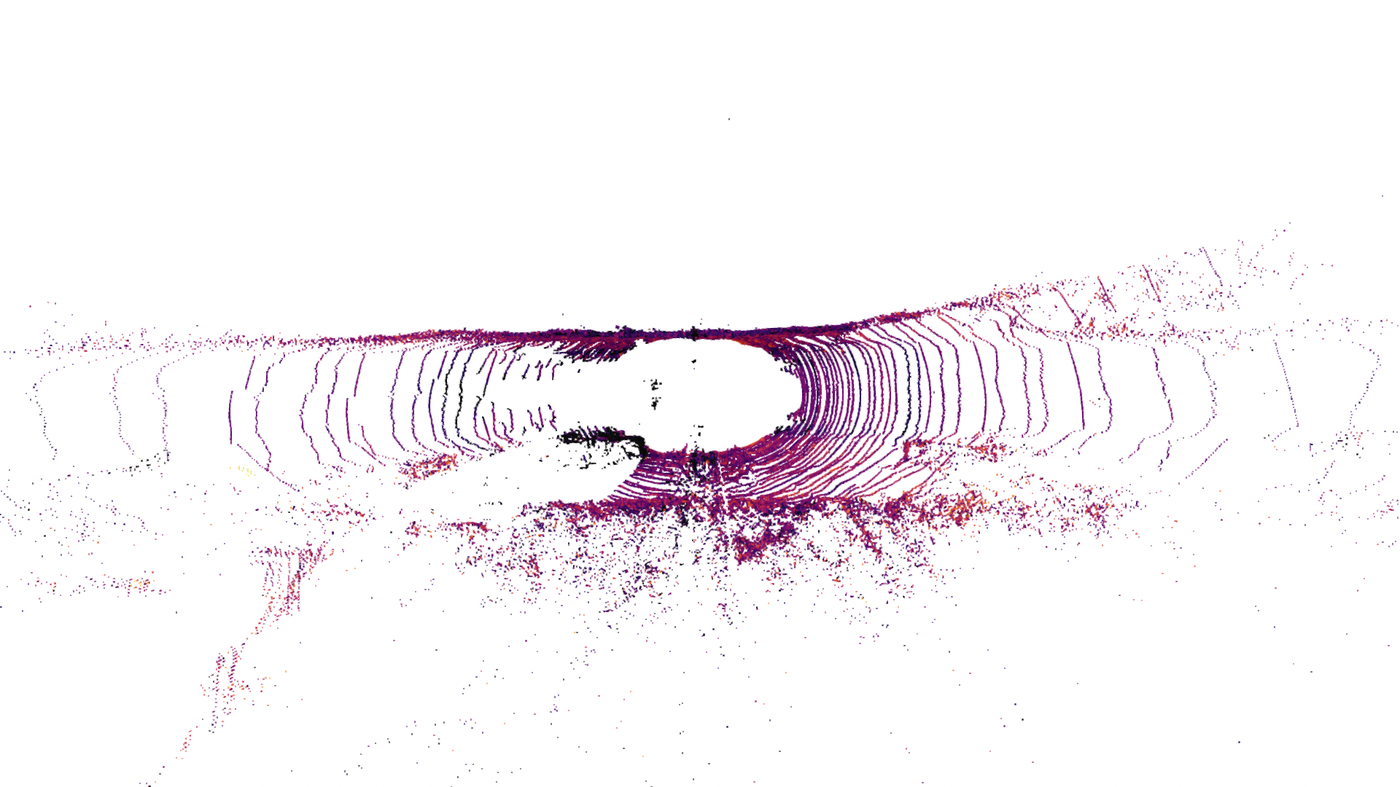} \\
\adjincludegraphics[width=0.33\textwidth, trim={{.2\width} {.2\height} {.2\width} {.2\height}},clip]{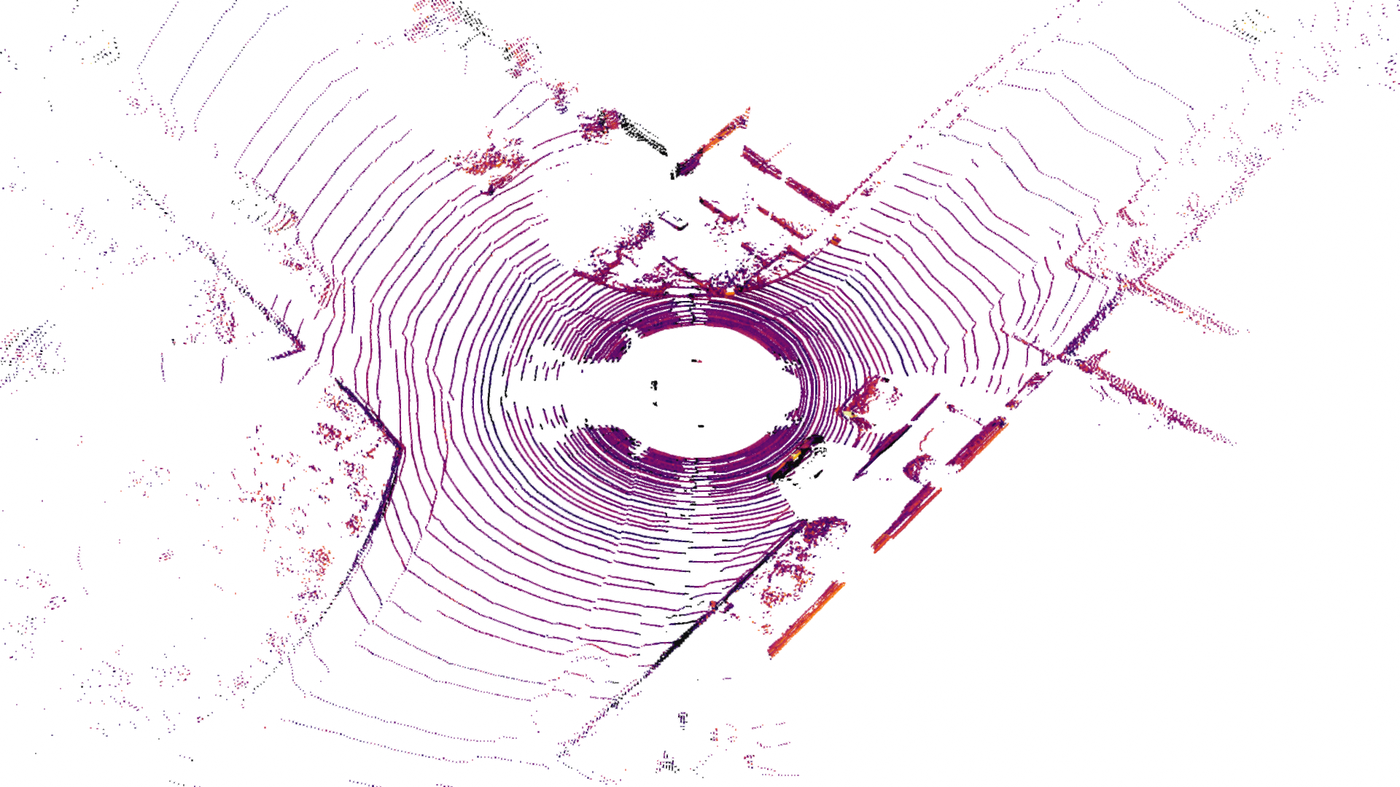} & 
\adjincludegraphics[width=0.33\textwidth, trim={{.2\width} {.2\height} {.2\width} {.2\height}},clip]{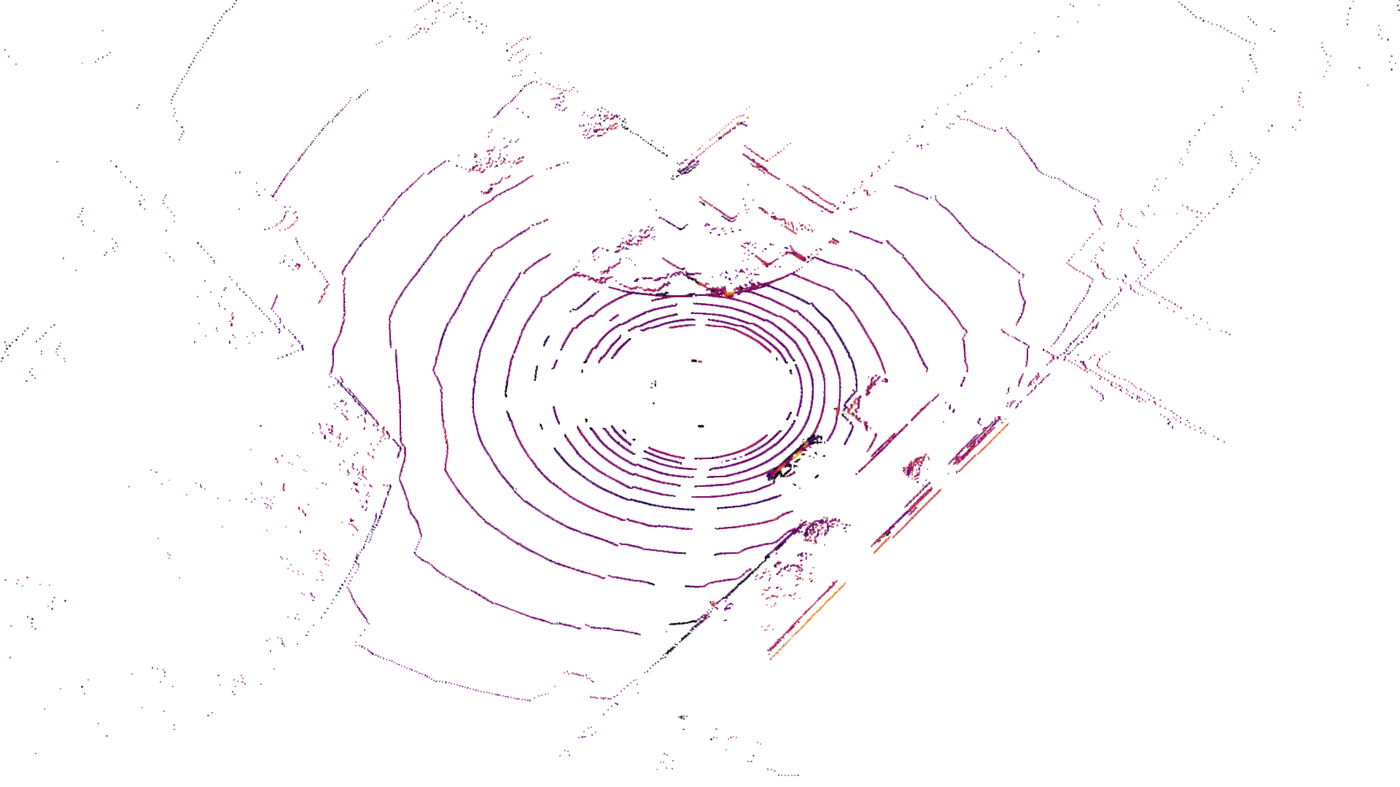} & 
\adjincludegraphics[width=0.33\textwidth, trim={{.2\width} {.2\height} {.2\width} {.2\height}},clip]{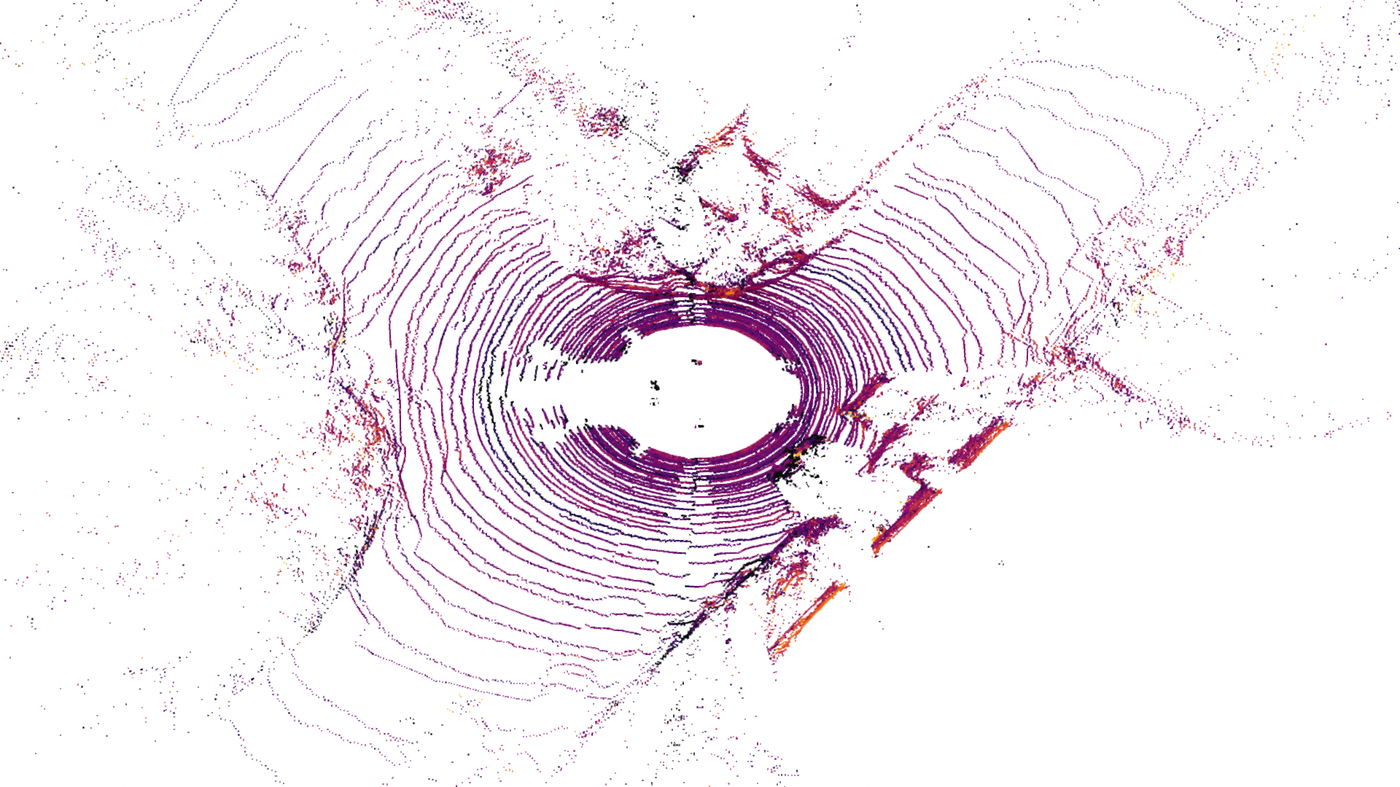} \\
\adjincludegraphics[width=0.33\textwidth, trim={{.2\width} {.2\height} {.2\width} {.2\height}},clip]{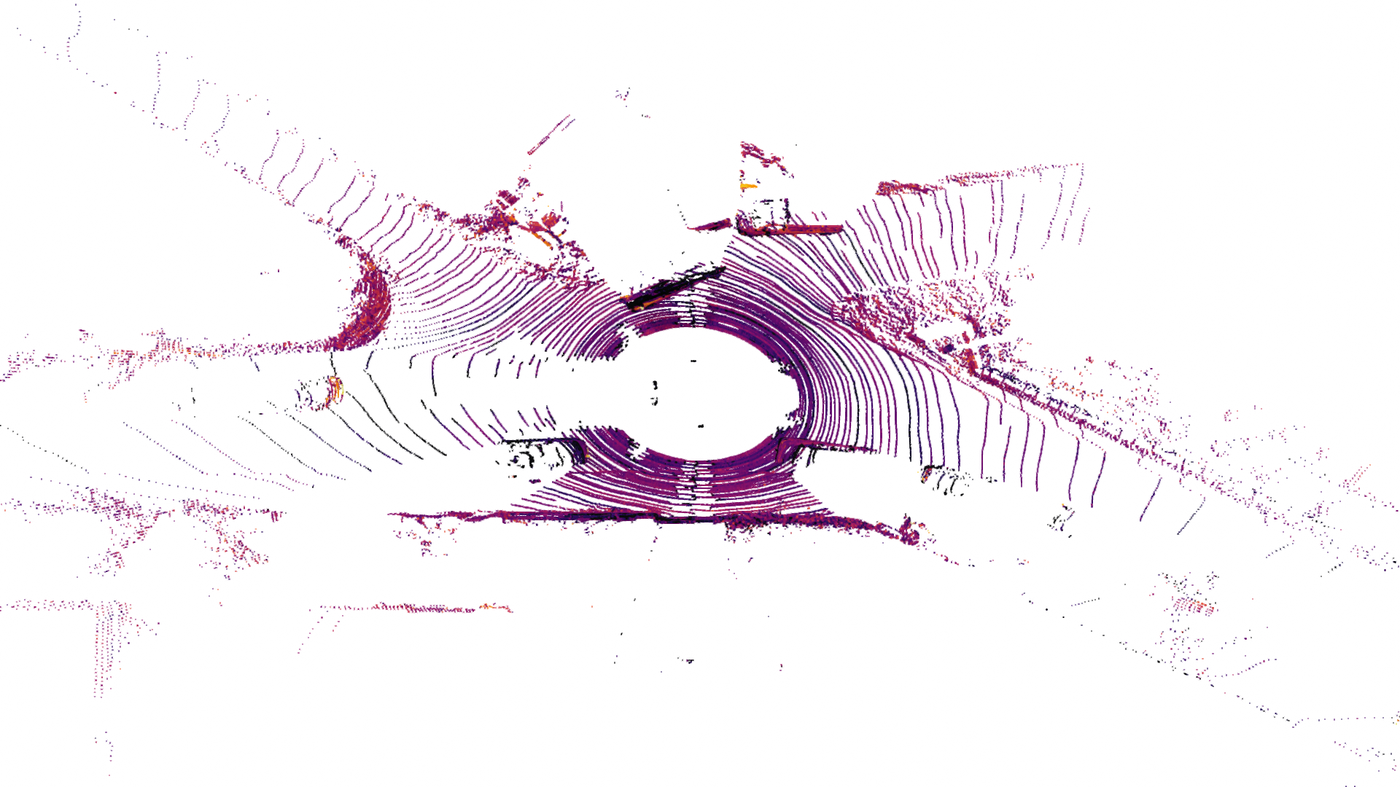} & 
\adjincludegraphics[width=0.33\textwidth, trim={{.2\width} {.2\height} {.2\width} {.2\height}},clip]{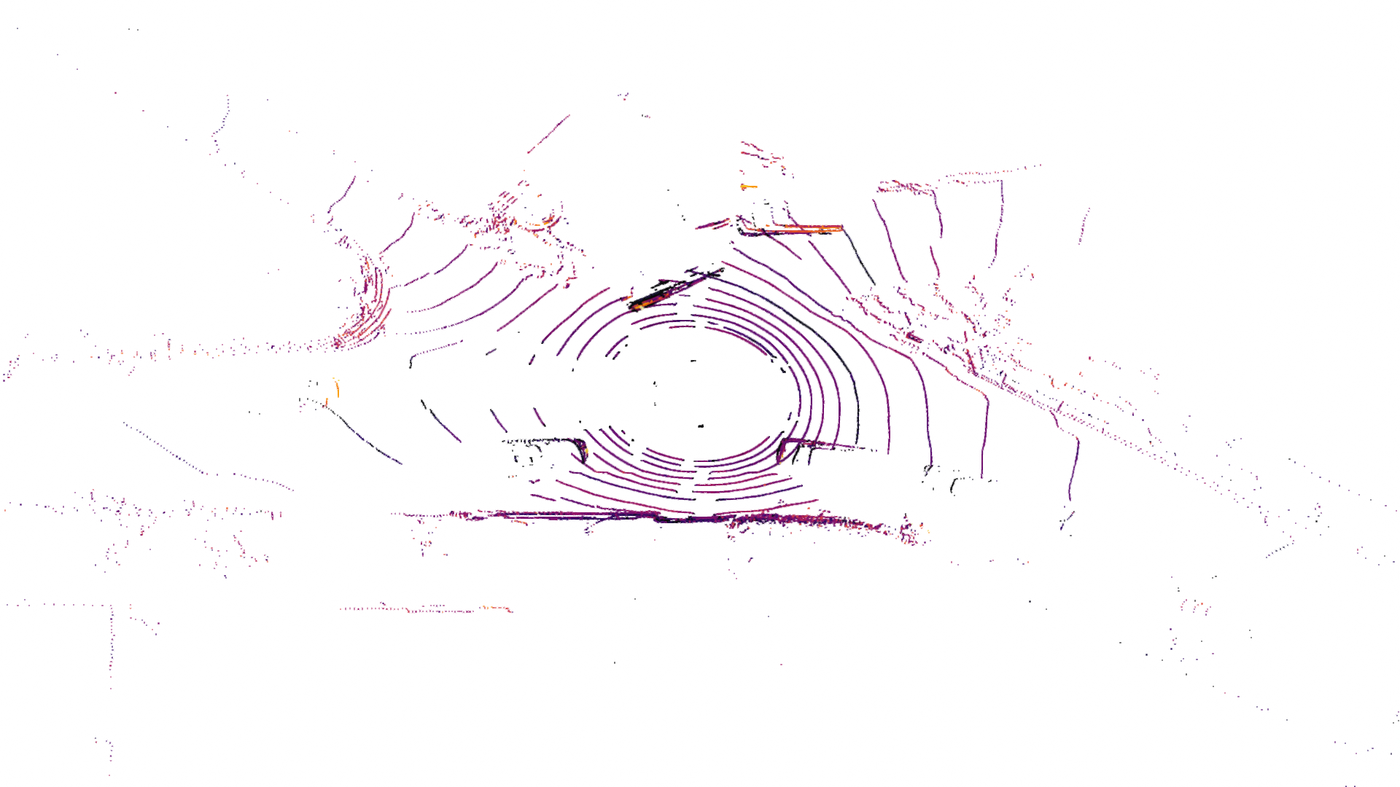} & 
\adjincludegraphics[width=0.33\textwidth, trim={{.2\width} {.2\height} {.2\width} {.2\height}},clip]{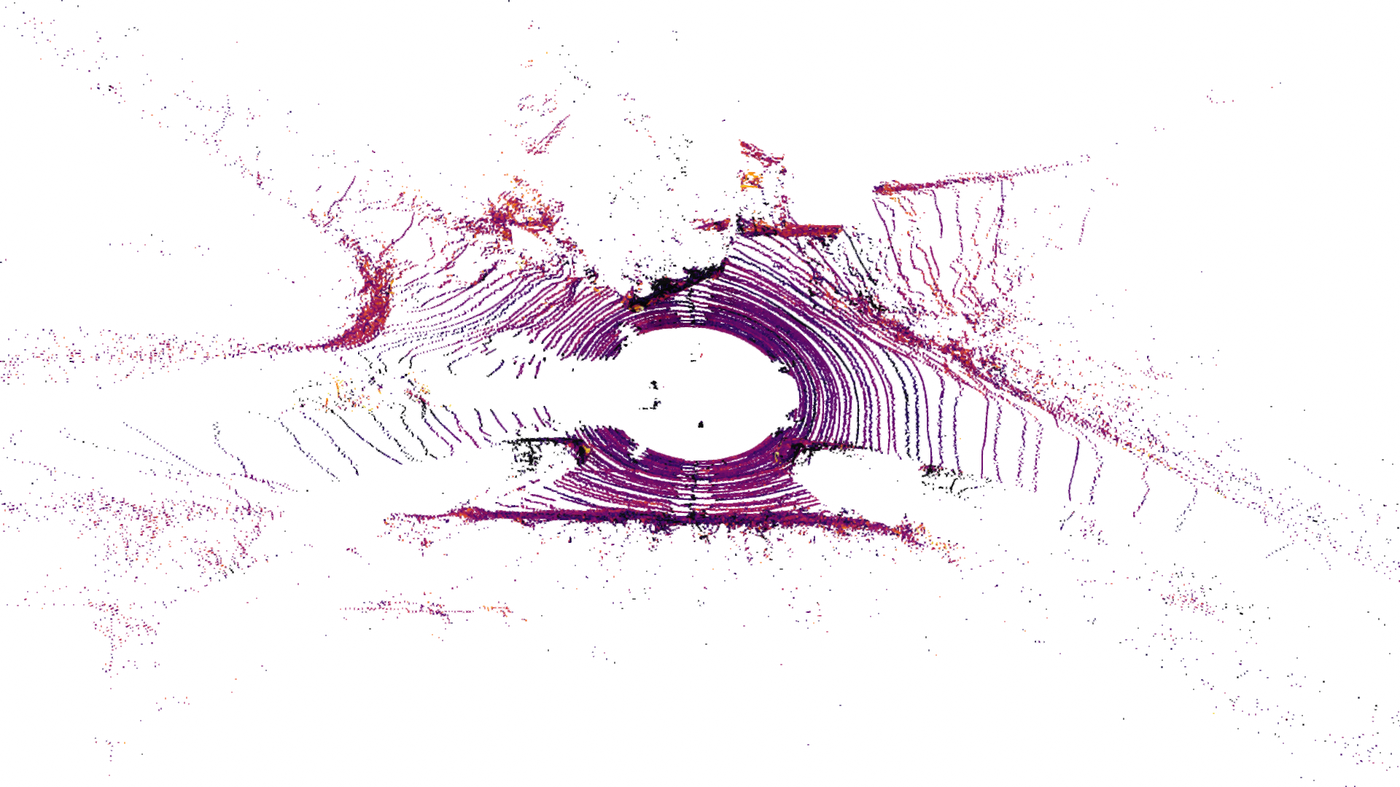} \\
\adjincludegraphics[width=0.33\textwidth, trim={{.2\width} {.2\height} {.2\width} {.2\height}},clip]{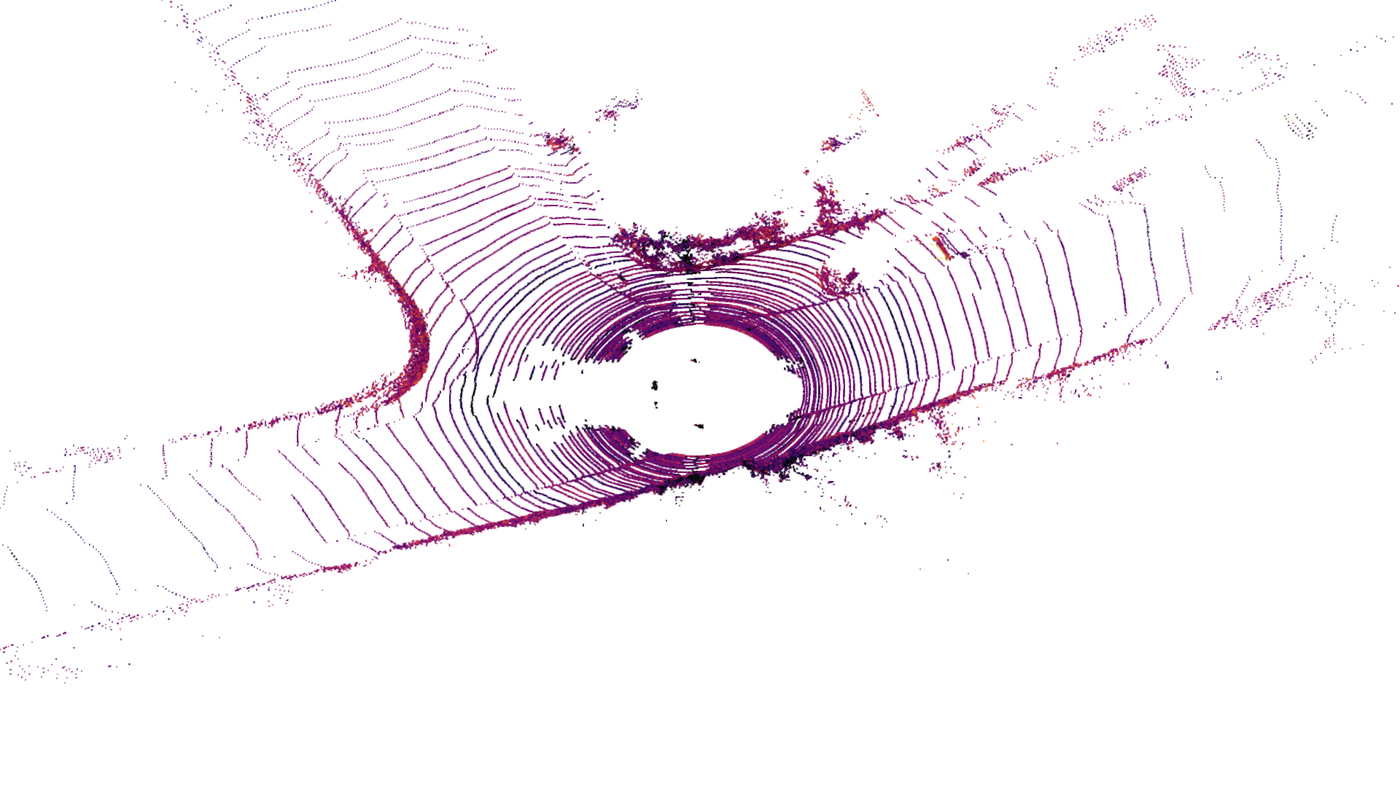} & 
\adjincludegraphics[width=0.33\textwidth, trim={{.2\width} {.2\height} {.2\width} {.2\height}},clip]{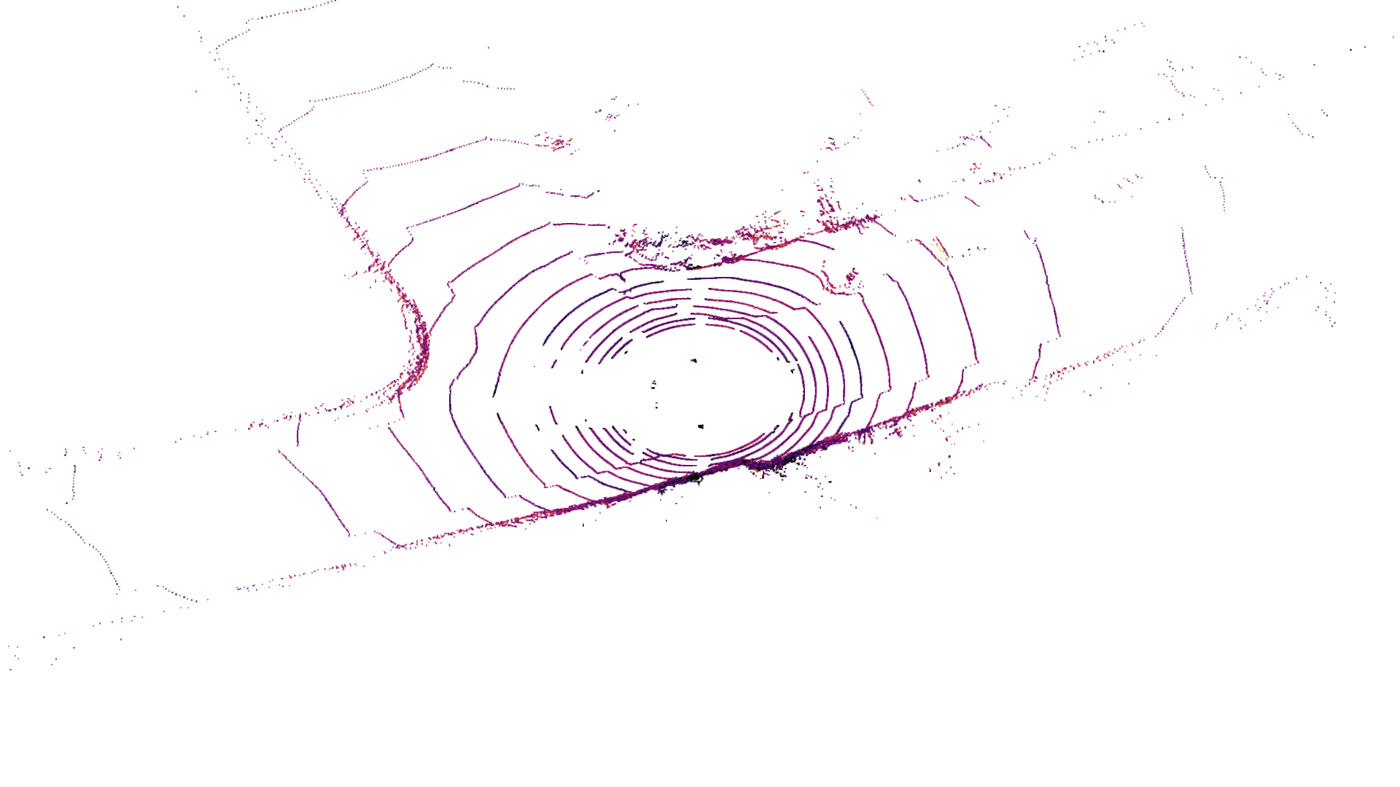} & 
\adjincludegraphics[width=0.33\textwidth, trim={{.2\width} {.2\height} {.2\width} {.2\height}},clip]{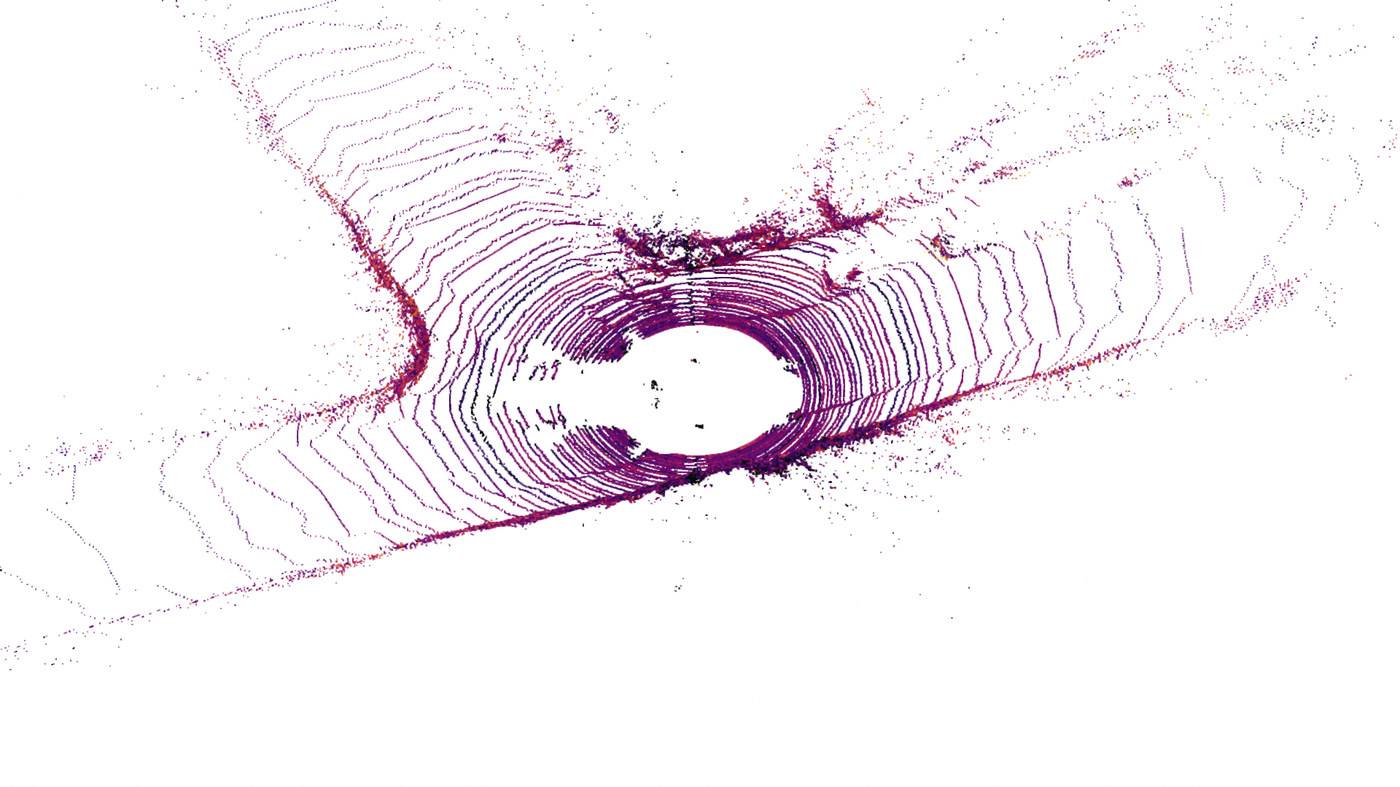} \\
\adjincludegraphics[width=0.33\textwidth, trim={{.2\width} {.2\height} {.2\width} {.2\height}},clip]{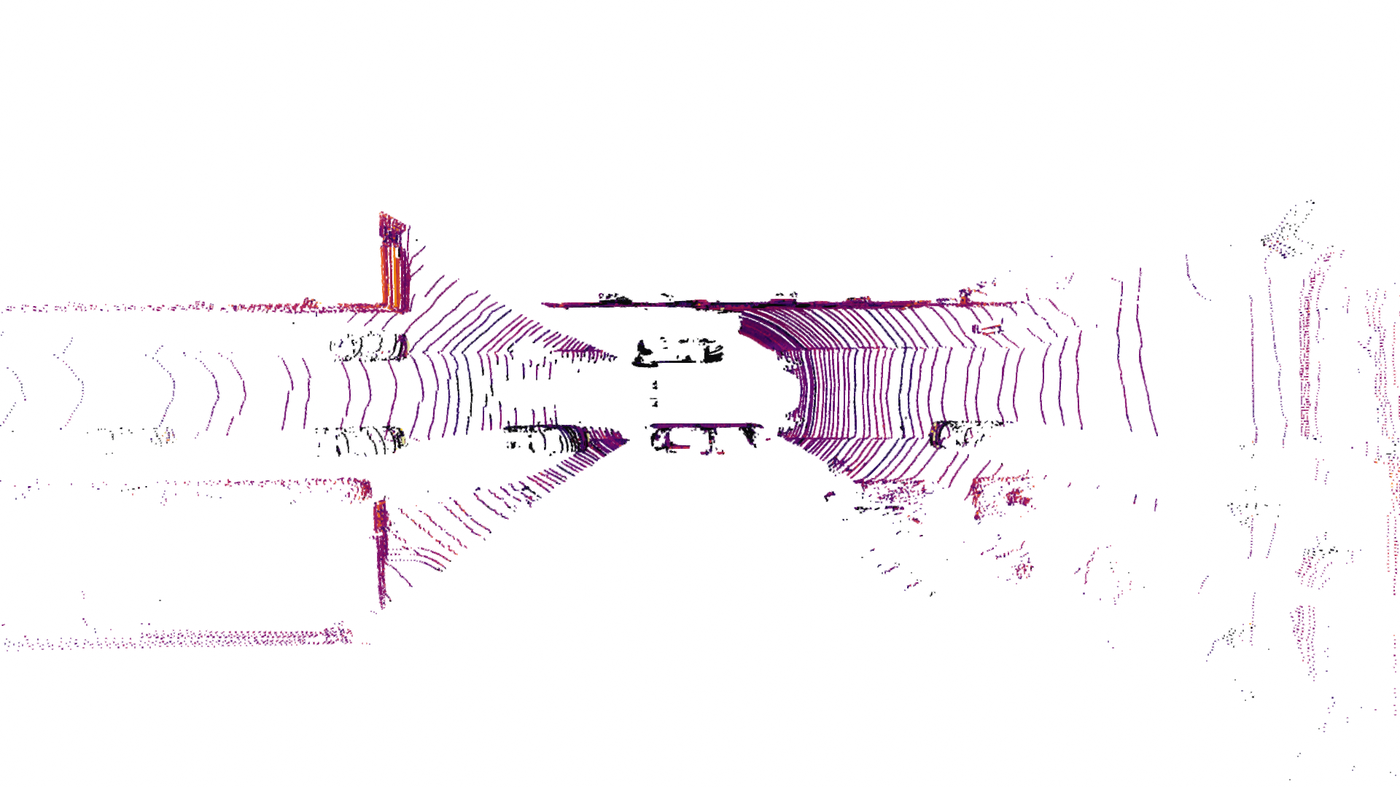} & 
\adjincludegraphics[width=0.33\textwidth, trim={{.2\width} {.2\height} {.2\width} {.2\height}},clip]{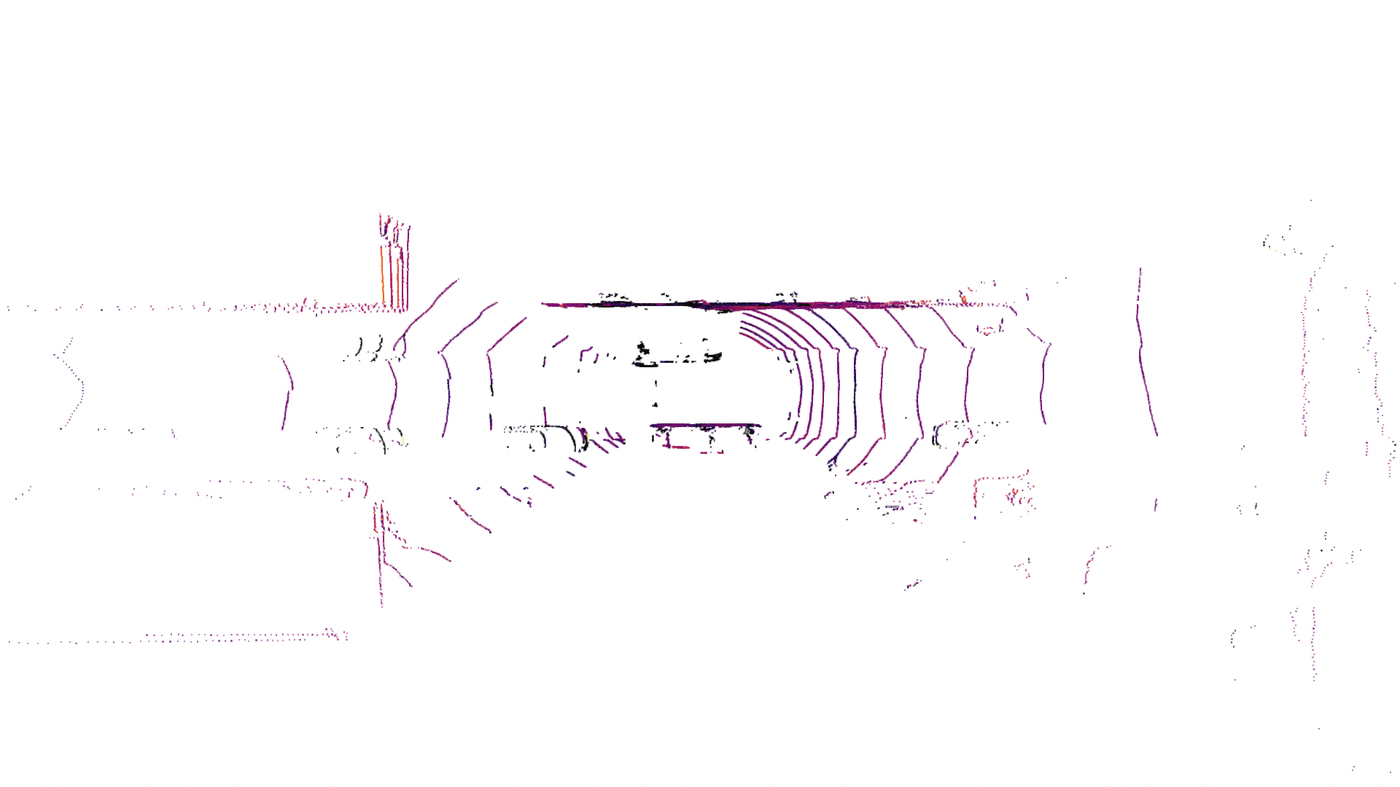} & 
\adjincludegraphics[width=0.33\textwidth, trim={{.2\width} {.2\height} {.2\width} {.2\height}},clip]{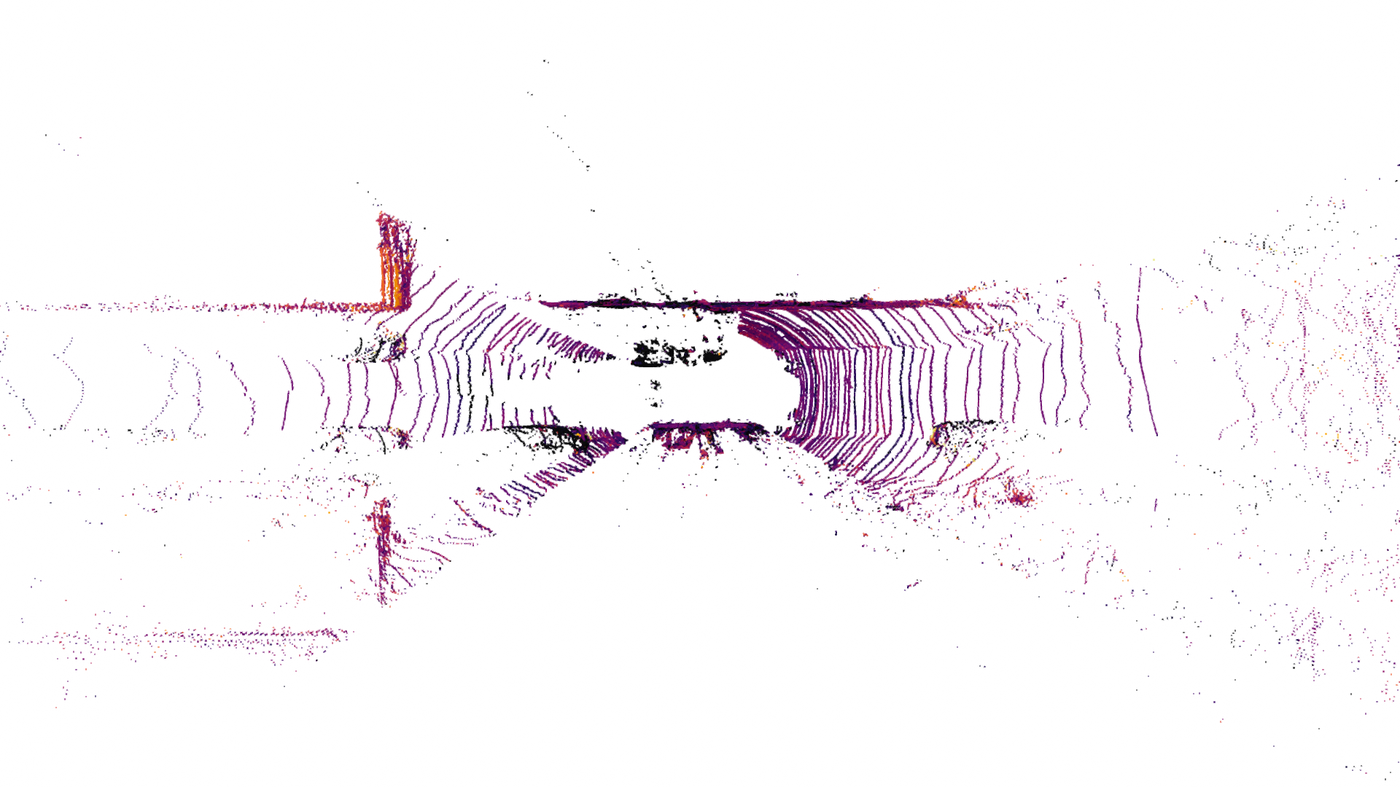} \\
\label{tab:densify}
\end{tabular}
\captionof{figure}{Additional Results for Unsupervised LiDAR Densification.}
\label{fig:more_densification}
\end{table}

\begin{table}[!t]
\small
\centering
\begin{tabular}{ccc}
Reference & 16-Beam Input & \textbf{Ours} \\
\adjincludegraphics[width=0.33\textwidth, trim={{.2\width} {.2\height} {.2\width} {.2\height}},clip]{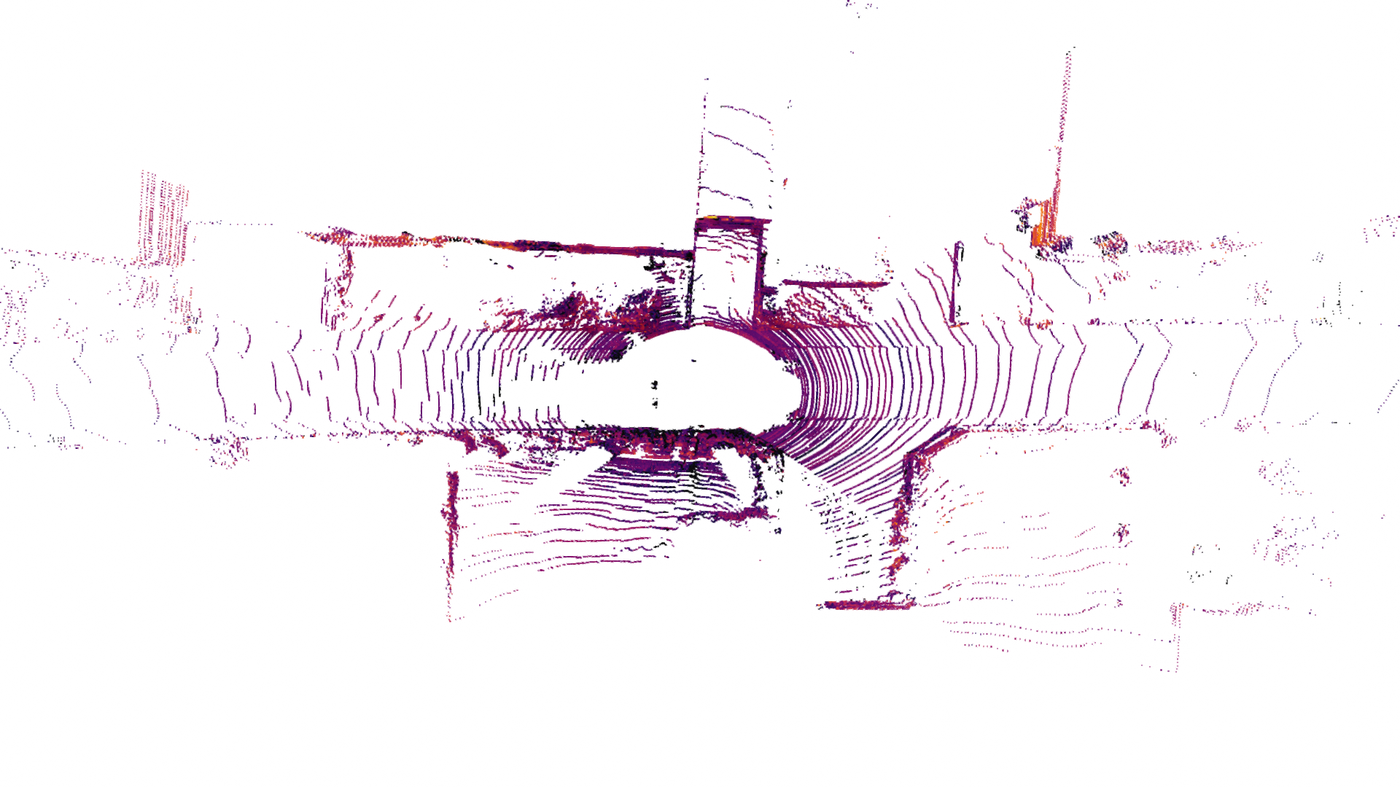} & 
\adjincludegraphics[width=0.33\textwidth, trim={{.2\width} {.2\height} {.2\width} {.2\height}},clip]{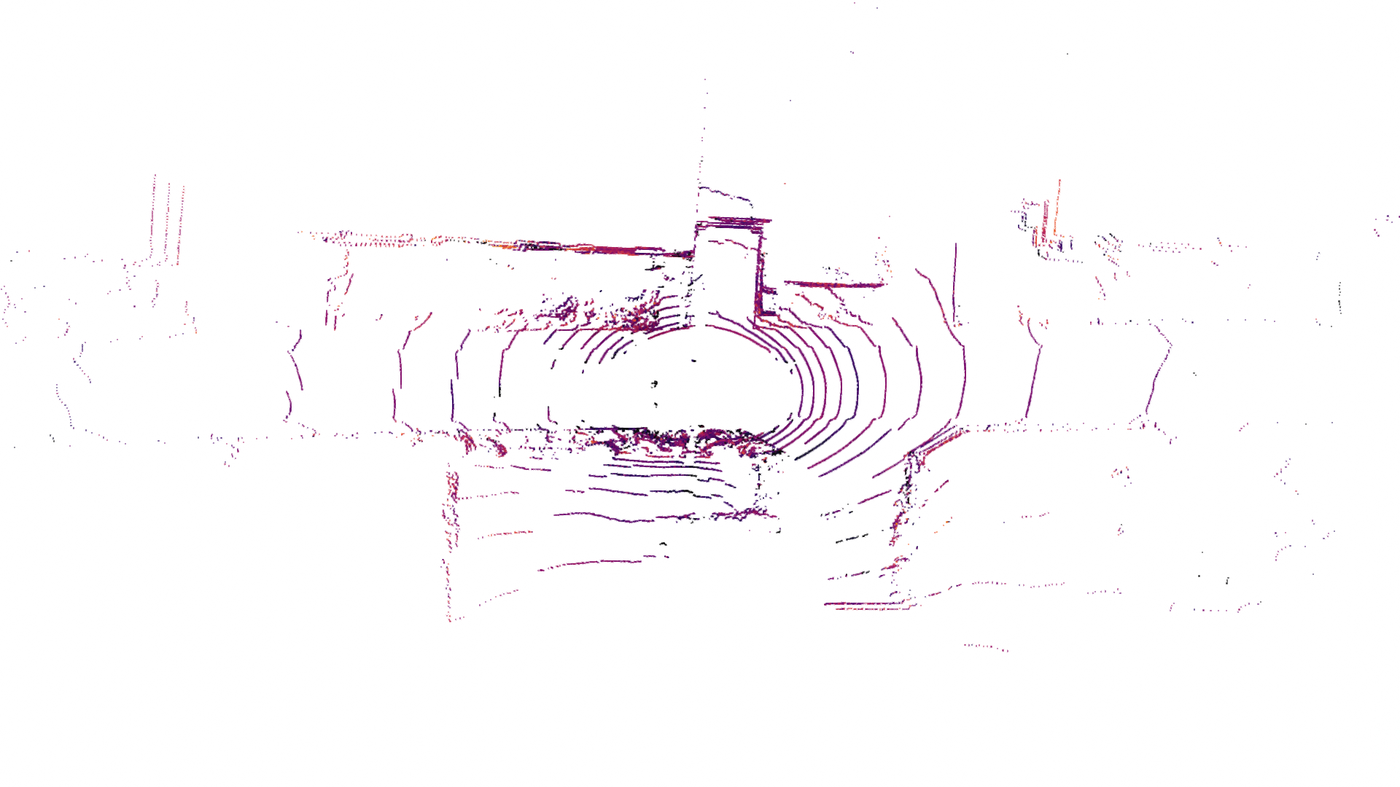} & 
\adjincludegraphics[width=0.33\textwidth, trim={{.2\width} {.2\height} {.2\width} {.2\height}},clip]{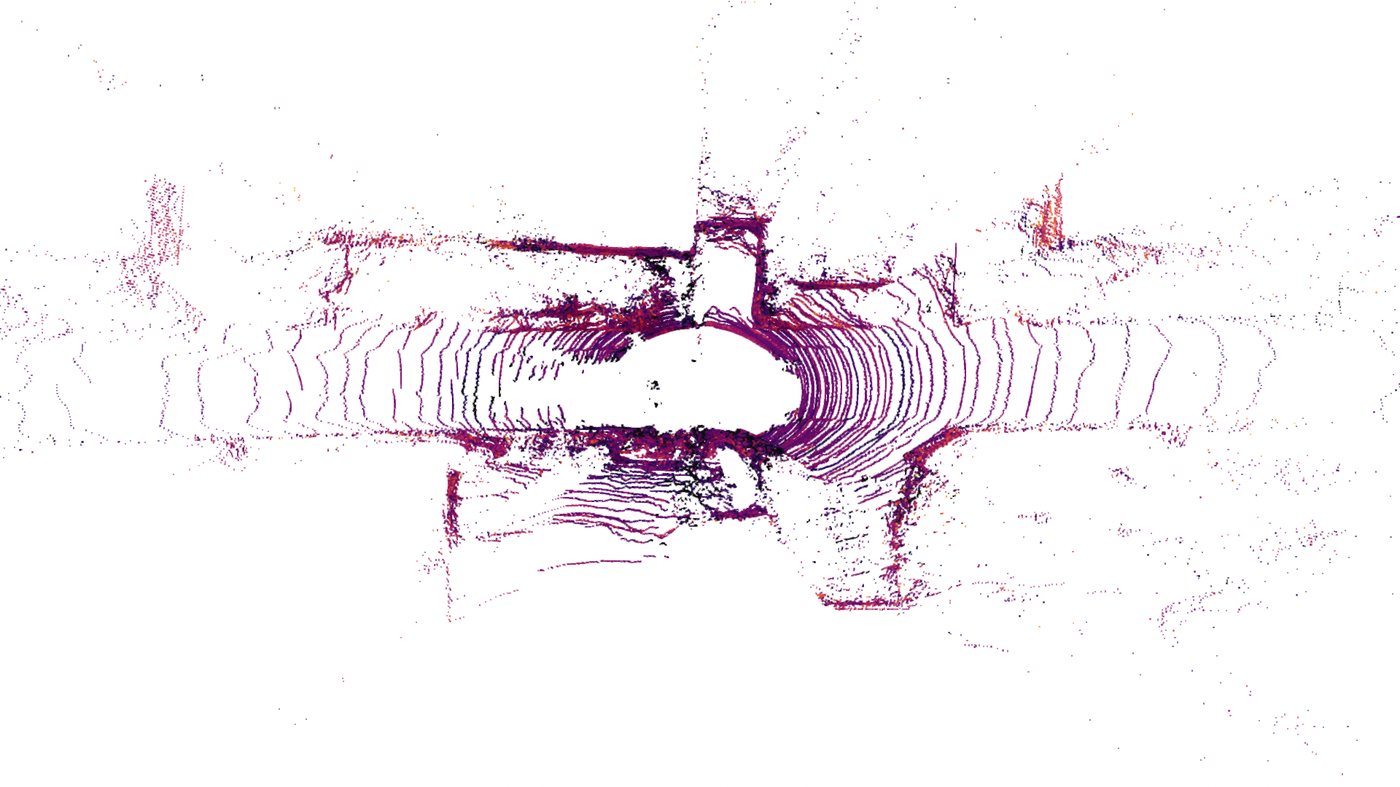} \\
\adjincludegraphics[width=0.33\textwidth, trim={{.2\width} {.2\height} {.2\width} {.2\height}},clip]{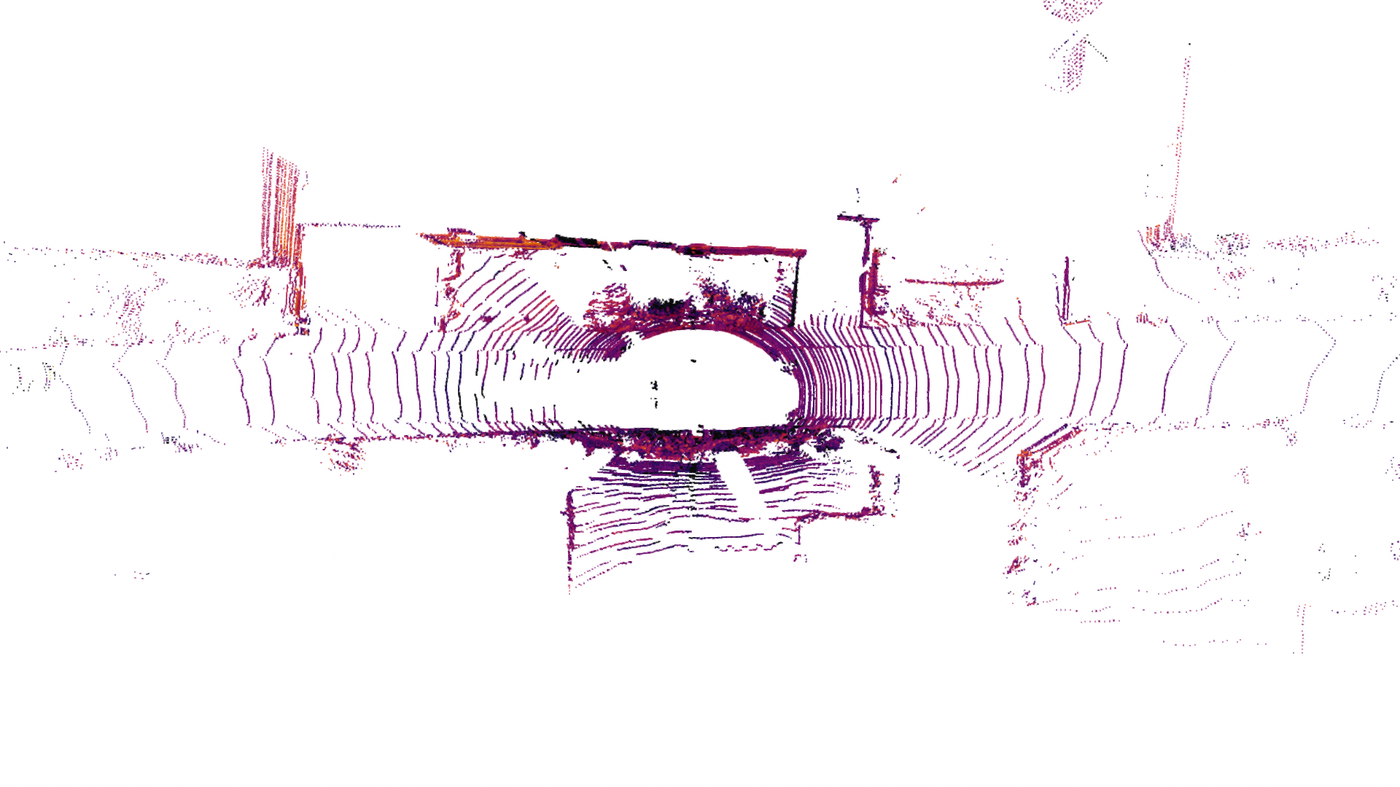} & 
\adjincludegraphics[width=0.33\textwidth, trim={{.2\width} {.2\height} {.2\width} {.2\height}},clip]{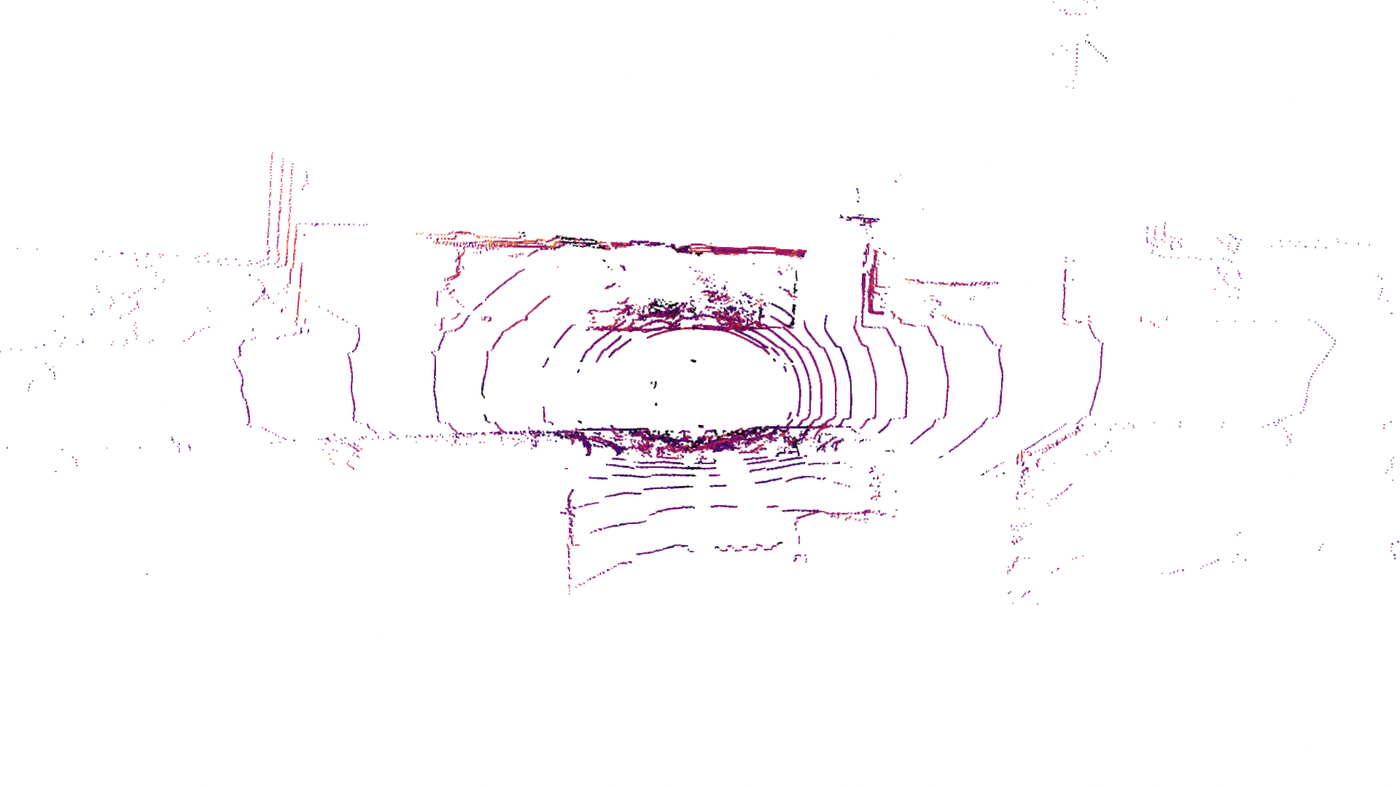} & 
\adjincludegraphics[width=0.33\textwidth, trim={{.2\width} {.2\height} {.2\width} {.2\height}},clip]{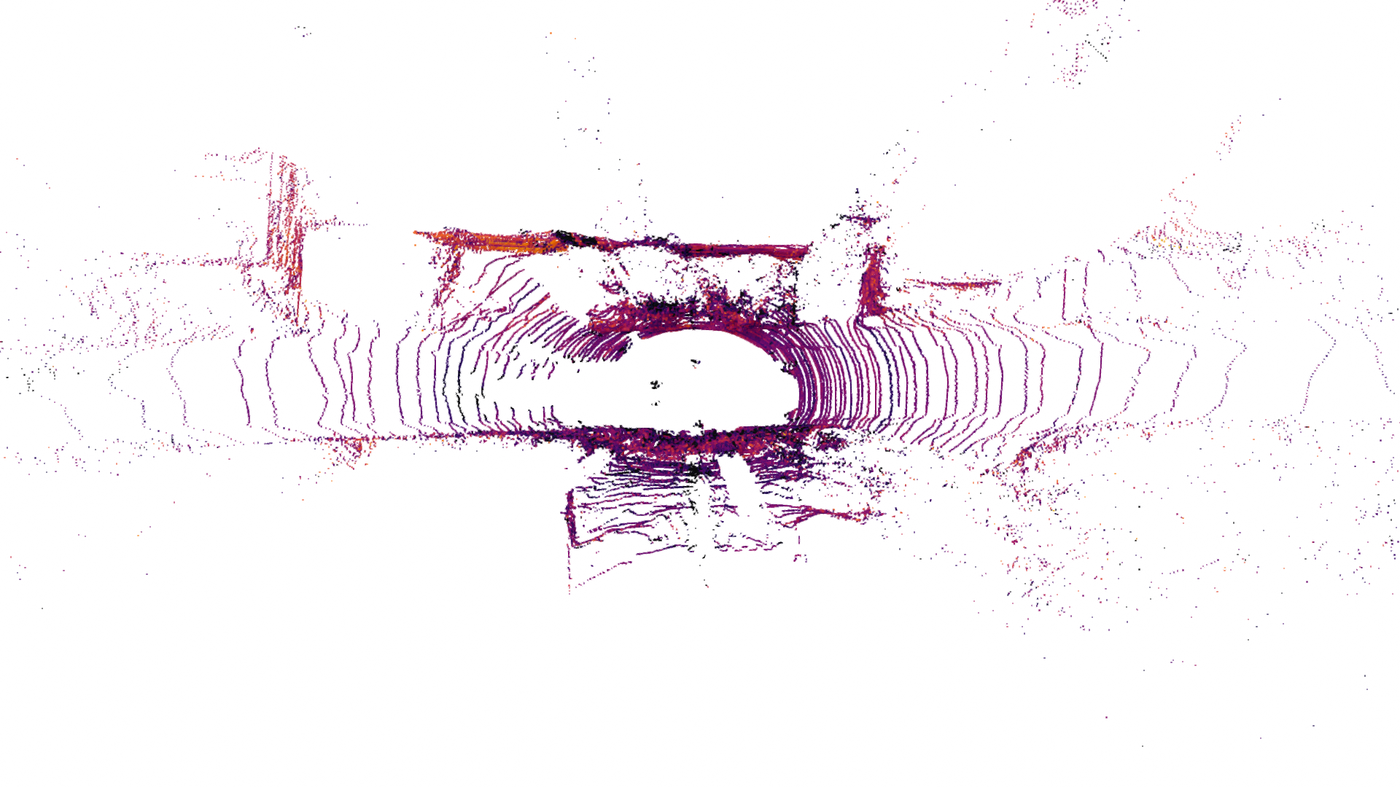} \\
\adjincludegraphics[width=0.33\textwidth, trim={{.2\width} {.2\height} {.2\width} {.2\height}},clip]{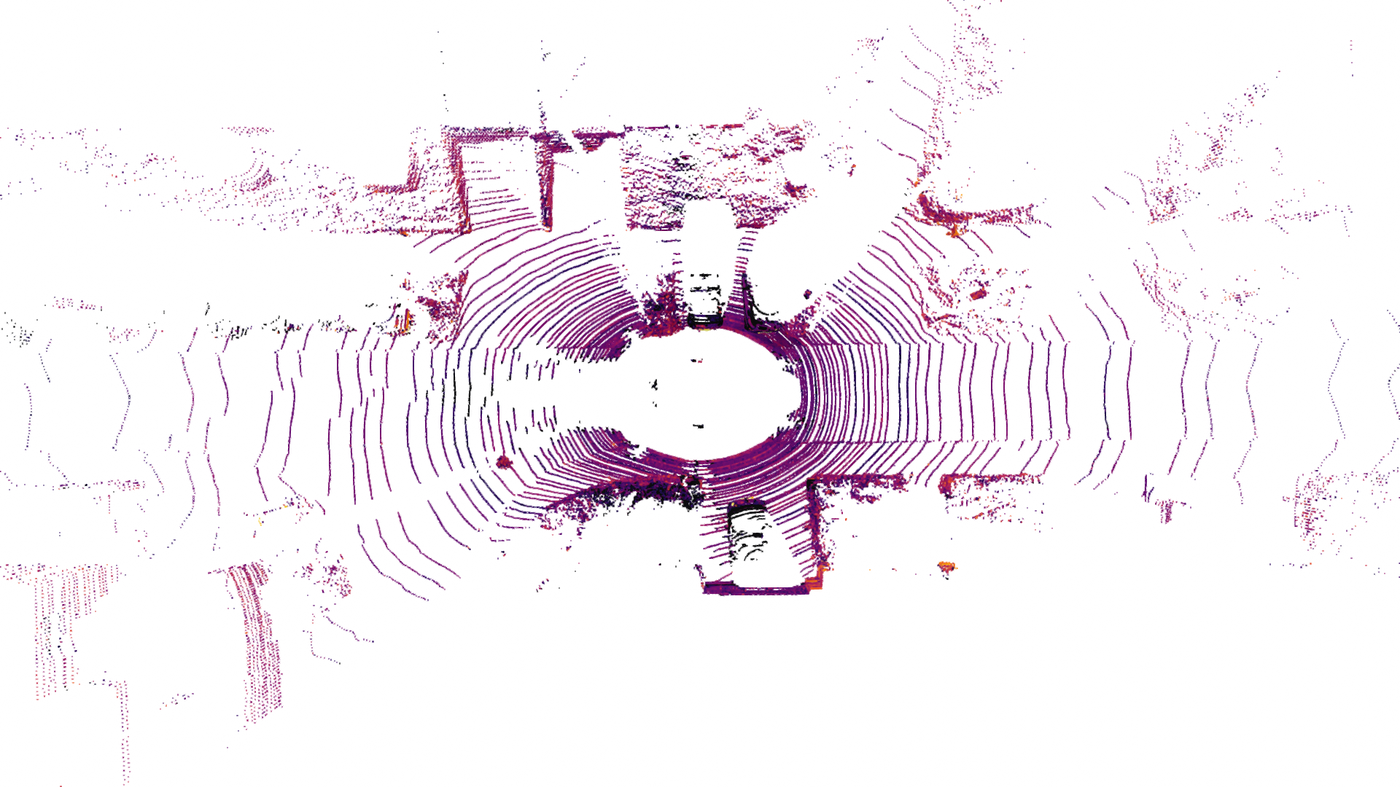} & 
\adjincludegraphics[width=0.33\textwidth, trim={{.2\width} {.2\height} {.2\width} {.2\height}},clip]{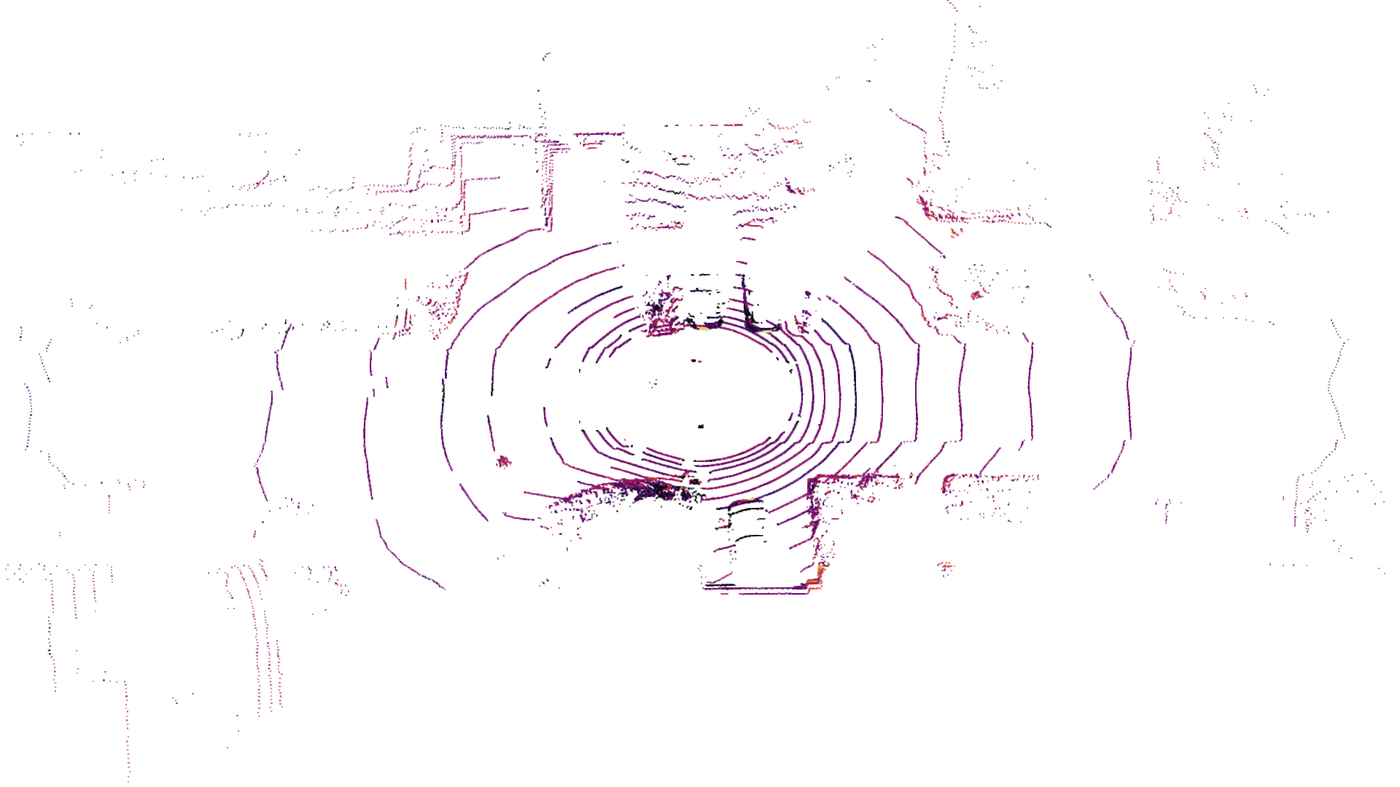} & 
\adjincludegraphics[width=0.33\textwidth, trim={{.2\width} {.2\height} {.2\width} {.2\height}},clip]{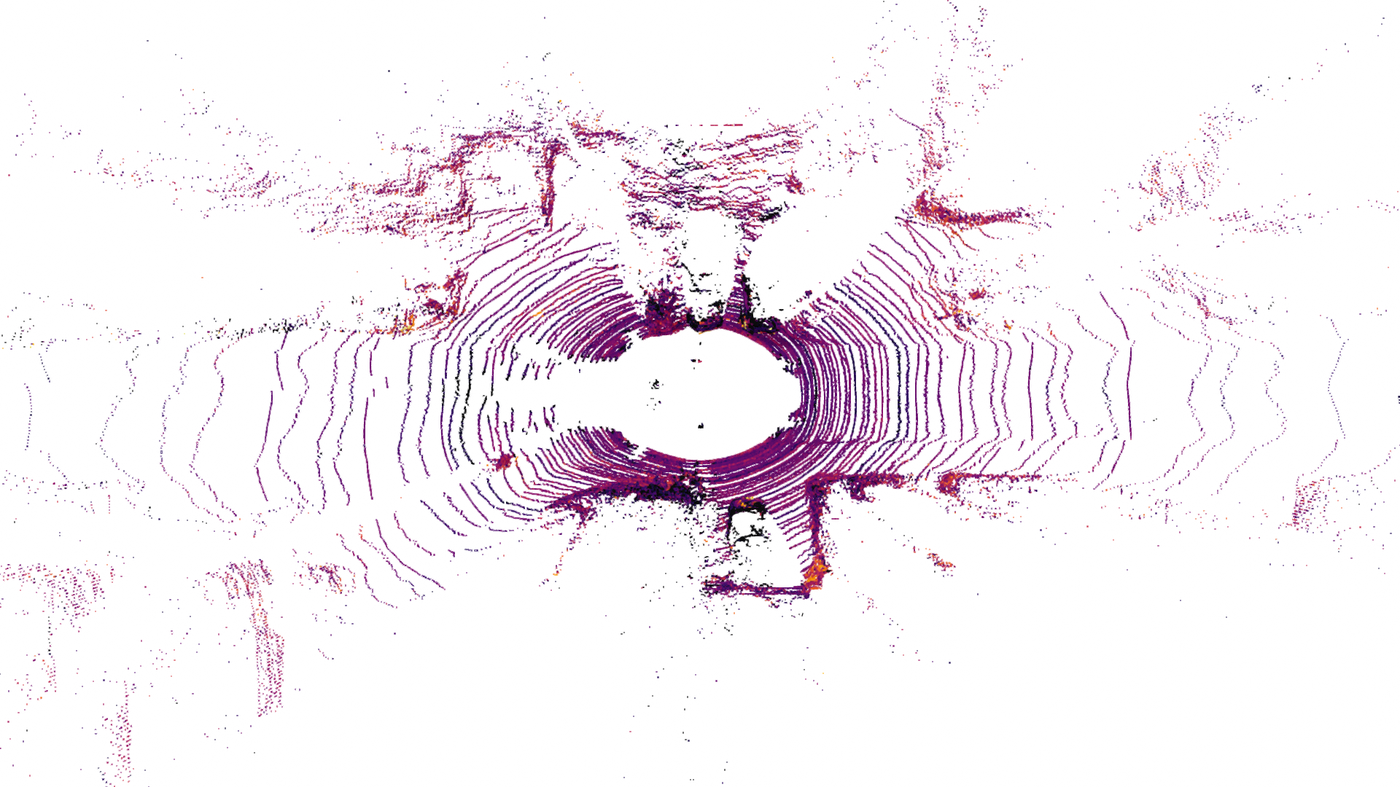} \\
\adjincludegraphics[width=0.33\textwidth, trim={{.2\width} {.2\height} {.2\width} {.2\height}},clip]{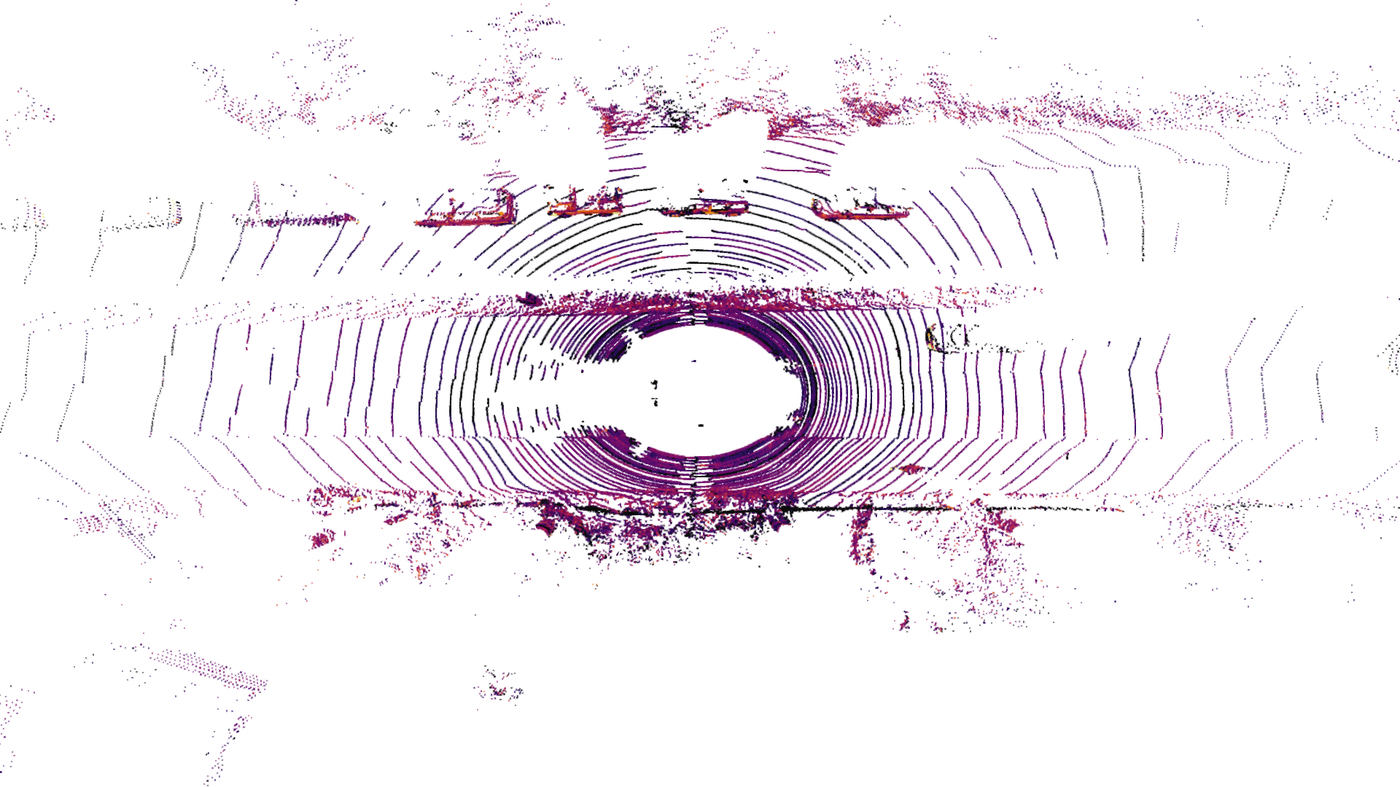} & 
\adjincludegraphics[width=0.33\textwidth, trim={{.2\width} {.2\height} {.2\width} {.2\height}},clip]{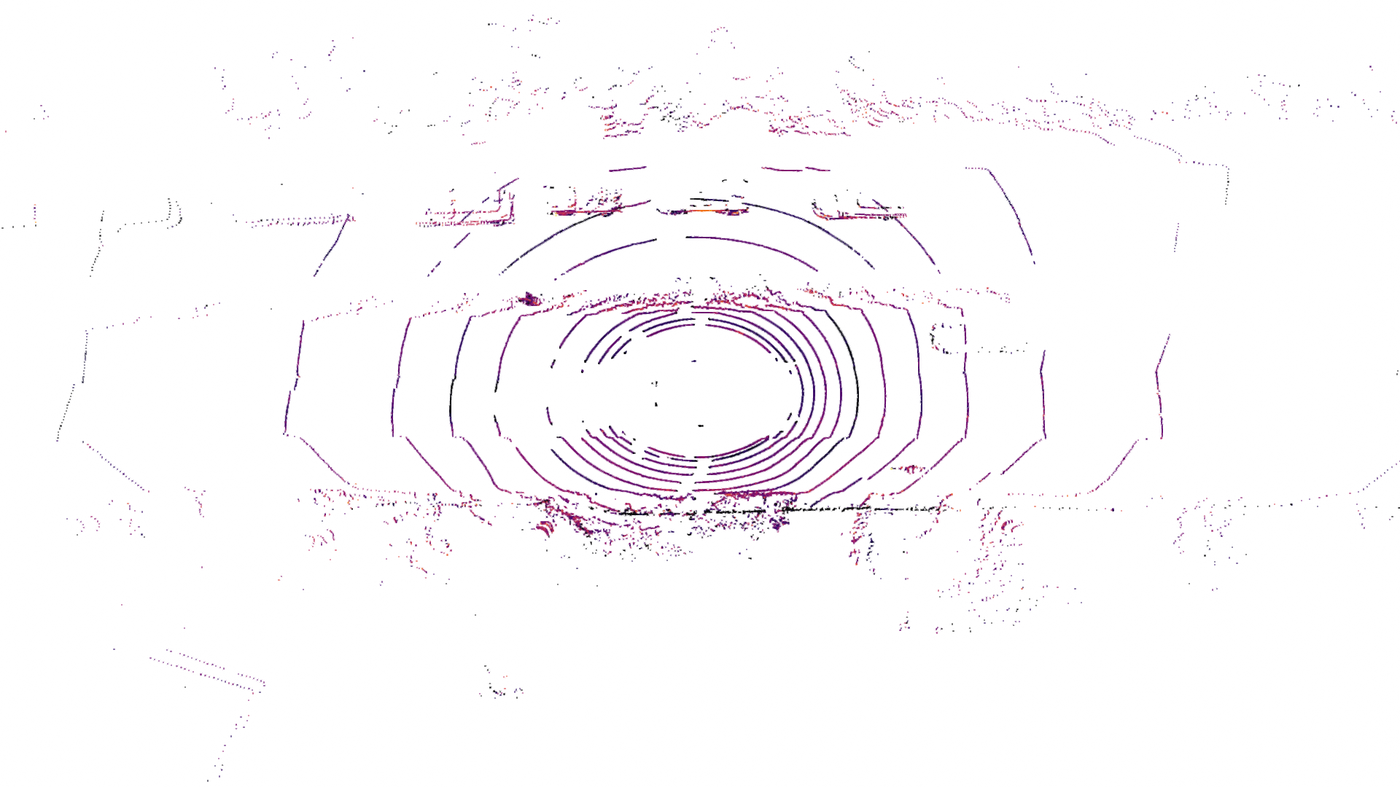} & 
\adjincludegraphics[width=0.33\textwidth, trim={{.2\width} {.2\height} {.2\width} {.2\height}},clip]{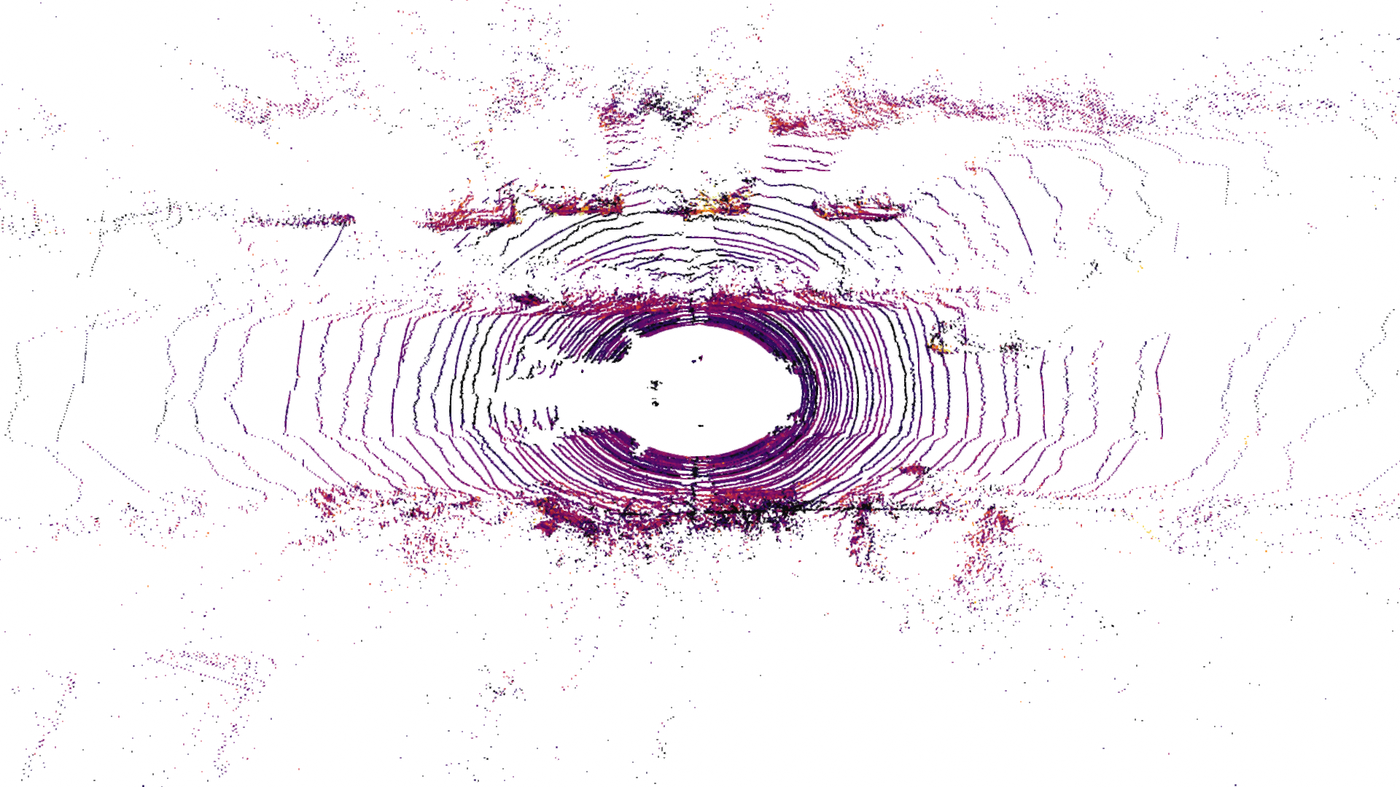} \\
\adjincludegraphics[width=0.33\textwidth, trim={{.2\width} {.2\height} {.2\width} {.2\height}},clip]{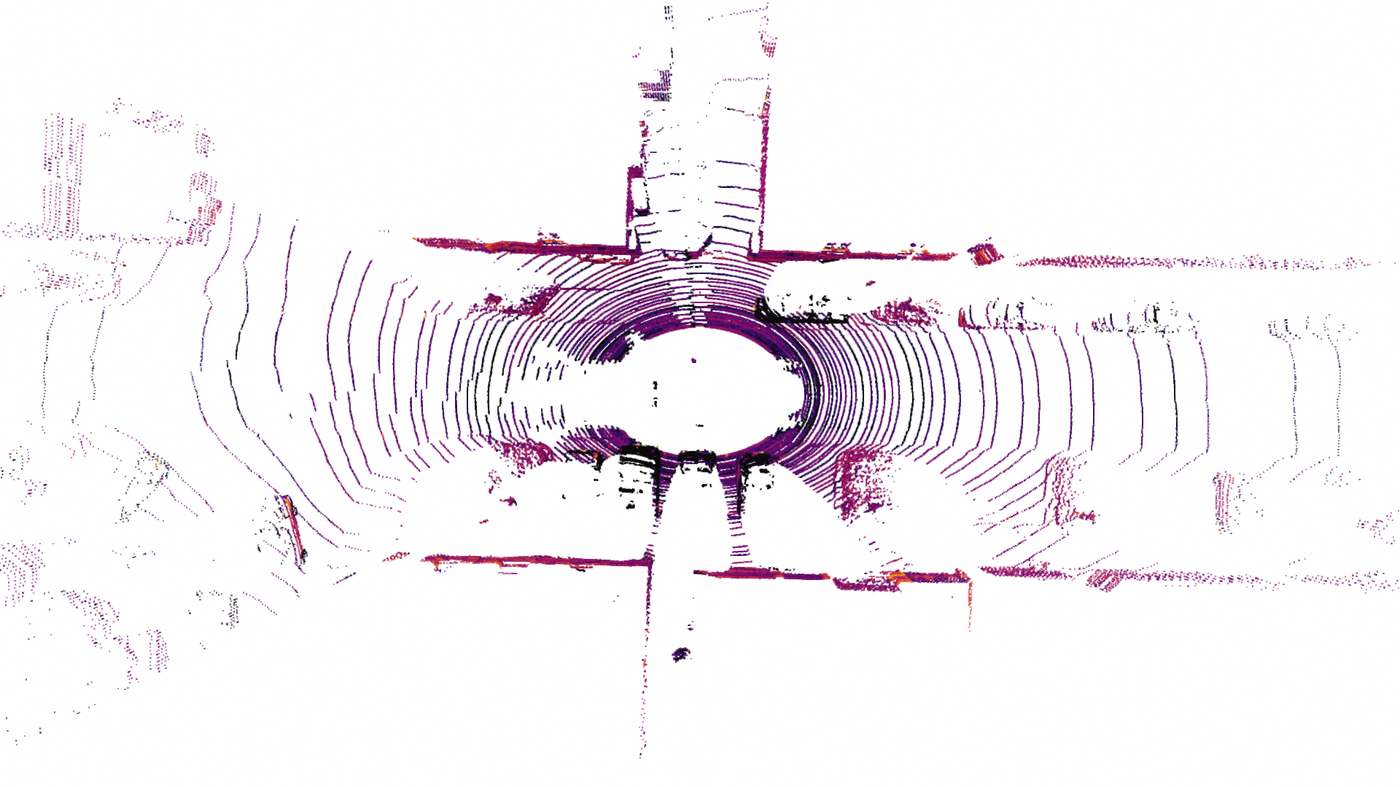} & 
\adjincludegraphics[width=0.33\textwidth, trim={{.2\width} {.2\height} {.2\width} {.2\height}},clip]{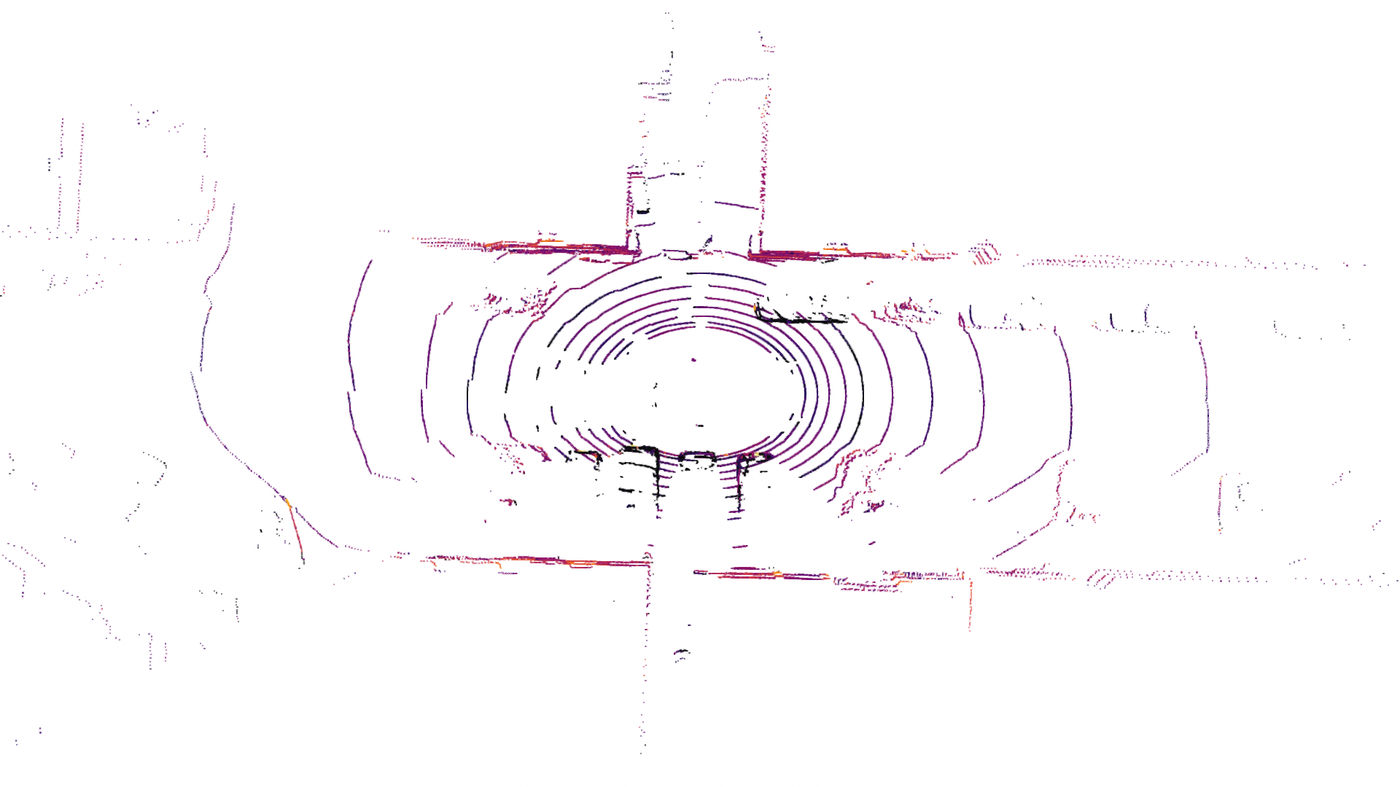} & 
\adjincludegraphics[width=0.33\textwidth, trim={{.2\width} {.2\height} {.2\width} {.2\height}},clip]{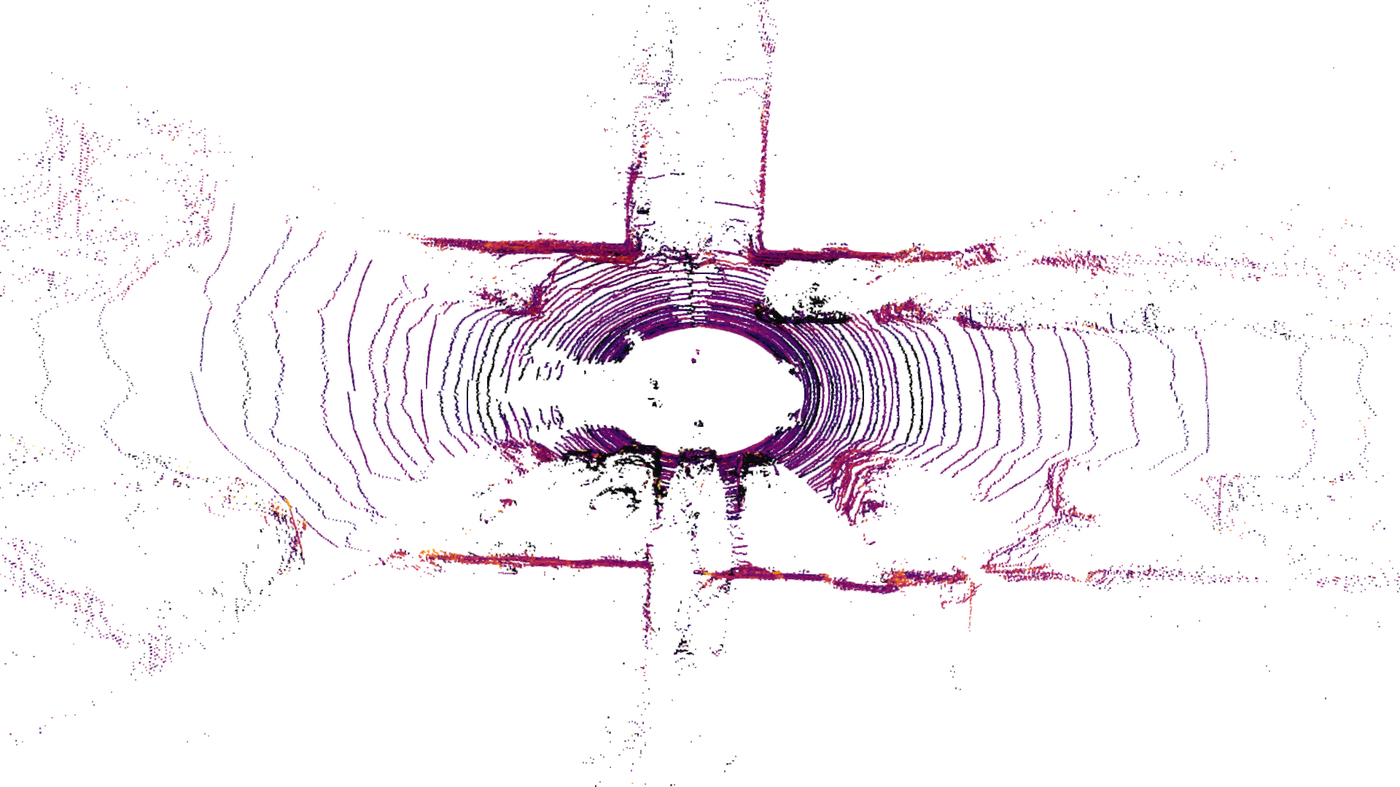} \\
\adjincludegraphics[width=0.33\textwidth, trim={{.2\width} {.2\height} {.2\width} {.2\height}},clip]{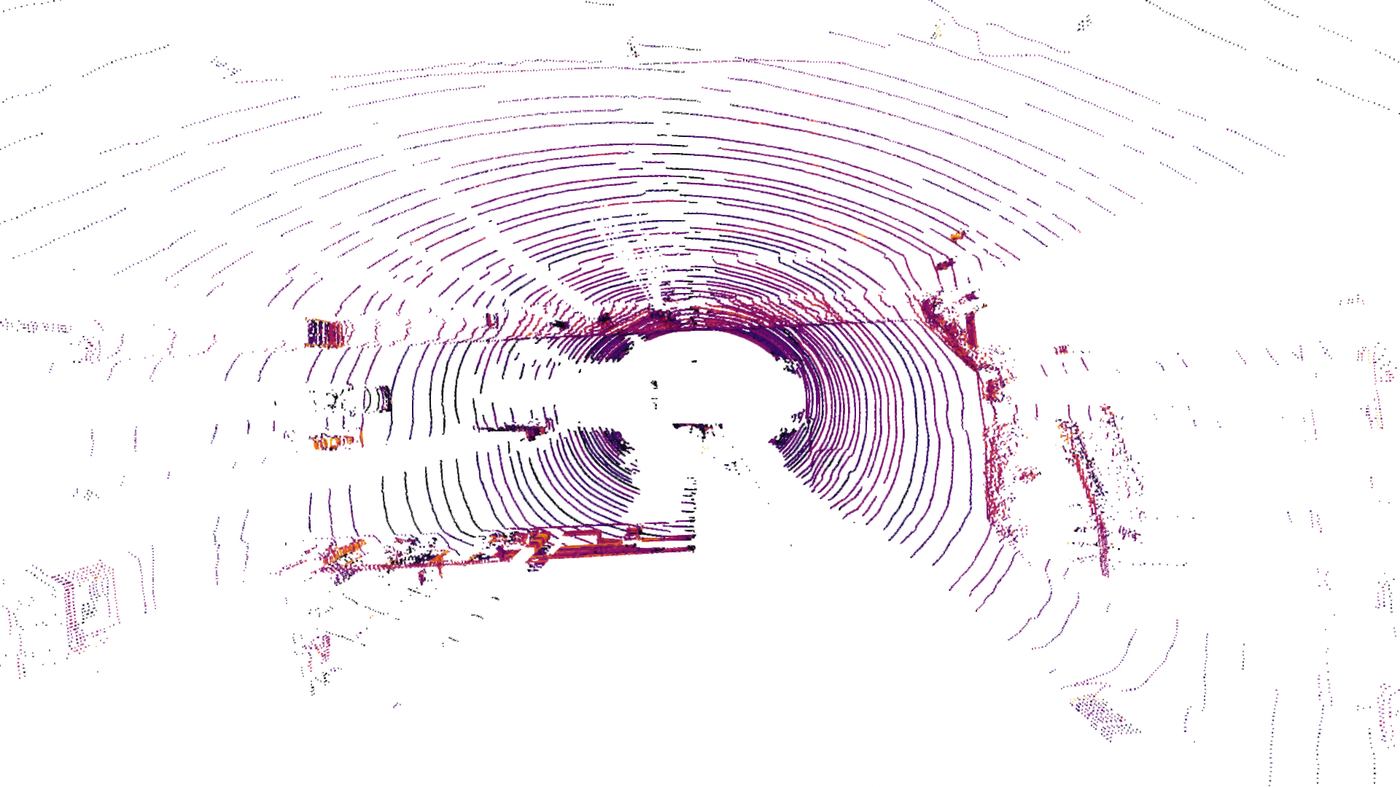} & 
\adjincludegraphics[width=0.33\textwidth, trim={{.2\width} {.2\height} {.2\width} {.2\height}},clip]{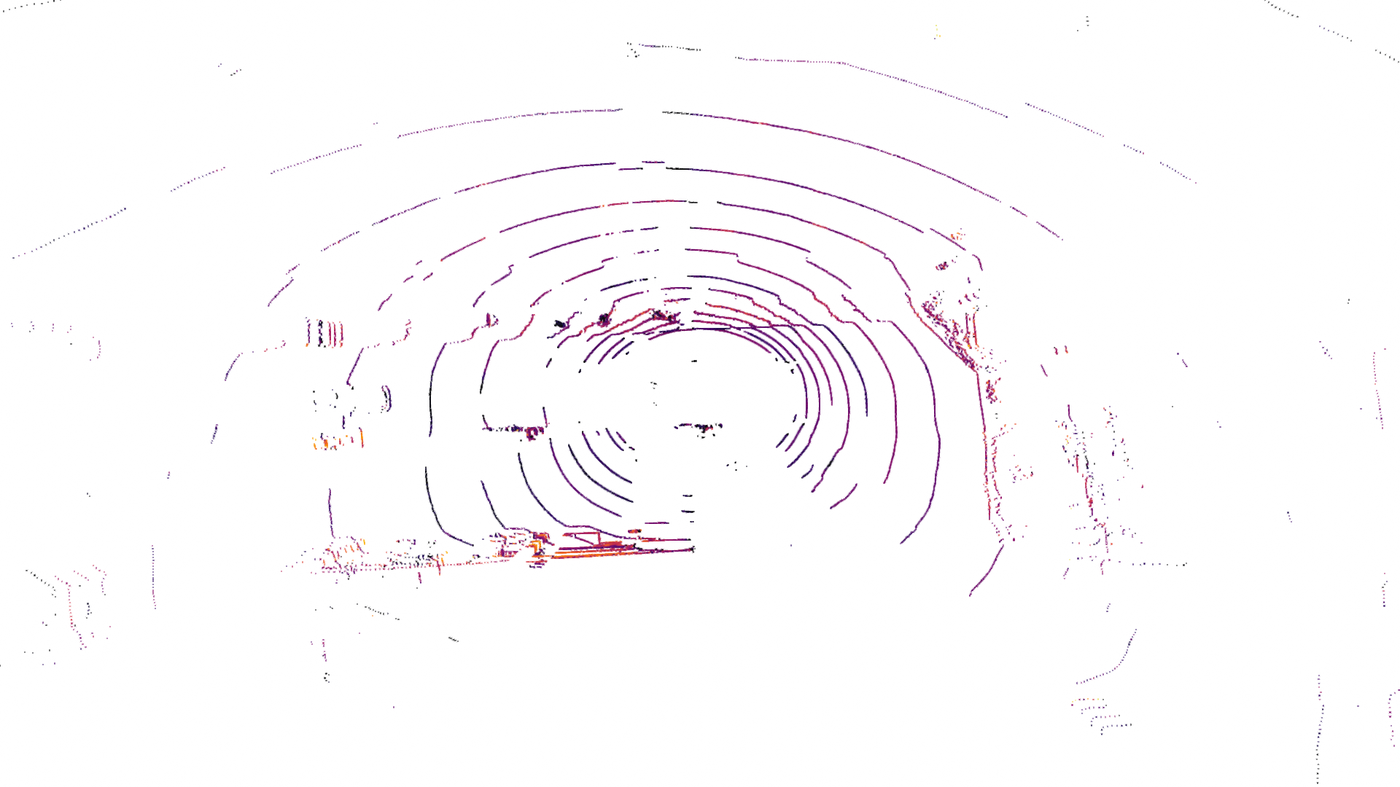} & 
\adjincludegraphics[width=0.33\textwidth, trim={{.2\width} {.2\height} {.2\width} {.2\height}},clip]{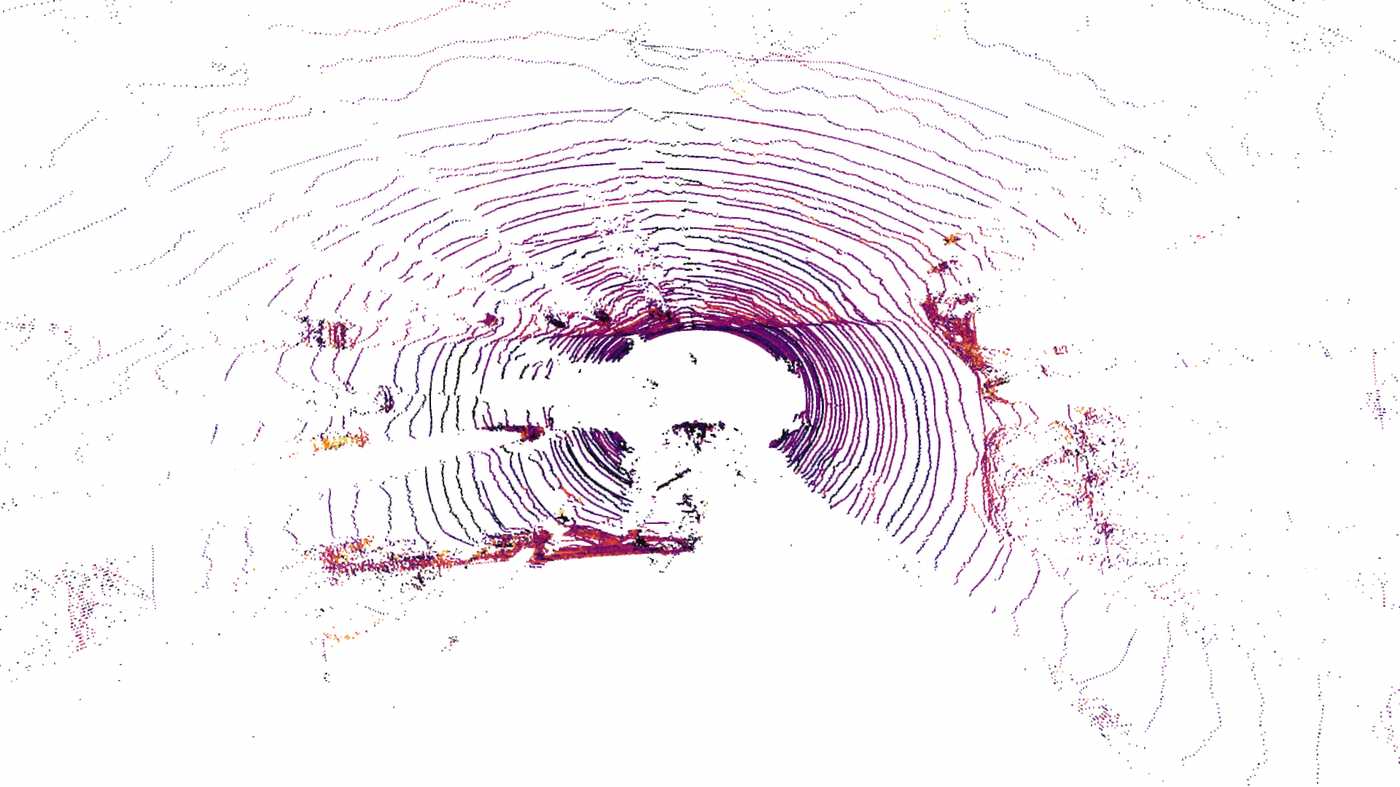} \\
\label{tab:densify}
\end{tabular}
\captionof{figure}{Additional Results for Unsupervised LiDAR Densification (continued).}
\label{fig:more_densification2}
\end{table}

\clearpage
\bibliographystyle{splncs04}
\bibliography{main}